\documentclass[10pt]{article}
\usepackage[preprint]{tmlr}%
\usepackage[utf8]{inputenc}
\usepackage[T1]{fontenc}
\usepackage{amsmath,amssymb}
\usepackage{booktabs}
\usepackage{longtable}
\usepackage{array}
\usepackage{graphicx}
\usepackage{xcolor}
\usepackage{microtype}
\usepackage{enumitem}
\usepackage{url}
\usepackage{tikz}
\usetikzlibrary{arrows.meta,positioning}
\usepackage{placeins}
\usepackage[colorlinks=true,linkcolor=blue!60!black,citecolor=blue!60!black,urlcolor=blue!60!black]{hyperref}

\newcommand{\yes}{\ensuremath{\checkmark}}
\newcommand{\half}{\ensuremath{\odot}}
\newcommand{\no}{\ensuremath{\times}}

\newif\ifdraftnotes
\draftnotestrue
\newcommand{\ph}[2]{\ifdraftnotes\textcolor{red!70!black}{[#1\ifx\\#2\\\else: #2\fi]}\fi}

\usepackage{xcolor}

\title{Modeling Authentic Research Trajectories: A Graph View of Research Ideation}

\author{\name Ruochen Li\email ruochen.li@utdallas.edu \\[2pt]
      \name Zhiyu Chen  \email zhiyu.chen2@utdallas.edu \\[2pt]
      \addr Department of Computer Science, University of Texas at Dallas}

\def\month{10}
\def\year{2026}
\begin{document}

\maketitle

\begin{abstract}
The research idea presented in a published paper is rarely the initial idea of the project. Between the two lies an iterative, closed-loop cognitive process, in which researchers conduct experiments, derive insights from the results, identify new challenges, explore the literature, and revise the idea, with each transition motivated by evidence and reasoning that they can articulate.
We refer to the path that one idea takes through this process, with the evidence and reasoning behind each step, as its authentic research trajectory. Although it underlies every scientific contribution, it is rarely recorded.
Existing LLM-based research agents generate an idea once, refine it against reviews, retrieved literature, or the results of their own experiments, or search over candidate ideas and experiments. In all of them the object that evolves is the text of the idea, yet the evidence and reasoning that connected one version to the next are discarded.
In this paper, we propose the Research Progress Graph (RPG), a new representation of authentic and complete research trajectories.
An RPG records the evolution of an idea as typed research states connected by cognitive transitions, each annotated with a temporal index and the reasoning process behind it.
The representation thus makes the evolution of the idea an explicit object of study, preserving the intermediate states, their temporal order, and the reasoning that connects each piece of evidence to the resulting idea update.
We ground the representation in authentic research trajectories whose steps their authors justify in their annotations and papers.
We compare the representations of thirty-five existing research agents with the RPG and find that each is an incomplete RPG; none maintains the state that changed, the evidence that changed it, and the reasoning that connects the two as a single evolving record.
Modeling research ideation with the RPG changes how research agents should be designed, evaluated, and trained. Agents predict the next research state instead of a complete idea, ideas are evaluated as trajectories instead of final texts, and authentic trajectories serve as training data.
Together, these directions shift research agents from generating static research ideas to explicitly modeling the evolution of research ideas.

\end{abstract}

\section{Introduction}\label{sec:intro}
\begin{figure}[t]
\centering
\definecolor{cogF}{HTML}{E9EEF7}\definecolor{cogD}{HTML}{7F91B2}%
\definecolor{litF}{HTML}{F8EDE0}\definecolor{litD}{HTML}{C6A271}%
\definecolor{wldF}{HTML}{E5F1EB}\definecolor{wldD}{HTML}{82A896}%
\definecolor{genF}{HTML}{F3F3F3}\definecolor{genD}{HTML}{B3B3B3}%
\definecolor{rsF}{HTML}{FAF7F0}\definecolor{rsD}{HTML}{D1C9B6}%
\begin{tikzpicture}[x=1cm,y=1cm,
  font=\footnotesize,
  n/.style={circle, line width=0.7pt, minimum size=0.72cm, inner sep=0pt, font=\footnotesize, text=black!80},
  cog/.style={fill=cogF, draw=cogD},
  lit/.style={fill=litF, draw=litD},
  wld/.style={fill=wldF, draw=wldD},
  gen/.style={fill=genF, draw=genD},
  e/.style={-{Latex[length=2mm]}, line width=0.65pt, draw=black!60},
  ed/.style={-{Latex[length=2mm]}, line width=0.65pt, dashed, draw=black!35},
  t/.style={font=\footnotesize, fill=white, inner sep=1pt, text=black!65},
  note/.style={font=\footnotesize, text=black!50},
  lbl/.style={font=\footnotesize, align=center, text width=3.0cm, anchor=north},
  hd/.style={font=\small\bfseries, anchor=west, text=black!80}
]
\node[hd] at (0,0) {Research Progress Graph};
\node[n,cog] (rq)  at (0.9,-1.3) {RQ};
\node[n,cog] (i0)  at (3.2,-1.3) {I};
\node[n,wld] (ce0) at (5.5,-1.3) {CE};
\node[n,wld] (r0)  at (7.8,-1.3) {R};
\node[n,cog] (in0) at (10.1,-1.3) {IN};
\node[n,cog] (c1)  at (12.4,-1.3) {C};
\node[n,lit] (ls1) at (14.7,-1.3) {LS};
\node[n,lit] (l0)  at (3.2,-2.9) {L};
\node[n,lit] (l1)  at (14.7,-4.5) {L};
\node[n,cog] (in1) at (12.4,-4.5) {IN};
\node[n,cog] (c2)  at (10.1,-4.5) {C};
\node[n,cog] (in2) at (7.8,-4.5) {IN};
\node[n,cog] (i1)  at (5.5,-4.5) {I};
\node[n,wld] (ce1) at (3.2,-4.5) {CE};
\node[n,wld] (r1)  at (0.9,-4.5) {R};
\draw[e] (rq) -- node[t]{1} (i0);
\draw[e] (l0) -- node[t]{1} (i0);
\draw[e] (i0) -- node[t]{2} (ce0);
\draw[e] (ce0) -- node[t]{3} (r0);
\draw[e] (r0) -- node[t]{4} (in0);
\draw[e] (in0) -- node[t]{5} (c1);
\draw[e] (c1) -- node[t]{6} (ls1);
\draw[e] (ls1) -- node[t]{7} (l1);
\draw[e] (l1) -- node[t]{8} (in1);
\draw[e] (in1) -- node[t]{9} (c2);
\draw[e] (c2) -- node[t]{10} (in2);
\draw[e] (l0) -- node[t,pos=0.5]{10} (in2);
\draw[e] (in2) -- node[t]{11} (i1);
\draw[e] (i1) -- node[t]{12} (ce1);
\draw[e] (ce1) -- node[t]{13} (r1);
\node[note, anchor=south] at (7.8,-0.85) {evidence from the world};
\node[note, anchor=south east] at (15.1,-0.85) {evidence from the literature};
\node[note, anchor=west] at (3.75,-2.9) {a state is read again later (10)};
\node[note, anchor=west] at (0.1,-3.6) {numbers: temporal index};
\node[font=\footnotesize, align=left, text width=4.0cm, draw=rsD, fill=rsF, rounded corners=2pt, inner sep=3pt, anchor=north] (rs) at (6.3,-5.3) {reasoning process on transition 11: why this evidence changes this part of the idea};
\draw[ed] (6.65,-5.3) -- (6.65,-4.78);
\node[n,cog,minimum size=0.42cm] at (9.1,-5.6) {};
\node[anchor=west] at (9.35,-5.6) {cognition (RQ, I, IN, C)};
\node[n,lit,minimum size=0.42cm] at (13.0,-5.6) {};
\node[anchor=west] at (13.25,-5.6) {literature (LS, L)};
\node[n,wld,minimum size=0.42cm] at (9.1,-6.2) {};
\node[anchor=west] at (9.35,-6.2) {world (CE, R)};
\node[n,gen,minimum size=0.42cm] at (13.0,-6.2) {};
\node[anchor=west] at (13.25,-6.2) {untyped text or score};
\draw[dashed, black!35] (0,-6.8) -- (16.4,-6.8);
\node[hd] at (0,-7.15) {Existing representations};
\begin{scope}[shift={(0.2,-8.0)}]
\node[note, anchor=south] at (1.4,0.3) {prompt, papers};
\node[n,gen] (a1) at (0.5,0) {};
\node[n,gen] (a2) at (1.4,0) {};
\node[n,gen] (a3) at (2.3,0) {};
\node[n,cog] (b)  at (1.4,-1.5) {I};
\draw[e] (a1) -- (b); \draw[e] (a2) -- (b); \draw[e] (a3) -- (b);
\node[lbl] at (1.4,-2.0) {(1) one-shot generation};
\end{scope}
\begin{scope}[shift={(3.5,-8.0)}]
\node[n,cog] (i) at (0.5,-0.75) {I};
\node[n,gen] (rv) at (2.5,-0.75) {};
\draw[e] (i) to[bend left=40] node[t]{review} (rv);
\draw[e] (rv) to[bend left=40] node[t]{regenerate} (i);
\node[lbl] at (1.5,-2.0) {(2) revision against self-critique};
\end{scope}
\begin{scope}[shift={(6.9,-8.0)}]
\node[n,cog] (i) at (1.4,0.1) {I};
\node[note, anchor=west] at (1.8,0.1) {fixed};
\node[n,wld] (c1) at (0.6,-0.95) {CE};
\node[n,wld] (c2) at (2.2,-0.95) {CE};
\node[n,wld] (r1) at (0.6,-1.95) {R};
\node[n,wld] (r2) at (2.2,-1.95) {R};
\draw[ed] (i) -- (c1); \draw[ed] (i) -- (c2);
\draw[e] (c1) -- (r1); \draw[e] (c2) -- (r2);
\draw[e] (r1) to[bend left=55] (c1);
\node[lbl] at (1.4,-2.35) {(3) fixed idea, tree of experiments};
\end{scope}
\begin{scope}[shift={(10.0,-8.0)}]
\node[n,cog] (i0) at (0.4,-0.75) {I};
\node[n,wld] (r0) at (1.5,-0.75) {R};
\node[n,cog] (i1) at (2.9,-0.75) {I};
\draw[e] (i0) -- (r0); \draw[e] (r0) -- node[t,above=2pt]{score} (i1);
\node[lbl] at (1.65,-2.0) {(4) results to a new idea through a score};
\end{scope}
\begin{scope}[shift={(13.3,-8.0)}]
\draw[dashed, black!40, rounded corners=3pt] (-0.1,0.45) rectangle (2.9,-1.95);
\node[n,cog] (p) at (0.5,0) {RQ};
\node[n,cog] (h) at (2.3,0) {I};
\node[n,cog] (m) at (0.5,-1.5) {C};
\node[n,cog] (a) at (2.3,-1.5) {I};
\draw[e] (p) -- (h); \draw[e] (h) -- (a); \draw[e] (p) -- (m); \draw[e] (m) -- (a);
\node[lbl] at (1.4,-2.35) {(5) typed idea graph, no evidence enters};
\end{scope}
\end{tikzpicture}
\caption{
The Research Progress Graph (top) and the structures by which existing research agents represent ideation (bottom). An RPG records typed states, the temporal order of the transitions between them, and the reasoning process behind each transition, with results and literature entering as states that change the idea. Each existing structure keeps a subset of these elements and can be viewed as an incomplete RPG.}
\label{fig:intro}
\end{figure}
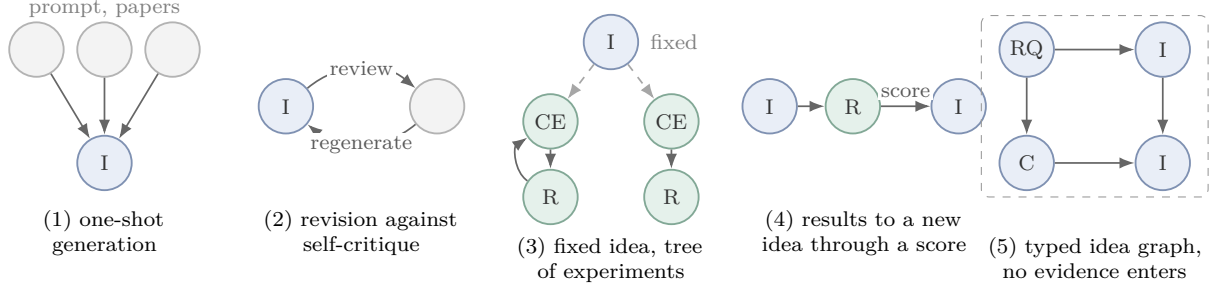

Scientific ideation is an iterative, closed-loop cognitive process, and the research idea presented in a published paper is rarely the initial idea of the project.
Instead, the idea evolves through a continuous cycle of literature exploration, hypothesis formulation, implementation, experimentation, insight generation, challenge identification, and idea revision, in which researchers recognize patterns in heterogeneous evidence and gradually transform an incomplete idea into a mature scientific contribution.
Each step of this cycle is driven by evidence, whether an experimental result or a paper, and by a reasoning process that the researcher can articulate.
We refer to the complete sequence of research states and transitions, together with the evidence and reasoning behind each transition, as the \emph{authentic research trajectory} of an idea.
Figure~\ref{fig:intro} (top) shows one such trajectory. This trajectory is only one example. Trajectories are free-form, in that the states an idea passes through, their order, and the number of rounds all vary from one idea to the next (\S\ref{sec:cases}).

Existing LLM-based research agents represent this process only partially, and in an artificial form.
Early work formulated ideation as one-shot generation, in which an LLM produces a research idea from a topic or a set of papers \citep{si2024novel}, so this formulation represents only the final idea. Other agents refine the idea iteratively against reviewer feedback \citep{baek2024researchagent}, retrieved literature \citep{wang2024scimon,hu2024nova}, or the results of experiments the agent runs itself~\citep{yuan2025dolphin,weng2025deepscientist}, and the most recent agents organize the refinement as a search over candidate ideas or experiments~\citep{yamada2025aiscientist2,rdagent2025}.
These agents differ in what they feed back, but they all represent the same object, the current text of the idea, and each round turns it into the next idea (Figure~\ref{fig:intro}, bottom).
In most of them, a round takes the whole idea as input and returns a new idea, while the evidence that caused the change and the reasoning that connected the two versions stay in a prompt that is discarded afterward.
Insights, challenges, and the literature searched for in response are rarely kept as objects of their own.
Even when an agent records a result or a reason, it stores only a score or a summary.
Each representation therefore keeps one part of the trajectory, whether the versions of the idea, the evidence, or the reasoning, and none records a change to the idea together with the evidence and reasoning behind it.

In this paper, we propose the Research Progress Graph (RPG), a new representation of authentic and complete research trajectories.
An RPG records the evolution of an idea as a graph of typed research states connected by cognitive transitions, each annotated with a temporal index and the reasoning process behind it. The states are of eight types: research question, idea update, insight, challenge, literature search, literature, code and experiment plan, and experimental result.
The representation makes the evolution of an idea an explicit object of study. It preserves the intermediate states that a paper omits and an agent discards, their chronological order, and the reasoning that connects each piece of evidence to the resulting idea update, so that a trajectory can be recorded as it happens, read by the next step, learned from, and evaluated.
The RPG is free-form, in that it prescribes no order of states and does not require a round to visit every type, and minimal, in that none of the eight state types can be removed without losing a stage of the ideation cycle, so that an RPG can be annotated by hand and generated by a model.

We ground the representation in authentic research trajectories from machine learning and natural language processing and use it to examine existing research agents and the design of future ones. \S\ref{sec:bg} describes how research ideas evolve and what existing research agents model of the process, and \S\ref{sec:rpg} defines the RPG.
We then use it in three ways.
First, we represent as RPGs eight authentic trajectories whose steps their authors justify in annotations or in their papers (\S\ref{sec:cases}).
In each of them, the decisive change is confined to one state, caused by a specific result or paper, and explained by the reasoning of the authors, which the RPG records on the transition. The eight state types cover every state of each trajectory.
Second, we compare the representations of thirty-five existing research agents with the RPG (\S\ref{sec:systems}).
Each is an incomplete RPG, and the comparison locates for each agent precisely which states and transitions it lacks.
Third, we show that modeling research ideation with the RPG changes how research agents should be designed, evaluated, and trained (\S\ref{sec:eval}, \S\ref{sec:insights}). Instead of a complete idea, an agent predicts the next research state from the trajectory so far. An idea is evaluated as a trajectory instead of a final text, and a blind study of 26 idea pairs shows that a judge reading only the final text rates an untested rewrite above the version its authors carried out in ten of the pairs, and that giving the judge retrieval does not prevent this. Authentic trajectories, recorded while a project runs, serve as training data. \S\ref{sec:disc} discusses open questions and limitations.

\section{Background: Research Idea Evolution}\label{sec:bg}
\subsection{Scientific discovery as idea evolution}
\label{sec:bg:evolution}\label{sec:bg:loop}

The view that research ideas are developed rather than generated has a long history in cognitive science and the history of science. \citet{boden2004creative} distinguishes three kinds of creativity, combinational, exploratory, and transformational, and in each of them a new idea is built from what already exists, by combining familiar ideas, exploring an existing conceptual space, or transforming that space. Computational models describe discovery as a step-by-step heuristic search through a space of possible hypotheses \citep{langley1987scientific}. Laboratory studies of scientific reasoning \citep{klahr1988dual}, observations of real research laboratories \citep{dunbar1997how}, and the history of a single mathematical idea \citep{lakatos1976proofs} describe a common cycle, in which a researcher proposes a conjecture, tests it, and revises it according to what the test shows. The trajectories in \S\ref{sec:cases} show this cycle in more detail. A researcher reads the literature, proposes a hypothesis, implements and tests it, draws insights from the results, identifies the challenges that these insights reveal, and revises the idea, often after a new literature search aimed at the challenge. Figure~\ref{fig:intro} (top) shows one round of this cycle. Two observations follow from it. First, the evidence that changes an idea arrives at specific moments, as an experimental result or a paper. Second, each revision is local and has a reason. It changes one part of the idea, and the researcher can explain how the evidence led to that change. A representation of how an idea evolves must therefore keep the states the idea passes through, their order in time, and the reasoning that links each piece of evidence to the change it caused.

\label{sec:bg:recover}This process is largely missing from the published record. A paper is written to present a result, not the path that led to it \citep{medawar1963fraud}. It reports the final idea and the experiments that support it, and it leaves out the intermediate reasoning, the implementation decisions, and the attempts that failed. The trajectory of an idea therefore has to be reconstructed, and reconstruction faces three problems. First, papers rarely say in what order the intermediate states occurred. Second, only the researchers know the reasons for each step. Third, any reconstruction happens after the fact, and people have limited access to their own past reasoning \citep{nisbett1977telling}. The order of the states and the evidence can be checked against code versions, experiment logs, and publication dates, but the stated reasons cannot. For this reason, each trajectory in this paper is annotated by its authors, traced and then confirmed by its authors, or traced from what its authors wrote in papers, reports, and blog posts. Existing datasets for hypothesis discovery reconstruct only the first step of a trajectory. They record, for each paper, the research background, the inspirations, and the hypothesis that combines them, annotated by PhD students for a few dozen papers \citep{yang2024moose,yang2025moosechem} or extracted by an LLM from over 100,000 papers \citep{yang2026moosestar}, but not the experiments, insights, and revisions that come after the hypothesis. Recording an RPG while the research is in progress avoids all three problems of reconstruction.

\subsection{LLM-based research agents}
\label{sec:bg:agents}

LLMs are now used to propose research ideas and to automate parts of the research process. The simplest approach generates an idea in a single pass, by prompting an LLM with a problem description, a topic, or a set of papers \citep{si2024novel}. Other agents revise the idea over several rounds, and they have developed in three directions. The first two change what drives the revision, feedback or literature, and the third changes how many candidates an agent keeps.

The first direction revises the idea in response to feedback, first critiques of the text and later the results of experiments. Outside ideation, Self-Refine and Reflexion showed that a model can improve its output by iterating on its own critique \citep{madaan2023selfrefine,shinn2023reflexion}. ResearchAgent applies this to ideas and revises them in response to the reviews and scores of reviewing agents \citep{baek2024researchagent}. The AI Scientist revises its idea through self-reflection and then revises its experiments according to their results \citep{lu2024aiscientist}. By late 2025, Dolphin, DeepScientist, ASI-Arch, AlphaResearch, Kosmos, and freephdlabor use the results of experiments to produce the next round of ideas \citep{yuan2025dolphin,weng2025deepscientist,liu2025asiarch,yu2025alpharesearch,kosmos2025,li2025freephdlabor}. The second direction gives the agent literature, first as retrieved papers and later as structured collections selected for the current idea. \citet{swanson1986fish} proposed, long before LLMs, that new hypotheses can come from connecting literatures that do not cite each other. MOOSE-Chem, which draws this parallel, pairs a research background with inspiration papers not previously known to be related to it \citep{yang2025moosechem}. SciMON retrieves related work along several routes \citep{wang2024scimon}, ResearchAgent builds a store of entities from the literature, and Chain of Ideas arranges papers into chains in the order in which the field developed \citep{li2025coi}. In these agents, the literature is a structured object rather than text added to a prompt. The third direction also changes how the candidates are related. Early agents keep one candidate and replace it in each round. Tree search keeps several branches and expands the most promising one, and graph search also allows branches to merge. Tree and graph search were first applied to reasoning steps \citep{yao2023tot,besta2024got} and later to ideas and experiments. Nova and IRIS build trees of ideas \citep{hu2024nova,garikaparthi2025iris}. AIDE and AI Scientist v2 build trees of code and experiments while the task or idea stays fixed \citep{jiang2025aide,yamada2025aiscientist2}. R\&D-Agent merges parallel traces and, in its second version, chooses parent nodes in a directed acyclic graph of solutions \citep{rdagent2025}.

Beyond ideation, end-to-end research agents combine LLMs with tools for literature search, coding, and the analysis of results, and coding agents increasingly carry out the implementation \citep{wang2024openhands}. Across the three directions, agents have gained richer inputs and more candidates to compare, but they judge each revision by its outcome. The insights, challenges, and reasoning through which a researcher moves from one version of an idea to the next are represented only in part. \S\ref{sec:systems} compares how each agent represents ideation.

\section{The Research Progress Graph}\label{sec:rpg}

The Research Progress Graph (RPG) represents the authentic research trajectory of an idea as a directed graph. Its nodes are research states of eight types, namely the research question, idea updates, insights, challenges, literature searches, literature, code and experiment plans, and experimental results. Its edges are the cognitive transitions by which the researcher moves from one or more states to the next, such as analyzing an experimental result, identifying a challenge, searching the literature, or integrating these signals into an idea update. Each transition carries two annotations. A \emph{temporal index} gives the global chronological order of the transition, and each state inherits the index of the transition that produced it, so that the indices define the temporal progression of the project and the graph can be replayed as a timeline. A \emph{reasoning process} describes the cognitive step by which the researcher moved from the source states to the target state. At the start of a project, the researcher provides a research question, optionally with a set of relevant papers, an initial idea, and an initial code and experiment plan with its first results, and these form the \emph{initial RPG}.

Figure~\ref{fig:case1} shows the RPG of Case 1. Its initial RPG holds the question, the pointer-generator paper, the initial idea, its implementation, and the first result. Transitions 4 to 11 take the project from that result to the final idea, and transitions 12 and 13 implement and run it.

The representation is not tied to a field. Its states and transitions are those of empirical research in general, and the trajectories and agents in this paper come from machine learning and natural language processing, the fields in which we can read the papers and code closely and have the authors confirm a trajectory. Nor is the representation limited to ideation. The code and experiment plan state and the experimental result state hold the implementation and the experiments of a project, so the same graph can represent the research process as a whole.
This paper centers on ideation, the part of the process that existing agents represent least (\S\ref{sec:systems}).

\subsection{Research states}\label{sec:rpg:states}

Table~\ref{tab:states} defines the eight state types with the running example of Case 1. The types fall into three groups by where a state comes from. The research question, idea updates, insights, and challenges are produced by the researcher's own cognition. Literature searches and literature are how the written record, and existing resources such as model and dataset repositories, enter the graph. Code and experiment plans and experimental results are how the world, or a derivation that can be checked, enters it. The eight types are minimal, in that none of them can be removed without losing a stage of the ideation cycle, and they have been sufficient for every trajectory we have examined.
They are not a closed ontology. Finer distinctions, such as an assumption, a mechanism, or a decision between alternatives, can be introduced as subtypes without changing the graph. A state records what was known or produced at a point in the project, not how it was produced. The representation does not specify which model generated a state or whether a person or an agent produced it.

\begin{table}[t]
\centering\small
\caption{State types of an RPG, with the running example of Case 1 (few-shot table-to-text generation).}
\label{tab:states}
\begin{tabular}{>{\raggedright\arraybackslash}p{0.17\linewidth}|>{\raggedright\arraybackslash}p{0.06\linewidth}|>{\raggedright\arraybackslash}p{0.35\linewidth}|>{\raggedright\arraybackslash}p{0.31\linewidth}}
\toprule
\textbf{Type} & \textbf{Abbr.} & \textbf{Definition} & \textbf{Example} \\
\midrule
Research question & RQ & The central scientific problem that motivates the project & How to do table-to-text generation with only a few hundred examples? \\
\hline
Idea update & I & A new or revised research idea proposed at a particular stage of the project, often based on prior insights, challenges, literature, and results & A switch between a pre-trained LM and a copy mechanism, as in the pointer-generator network \\
\hline
Insight & IN & A key observation or higher-level conclusion derived from the current RPG & Copying is easy to learn from few shots; language modeling is not \\
\hline
Challenge & C & An open problem, bottleneck, or unanswered question that arises during the research process & How to combine a pre-trained LM with copying? \\
\hline
Literature search & LS & The act of searching prior work; holds the search query & ``Few-shot language modeling'' \\
\hline
Literature & L & A unit of literature, such as one paper, optionally with a summary, or an existing external resource, such as a repository of models and datasets & The pointer-generator paper \\
\hline
Code and experiment plan & CE & The code implementation and experimental design used to test the current idea & Train with 50, 100, 200, and 500 examples \\
\hline
Experimental result & R & Outcomes obtained from running the experiments, or from working out a derivation that can be checked & Results of the model and baselines at each size \\
\bottomrule
\end{tabular}
\end{table}

\subsection{Cognitive transitions}\label{sec:rpg:edges}

A transition connects one or more source states to one target state. Its source states are the evidence for the target. An insight drawn from a result has that result as a source, and an idea update justified by an insight and a paper has both as sources, so the evidence behind a step is read off the graph rather than stated separately. Many transitions in the case trajectories have more than one source, and the note under each case figure names them. The temporal index orders transitions, not states, which matters because a state can be the source of transitions long after it was produced. In Figure~\ref{fig:case1}, Literature \#0 enters at the start of the project and is read again at transition 10. The reasoning process annotated on a transition describes the cognitive step as the researcher took it, and it is specific to that step.
For the transition that produced Idea Update \#1 in Case 1, the reasoning records not that a pointer-generator network and a pre-trained language model were combined but why they were compatible. The pointer-generator is itself a switch between a language model and a copy distribution, so its language-model half could be replaced while the switch and the copy half were kept.

Grouped by the type of the target state, transitions are a small set of operations. \emph{Deriving an insight} reads one or more results, papers, or earlier states into an insight. \emph{Identifying a challenge} turns an insight, a result, or the research question into a challenge about one part of the current idea. \emph{Searching the literature} turns a challenge into a literature search whose query is recorded, and the search returns literature. \emph{Updating the idea} produces a new idea update from the previous one and from the insights and literature that justify the change; the update is localized, in that the new version differs from the previous one in named components and keeps the rest. \emph{Implementing and experimenting} produce a code and experiment plan and an experimental result from the current idea. A \emph{re-reading} takes a state that is already in the graph, usually a paper, as the source of a new transition under a new challenge. Re-readings are common in the recorded trajectories and are how a graph reuses what it already holds.

\subsection{Idea evolution on the graph}\label{sec:rpg:loop}

Idea evolution is a path on the graph that returns to an idea update. The initial RPG holds what the researcher starts with. From there the project adds states and transitions until a new idea update is reached, and a mature idea is the result of one or more such rounds. The RPG is \emph{free-form}. It prescribes no order of state types, any type may follow any other, several sources may feed one target, and a round need not visit every type. What the recorded trajectories most often show is the shape of Figure~\ref{fig:intro} (top). An experimental result or a paper enters the graph, an insight is derived from it, the insight raises a challenge about one component of the idea, the challenge directs a literature search, the literature yields a further insight, and the insights justify a localized change to the idea. The two states that enter from outside the researcher's cognition, the result and the literature, are where the evidence arrives, and every change to the idea in the recorded trajectories traces back, usually through insights and challenges, to one of them.

The trajectories depart from this shape in several ways, and the representation accommodates each without a new construct. A round can re-read a state already in the graph instead of searching, as transition 10 of Case 1 does. A challenge can be answered by a derivation rather than by a paper or an experiment, as when the closed-form optimum of the KL-constrained objective let direct preference optimization train the policy on preferences without a separate reward model \citep{rafailov2023dpo}. A derivation that can be checked is recorded as a result state, since, like an experiment, it returns an outcome that can contradict what the researcher expected. A result can come from people rather than from an experiment, as in Case 2, and an external resource rather than a paper can enter as a literature state, as the Hugging Face Hub does in Case 4. And a trajectory can span several papers, with each earlier paper recorded as a literature state that the next paper reads, as Case 5 records the first paper of its group as a literature state of the second; the same holds when the papers come from different groups. What the rounds share is that the new idea is a change to the previous one, that the change is caused by evidence that entered the graph, and that the reasoning behind it is recorded on the transition.

\subsection{Properties of the representation}\label{sec:rpg:props}\label{sec:rpg:agnostic}
An authentic research trajectory consists of the states an idea passes through, their order, and the evidence and reasoning behind each transition. A representation of such a trajectory must preserve each of these, and the five properties below state what this requires. The first three keep the parts of a trajectory that a paper omits, the intermediate states, their order, and the reasoning between them. The fourth makes idea evolution a closed loop driven by evidence rather than a text revised against itself. The fifth concerns who writes the graph, a researcher or an agent. Each property is stated with what is lost without it.

\begin{enumerate}[itemsep=1pt]
\item \emph{Typed states.} The trajectory passes through states of different kinds, and a question, an insight, a challenge, a paper, a plan, and a result play different roles in it. Keeping them as distinct objects is what lets a later step act on one of them in particular. Without types, a challenge is a sentence in a paragraph and cannot direct a search, and an insight cannot be checked against the result it was drawn from.
\item \emph{A temporal order that later steps read.} The order of the states is part of the trajectory, since what the researcher knew at each step is what had entered the graph by then. States and transitions are therefore kept in chronological order, and the next step takes the earlier ones as input. A stored history that nothing reads does not have this property, because the order then plays no part in what happens next.
\item \emph{Reasoning bound to the change it explains.} The reasoning behind a step is the part of the trajectory that only its author can supply, and it is specific to one change. It is therefore annotated on the transition that produced the change, not kept as a review detached from any particular edit or in a chain of thought that is discarded at the end of the round. Without this property, the record shows that the idea changed but not why, which is exactly what a published paper already shows.
\item \emph{Evidence enters the graph and changes the idea.} In an authentic trajectory, every change to the idea traces back to a result or a literature state. After an initial idea exists, literature retrieved for it or results of running it lead, through insights and challenges, to a new idea update. Self-critique is not evidence, and a result that changes the code but not the idea does not count, since in neither case has the idea evolved in response to something that could contradict it.
\item \emph{Humans can intervene at any stage.} The trajectories in this paper were produced by researchers, and in a project that runs an agent the researchers still edit, veto, or redirect the idea between steps. A representation that held only an agent's steps could record neither the researchers' trajectories nor their interventions. The RPG therefore does not distinguish who produced a state. A coding agent can produce code and experiment plans and experimental results, a retriever can produce literature, an ideation agent can produce insights, challenges, literature searches, and idea updates, and a person can produce any of them. The graph is the interface between them, so whether a person intervenes is a question of which transitions the person takes, not of whether a person is present. The same holds across research groups. A line of work handed from one group to another is one graph in which the hand-overs are literature states, and updating another group's idea is a transition that every follow-up paper takes.
\end{enumerate}
These are properties of the representation rather than of the evidence we offer for it. The trajectories of \S\ref{sec:cases} are taken from what their authors annotated or wrote, because only the authors can state the reasoning behind a transition.

\paragraph{Why the structure is stored explicitly.} A model with a long context window, or an agent with a memory store, could be given the whole history of an idea as text and left to find the structure itself. Such a transcript lacks three things that the properties above require. It does not say which sentence is an assumption, which is a claim already tested, and which the last result contradicted, so a challenge cannot be addressed to one component and an insight cannot be checked against its source (property 1). It records order but not attribution, since it contains the result and the next version but not the edge saying that one caused the other (properties 3 and 4). Finally, it cannot be edited locally, so the operation the cases show to be decisive, replacing one component for a stated reason while keeping the rest, has no counterpart in regenerating a text from a longer prompt. The RPG is compatible with long contexts, which are the natural way to present a linearized graph to a model. What it requires is only that the structure be explicit in what is stored and read.

\section{Case Studies: Authentic Research Trajectories}\label{sec:cases}

In this section we represent the trajectories of real ideas as RPGs. The resulting RPGs show that real trajectories fit the representation without a new construct, which is the evidence for the claim that the representation is authentic and complete. For each idea we traced the path from the initial idea to the published idea, typed every state, and drew the result as an RPG. Table~\ref{tab:cases} summarizes eight cases. Five are presented in this section. Case 1 is given in full as the worked example, and Cases 2 to 5 are each given as one paragraph and one figure. Cases 6 to 8 are given in Appendix~\ref{app:cases}, which also contains the state and transition listings from which every figure is drawn. For each case we ask whether the idea evolved through the cycle of \S\ref{sec:bg:evolution}, where evidence entered, and which parts of the trajectory an agent that regenerates the whole idea from a score or a summary could not have recorded.

\paragraph{Scope of the evidence.} Every case is an authentic trajectory, that is, a trajectory produced by the idea's own authors, with the reason for each step taken from what they annotated or wrote. This is required because the claim under test is that the reasoning behind an idea update is bound to the evidence that caused it, and only the authors can state what the reasoning was. Cases 1 to 4 are annotated, or traced and confirmed, by their authors, Case 2 spans two papers by one group, and Case 4 spans two versions of one paper. Cases 5 to 8 are traced from what their authors published. Case 5 spans two papers by one group, linked by their shared title, authors, and citations, and of the cases in the appendix, Case 6 is built from its authors' published narrative, Case 7 spans two papers by the same three authors, and Case 8's report records two abandoned attempts with their reasons.

\begin{table}[t]
\centering\footnotesize\setlength{\tabcolsep}{4pt}
\caption{Eight real ideas as RPGs. Cases 1 to 5 are presented in this section and Cases 6 to 8 in Appendix~\ref{app:cases}. Rounds counts the passes through the cycle of \S\ref{sec:bg:evolution}; Evidence lists the experimental result (R) and literature (L) states through which evidence entered.}
\label{tab:cases}
\begin{tabular}{>{\raggedright\arraybackslash}p{0.045\linewidth}|>{\raggedright\arraybackslash}p{0.10\linewidth}|>{\raggedright\arraybackslash}p{0.17\linewidth}|>{\raggedright\arraybackslash}p{0.275\linewidth}|>{\raggedright\arraybackslash}p{0.08\linewidth}|>{\raggedright\arraybackslash}p{0.22\linewidth}}
\toprule
\textbf{Case} & \textbf{Idea type} & \textbf{Initial idea} & \textbf{Published idea} & \textbf{Rounds} & \textbf{Evidence} \\
\midrule
1 & Method & Pointer-generator network for few-shot table-to-text & Pre-trained LM with a copy switch \citep{chen2020fewshot} & 1 & L (pointer-generator, re-read), R (own pilot), L (GPT-2), R (final) \\
\hline
2 & Application & Prompt GPT-4 to role-play a patient & PATIENT-$\Psi$, GPT-4 conditioned on clinician-written cognitive models and evaluated by mental-health experts and trainees \citep{wang2024patientpsi}; then CBT-Bench on the therapist's side \citep{zhang2024cbtbench} & 4 & R (expert interviews), L (CBT), L (transcripts), R (user study), L (prior CBT tasks), R (benchmark) \\
\hline
3 & Method (ideation agent) & Fine-tune an LLM on paper-to-idea pairs & LDC: reward models per dimension, dimensional controllers, sentence-level dynamic control \citep{li2024ldc} & 3 & L (prompting-based ideation), L (ResearchAgent's iterative feedback), L (trade-off in prior work), R $\times$2, R with human judges, R (later sweep) \\
\hline
4 & System (research agent) & A prompted idea agent and an experiment agent that runs the source paper's code & A ReAct experiment agent with retrieval and human feedback; the idea agent trained with fine-tuning and reward-model reinforcement learning \citep{li2024mlrcopilot} & 3 & L (prior systems), L (MLAgentBench), L (ResearchAgent), R (agent drift), L (Hugging Face Hub), R (five tasks), R (single-prompt baseline), L (the group's idea-generation method), R (retrained idea agent) \\
\hline
5 & Empirical insight & Apply the one-shot RLVR lens to on-policy distillation & ``Data-overfed but algorithm-starved''; inputs act as state generators \citep{fu2026rethinking2} & 3 & L (one-shot RLVR, Part I), R $\times$3 \\
\hline\hline
6 & Method & Fine-tune the student on teacher samples & On-policy distillation with per-token reverse KL, later used for recovery after mid-training (blog post, \citealp{thinkingmachines2025opd}) & 3.5 & R (SFT plateau), L (Qwen3 report), L (DAgger, GKD), R $\times$6 \\
\hline
7 & Evaluation of ideation & A large blind human study of LLM and expert ideas \citep{si2024novel} & Execute a subset of the same ideas; LLM ideas lose more on all four metrics rated at both stages \citep{si2025gap} & 1, a second left open & L, R (expert study), R (self-analysis), R (execution study), R (attribution) \\
\hline
8 & Method & GRPO after SFT on mathematics \citep{shao2024deepseekmath} & DeepSeek-R1: pure-RL R1-Zero, then cold start and multi-stage RL; distillation for small models; two branches abandoned \citep{guo2025deepseekr1} & 3, two abandoned branches, one fork & L (own prior work), R $\times$4 \\
\bottomrule
\end{tabular}
\end{table}

\subsection{Case 1, worked example: few-shot table-to-text generation}\label{sec:cases:1}

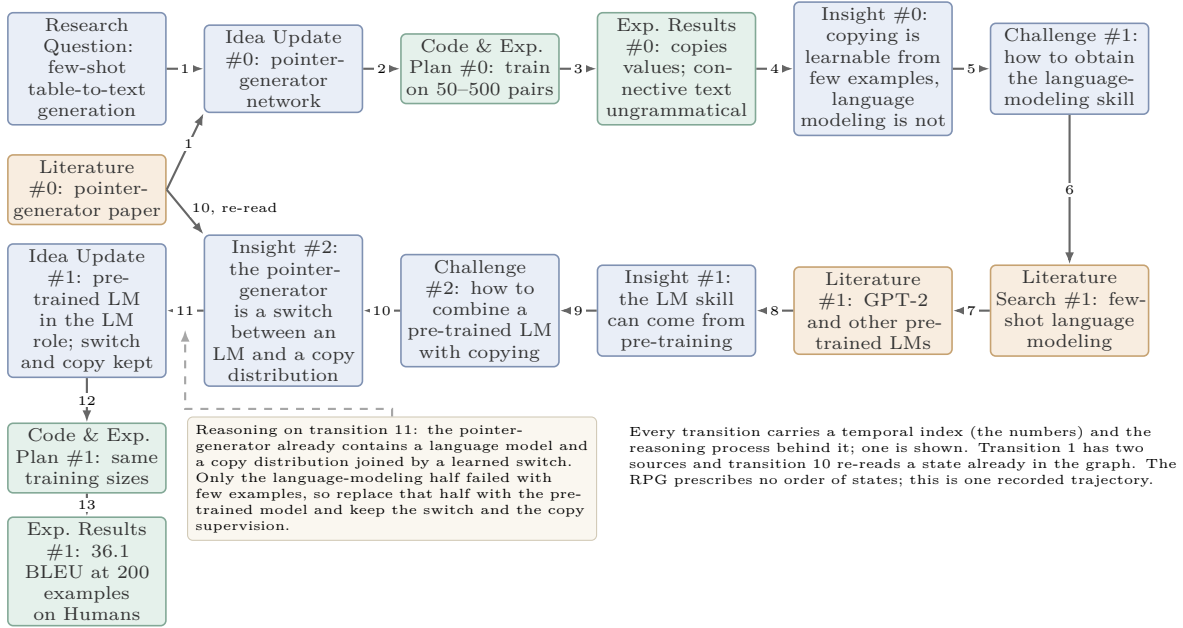
\begin{figure}[t]
\centering
\definecolor{cogF}{HTML}{E9EEF7}\definecolor{cogD}{HTML}{7F91B2}%
\definecolor{litF}{HTML}{F8EDE0}\definecolor{litD}{HTML}{C6A271}%
\definecolor{wldF}{HTML}{E5F1EB}\definecolor{wldD}{HTML}{82A896}%
\definecolor{rsF}{HTML}{FAF7F0}\definecolor{rsD}{HTML}{D1C9B6}%
\begin{tikzpicture}[
  font=\scriptsize,
  st/.style={line width=0.6pt, rounded corners=2pt, text=black!80, text width=1.95cm, minimum height=0.62cm, align=center, inner sep=2pt},
  ty/.style={draw, rounded corners=2pt, text width=1.9cm, minimum height=0.6cm, align=center, inner sep=2pt},
  cog/.style={fill=cogF, draw=cogD},
  lit/.style={fill=litF, draw=litD},
  wld/.style={fill=wldF, draw=wldD},
  txt/.style={draw, rounded corners=2pt, text width=2.2cm, minimum height=0.62cm, align=center, inner sep=2pt, fill=gray!8},
  pr/.style={draw, dashed, gray!70, rounded corners=2pt, text width=2.6cm, minimum height=0.62cm, align=center, inner sep=2pt, text=gray!70},
  e/.style={-{Latex[length=1.6mm]}, thick, draw=black!60},
  ed/.style={-{Latex[length=1.6mm]}, thick, dashed, draw=black!35},
  t/.style={font=\tiny, fill=white, inner sep=1pt},
  rs/.style={font=\tiny, align=left, text width=5.2cm, draw=rsD, fill=rsF, rounded corners=2pt, inner sep=3pt},
  grp/.style={font=\tiny, anchor=west, text width=1.7cm, align=left}
]
\node[st,cog] (crq) at (1.2,-0.2)  {Research Question: few-shot table-to-text generation};
\node[st,lit] (l0)  at (1.2,-1.8)  {Literature \#0: pointer-generator paper};
\node[st,cog] (i0)  at (3.8,-0.2)  {Idea Update \#0: pointer-generator network};
\node[st,wld] (ce0) at (6.4,-0.2)  {Code \& Exp.\ Plan \#0: train on 50--500 pairs};
\node[st,wld] (r0)  at (9.0,-0.2)  {Exp.\ Results \#0: copies values; connective text ungrammatical};
\node[st,cog] (in0) at (11.6,-0.2) {Insight \#0: copying is learnable from few examples, language modeling is not};
\node[st,cog] (c1)  at (14.2,-0.2) {Challenge \#1: how to obtain the language-modeling skill};
\node[st,lit] (ls1) at (14.2,-3.4) {Literature Search \#1: few-shot language modeling};
\node[st,lit] (l1)  at (11.6,-3.4) {Literature \#1: GPT-2 and other pre-trained LMs};
\node[st,cog] (in1) at (9.0,-3.4)  {Insight \#1: the LM skill can come from pre-training};
\node[st,cog] (c2)  at (6.4,-3.4)  {Challenge \#2: how to combine a pre-trained LM with copying};
\node[st,cog] (in2) at (3.8,-3.4)  {Insight \#2: the pointer-generator is a switch between an LM and a copy distribution};
\node[st,cog] (i1)  at (1.2,-3.4)  {Idea Update \#1: pre-trained LM in the LM role; switch and copy kept};
\node[st,wld] (ce1) at (1.2,-5.35) {Code \& Exp.\ Plan \#1: same training sizes};
\node[st,wld] (r1)  at (1.2,-6.85) {Exp.\ Results \#1: 36.1 BLEU at 200 examples on Humans};
\draw[e] (crq) -- node[t]{1} (i0);
\draw[e] (l0.east) -- node[t,pos=0.6]{1} (i0.south west);
\draw[e] (i0) -- node[t]{2} (ce0);
\draw[e] (ce0) -- node[t]{3} (r0);
\draw[e] (r0) -- node[t]{4} (in0);
\draw[e] (in0) -- node[t]{5} (c1);
\draw[e] (c1) -- node[t]{6} (ls1);
\draw[e] (ls1) -- node[t]{7} (l1);
\draw[e] (l1) -- node[t]{8} (in1);
\draw[e] (in1) -- node[t]{9} (c2);
\draw[e] (c2) -- node[t]{10} (in2);
\draw[e] (l0.east) -- node[t,pos=0.62,anchor=south west,inner sep=0.5pt]{10, re-read} (in2.north west);
\draw[e] (in2) -- node[t]{11} (i1);
\draw[e] (i1) -- node[t]{12} (ce1);
\draw[e] (ce1) -- node[t]{13} (r1);
\node[rs, anchor=north east] (rs11) at (7.95,-4.8) {Reasoning on transition 11: the pointer-generator already contains a language model and a copy distribution joined by a learned switch. Only the language-modeling half failed with few examples, so replace that half with the pre-trained model and keep the switch and the copy supervision.};
\draw[ed] (rs11.north) -- ++(0,0.1) -| (2.5,-3.65);
\node[font=\tiny, anchor=north west, text width=7.25cm] at (8.25,-4.8) {Every transition carries a temporal index (the numbers) and the reasoning process behind it; one is shown. Transition 1 has two sources and transition 10 re-reads a state already in the graph. The RPG prescribes no order of states; this is one recorded trajectory.};
\end{tikzpicture}
\caption{Case 1, few-shot table-to-text generation, as an RPG. One round, two experimental and two literature states of evidence. Transitions are numbered by temporal index; transition 1 has two sources, transition 10 re-reads a state already in the graph, and the reasoning process on transition 11 is shown. Listing in Appendix~\ref{app:cases:1}.}
\label{fig:case1}
\end{figure}

Case 1 is a method idea \citep{chen2020fewshot}, annotated, or traced and confirmed, by its authors. The project started from the question of how to generate text from a table with only a few hundred training pairs, at the moment GPT-2 had just appeared. The initial idea was the natural one in 2019, a pointer-generator network \citep{see2017pointer}, since the output mostly copies values from the table. The initial RPG therefore held a research question, a small literature set around the pointer-generator, an idea, an implementation, and a first result.

The first result was where outside information entered. With 50 to 500 examples the pointer-generator copied values reasonably but produced ungrammatical connective text. Analyzing that result gave the insight that the task needs two separable skills, copying and language modeling, and that copying is the easy one to learn from little data. That insight defined the challenge of learning the language-modeling skill from few examples. The challenge triggered a literature search, which returned the second piece of outside information, the pre-trained language models that had just been released. Reading them gave the insight that the language-modeling skill need not be learned from the task data at all, which narrowed the challenge to how a pre-trained language model can be combined with copying.

Resolving that challenge needed no new information. It needed a state already in the graph to be read again. Revisiting the pointer-generator paper with the new challenge in mind gave the decisive insight that the pointer-generator is structurally a switch between a language model and a copy distribution, so its language-model half can be replaced by the pre-trained model while the switch and the copy half are kept. The final idea is that hybrid, with the copy switch supervised explicitly, and the second implementation and its result closed the project (Figure~\ref{fig:case1}).

The trajectory is one round with two experimental and two literature states of evidence. The decisive step, transition 11, combines components, but it is not ``pointer-generator plus GPT-2''. It maps the new component onto one slot of the functional structure that transition 10 abstracted from the pointer-generator.

\emph{What a whole-idea regeneration would have lost.} A loop that receives the first result as a score (``BLEU 21 at 200 examples'') and regenerates the idea might well arrive at ``use GPT-2''. It cannot represent the actual edit, in which the language-model half of the switch was replaced and the switch and the copy half were kept, nor the reason on transition 11. That reason is why the final model keeps explicit copy supervision instead of fine-tuning GPT-2 alone, and it exists only as the annotation on the edge from the insight drawn from the re-read paper to the new idea. In a regeneration loop the diagnosis that copying was learned and language modeling was not would sit in a discarded prompt, so the next version could not be checked against it.

\subsection{Case 2: from a prompted patient to \texorpdfstring{PATIENT-$\Psi$}{PATIENT-Psi} and CBT-Bench}\label{sec:cases:4}

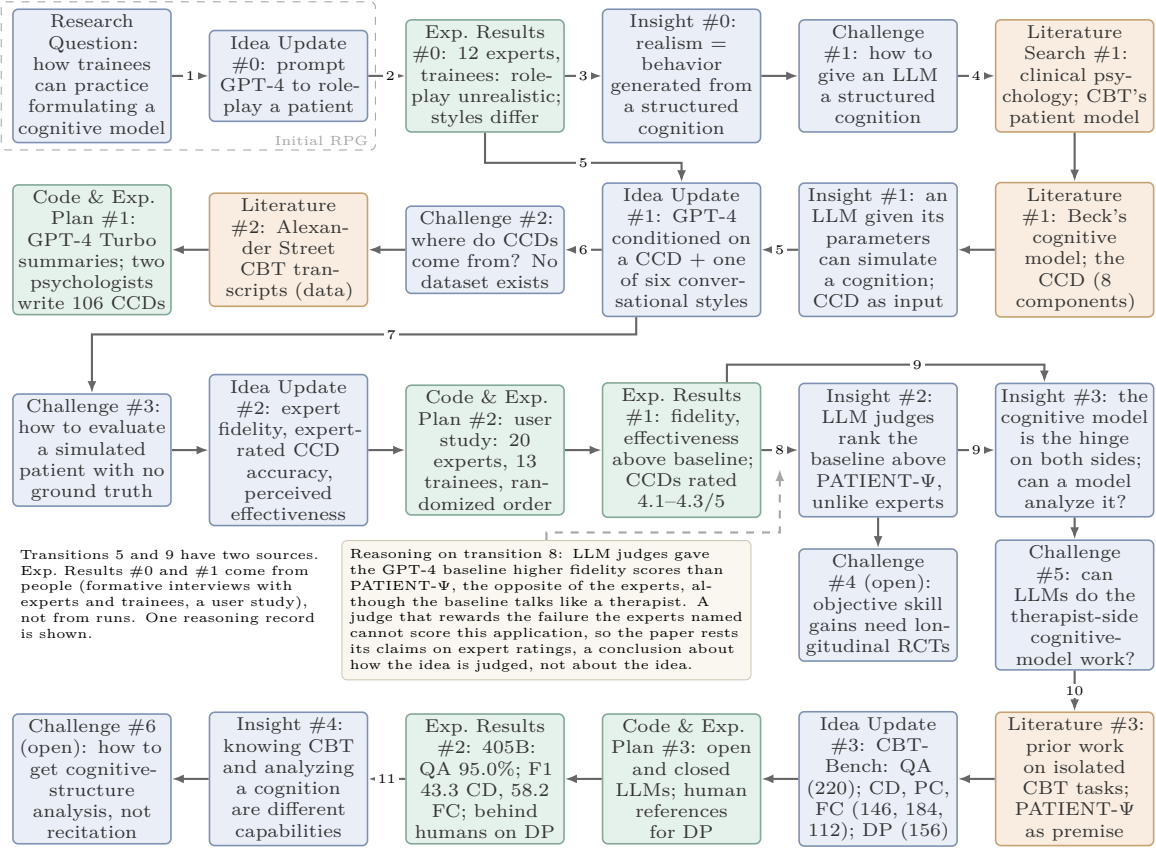
\begin{figure}[t]
\centering
\definecolor{cogF}{HTML}{E9EEF7}\definecolor{cogD}{HTML}{7F91B2}%
\definecolor{litF}{HTML}{F8EDE0}\definecolor{litD}{HTML}{C6A271}%
\definecolor{wldF}{HTML}{E5F1EB}\definecolor{wldD}{HTML}{82A896}%
\definecolor{rsF}{HTML}{FAF7F0}\definecolor{rsD}{HTML}{D1C9B6}%
\begin{tikzpicture}[
  font=\scriptsize,
  st/.style={line width=0.6pt, rounded corners=2pt, text=black!80, text width=1.95cm, minimum height=0.62cm, align=center, inner sep=2pt},
  ty/.style={draw, rounded corners=2pt, text width=1.9cm, minimum height=0.6cm, align=center, inner sep=2pt},
  cog/.style={fill=cogF, draw=cogD},
  lit/.style={fill=litF, draw=litD},
  wld/.style={fill=wldF, draw=wldD},
  e/.style={-{Latex[length=1.6mm]}, thick, draw=black!60},
  ed/.style={-{Latex[length=1.6mm]}, thick, dashed, draw=black!35},
  t/.style={font=\tiny, fill=white, inner sep=1pt},
  rs/.style={font=\tiny, align=left, text width=5.2cm, draw=rsD, fill=rsF, rounded corners=2pt, inner sep=3pt}
]
\node[st,cog] (rq0) at (1.20,-0.20) {Research Question: how trainees can practice formulating a cognitive model};
\node[st,cog] (i0) at (3.80,-0.20) {Idea Update \#0: prompt GPT-4 to role-play a patient};
\node[st,wld] (r0) at (6.40,-0.20) {Exp.\ Results \#0: 12 experts, trainees: role-play unrealistic; styles differ};
\node[st,cog] (in0) at (9.00,-0.20) {Insight \#0: realism = behavior generated from a structured cognition};
\node[st,cog] (c1) at (11.60,-0.20) {Challenge \#1: how to give an LLM a structured cognition};
\node[st,lit] (ls1) at (14.20,-0.20) {Literature Search \#1: clinical psychology; CBT's patient model};
\node[st,lit] (l1) at (14.20,-2.50) {Literature \#1: Beck's cognitive model; the CCD (8 components)};
\node[st,cog] (in1) at (11.60,-2.50) {Insight \#1: an LLM given its parameters can simulate a cognition; CCD as input};
\node[st,cog] (i1) at (9.00,-2.50) {Idea Update \#1: GPT-4 conditioned on a CCD + one of six conversational styles};
\node[st,cog] (c2) at (6.40,-2.50) {Challenge \#2: where do CCDs come from? No dataset exists};
\node[st,lit] (l2) at (3.80,-2.50) {Literature \#2: Alexander Street CBT transcripts (data)};
\node[st,wld] (ce1) at (1.20,-2.50) {Code \& Exp.\ Plan \#1: GPT-4 Turbo summaries; two psychologists write 106 CCDs};
\node[st,cog] (c3) at (1.20,-5.15) {Challenge \#3: how to evaluate a simulated patient with no ground truth};
\node[st,cog] (i2) at (3.80,-5.15) {Idea Update \#2: expert fidelity, expert-rated CCD accuracy, perceived effectiveness};
\node[st,wld] (ce2) at (6.40,-5.15) {Code \& Exp.\ Plan \#2: user study: 20 experts, 13 trainees, randomized order};
\node[st,wld] (r1) at (9.00,-5.15) {Exp.\ Results \#1: fidelity, effectiveness above baseline; CCDs rated 4.1--4.3/5};
\node[st,cog] (in2) at (11.60,-5.15) {Insight \#2: LLM judges rank the baseline above PATIENT-$\Psi$, unlike experts};
\node[st,cog] (in3) at (14.20,-5.15) {Insight \#3: the cognitive model is the hinge on both sides; can a model analyze it?};
\node[st,cog] (c4) at (11.60,-7.20) {Challenge \#4 (open): objective skill gains need longitudinal RCTs};
\node[st,cog] (c5) at (14.20,-7.20) {Challenge \#5: can LLMs do the therapist-side cognitive-model work?};
\node[st,lit] (l3) at (14.20,-9.50) {Literature \#3: prior work on isolated CBT tasks; PATIENT-$\Psi$ as premise};
\node[st,cog] (i3) at (11.60,-9.50) {Idea Update \#3: CBT-Bench: QA (220); CD, PC, FC (146, 184, 112); DP (156)};
\node[st,wld] (ce3) at (9.00,-9.50) {Code \& Exp.\ Plan \#3: open and closed LLMs; human references for DP};
\node[st,wld] (r2) at (6.40,-9.50) {Exp.\ Results \#2: 405B: QA 95.0\%; F1 43.3 CD, 58.2 FC; behind humans on DP};
\node[st,cog] (in4) at (3.80,-9.50) {Insight \#4: knowing CBT and analyzing a cognition are different capabilities};
\node[st,cog] (c6) at (1.20,-9.50) {Challenge \#6 (open): how to get cognitive-structure analysis, not recitation};
\draw[dashed, gray!70, rounded corners=2pt] (0.00,0.78) rectangle (4.95,-1.18);
\node[font=\tiny, text=gray!70, anchor=south east, inner sep=1pt] at (4.90,-1.16) {Initial RPG};
\draw[e] (rq0) -- node[t]{1} (i0);
\draw[e] (i0) -- node[t,pos=0.6]{2} (r0);
\draw[e] (r0) -- node[t]{3} (in0);
\draw[e] (in0) -- (c1);
\draw[e] (c1) -- node[t]{4} (ls1);
\draw[e] (ls1) -- (l1);
\draw[e] (l1) -- (in1);
\draw[e] (in1) -- node[t]{5} (i1);
\draw[e] (r0.south) -- (6.40,-1.35) -- node[t]{5} (9.00,-1.35) -- (i1.north);
\draw[e] (i1) -- node[t]{6} (c2);
\draw[e] (c2) -- (l2);
\draw[e] (l2) -- (ce1);
\draw[e] ([xshift=-0.6cm]i1.south) -- (8.40,-3.62) -- node[t,pos=0.45]{7} (1.20,-3.62) -- (c3.north);
\draw[e] (c3) -- (i2);
\draw[e] (i2) -- (ce2);
\draw[e] (ce2) -- (r1);
\draw[e] (r1) -- node[t]{8} (in2);
\draw[e] (in2) -- (c4);
\draw[e] (in2) -- node[t]{9} (in3);
\draw[e] ([xshift=0.6cm]r1.north) -- (9.60,-4.02) -- node[t,pos=0.6]{9} (13.80,-4.02) -- ([xshift=-0.4cm]in3.north);
\draw[e] (in3) -- (c5);
\draw[e] (c5) -- node[t]{10} (l3);
\draw[e] (l3) -- (i3);
\draw[e] (i3) -- (ce3);
\draw[e] (ce3) -- (r2);
\draw[e] (r2) -- node[t]{11} (in4);
\draw[e] (in4) -- (c6);
\node[rs, anchor=north west] (rs8) at (4.50,-6.34) {Reasoning on transition 8: LLM judges gave the GPT-4 baseline higher fidelity scores than PATIENT-$\Psi$, the opposite of the experts, although the baseline talks like a therapist. A judge that rewards the failure the experts named cannot score this application, so the paper rests its claims on expert ratings, a conclusion about how the idea is judged, not about the idea.};
\draw[ed] (rs8.north) -- ++(0,0.1) -| (10.30,-5.40);
\node[font=\tiny, anchor=north west, text width=3.9cm] at (0.13,-6.34) {Transitions 5 and 9 have two sources. Exp.\ Results \#0 and \#1 come from people (formative interviews with experts and trainees, a user study), not from runs. One reasoning record is shown.};
\end{tikzpicture}
\caption{Case 2, PATIENT-$\Psi$ and CBT-Bench, as an RPG. Four rounds, six states of evidence, two of them from people. Full account and listing in Appendix~\ref{app:cases:4}.}
\label{fig:case4}
\end{figure}

Case 2 is an application idea from one group \citep{wang2024patientpsi,zhang2024cbtbench}, annotated, or traced and confirmed, by its authors. Trainees in cognitive behavioral therapy have few realistic ways to practice formulating a patient's cognitive model before seeing real patients, and the initial idea was to prompt GPT-4 to role-play a patient. The first evidence was a formative interview study with twelve mental-health experts and trainees, who reported that peer role-play is unrealistic and that real patients differ in conversational style. The insight was that realism requires a structured cognition from which behavior is generated, and the challenge of giving an LLM one led to a literature search in clinical psychology, which returned Beck's cognitive conceptualization diagram as a parameterization an LLM can follow. Later rounds built the missing data, 106 diagrams written by two clinical psychologists from transcript summaries, and evaluated the simulated patient without ground truth. That evaluation found that LLM judges give the GPT-4 baseline higher fidelity scores than PATIENT-$\Psi$, the opposite of the experts, and the paper concludes that automatic evaluation with LLMs cannot assess the fidelity of a simulated patient, so its claims rest on the expert ratings. In the fourth round the group moved the object of the idea from the patient to the therapist. CBT-Bench tests whether a model can analyze a patient's cognitive structure, and its result separates knowing CBT from analyzing a patient's cognition. Llama-3.1-405B reaches 95.0\% on its CBT question set but an F1 of only 43.3 on cognitive-distortion classification and 58.2 on fine-grained core-belief classification (Figure~\ref{fig:case4}). Each change is confined to one state and tied to a specific result or paper. The conclusion about the judge, caused by a result about the judge, is the one a whole-idea regeneration would overwrite silently, and a judge failure of the same kind recurs in our pilot (\S\ref{sec:eval:judge}).

\subsection{Case 3: LDC, from fine-tuning an idea generator to controlling it}\label{sec:cases:5}

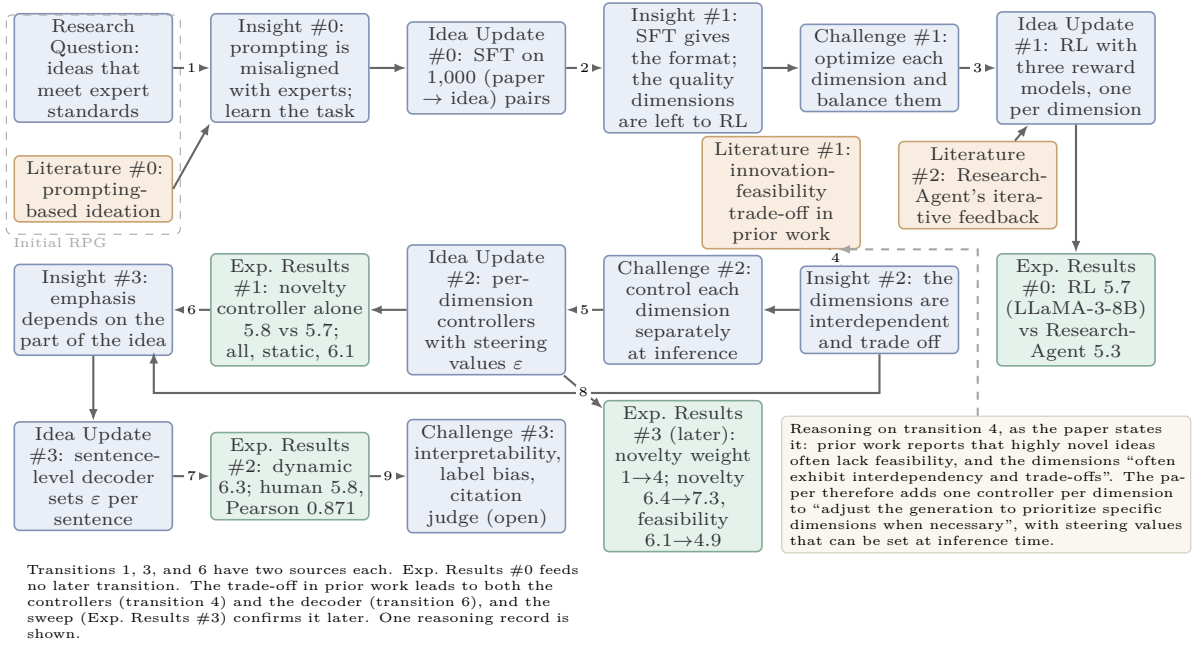
\begin{figure}[t]
\centering
\definecolor{cogF}{HTML}{E9EEF7}\definecolor{cogD}{HTML}{7F91B2}%
\definecolor{litF}{HTML}{F8EDE0}\definecolor{litD}{HTML}{C6A271}%
\definecolor{wldF}{HTML}{E5F1EB}\definecolor{wldD}{HTML}{82A896}%
\definecolor{rsF}{HTML}{FAF7F0}\definecolor{rsD}{HTML}{D1C9B6}%
\begin{tikzpicture}[
  font=\scriptsize,
  st/.style={line width=0.6pt, rounded corners=2pt, text=black!80, text width=1.95cm, minimum height=0.62cm, align=center, inner sep=2pt},
  ty/.style={draw, rounded corners=2pt, text width=1.9cm, minimum height=0.6cm, align=center, inner sep=2pt},
  cog/.style={fill=cogF, draw=cogD},
  lit/.style={fill=litF, draw=litD},
  wld/.style={fill=wldF, draw=wldD},
  txt/.style={draw, rounded corners=2pt, text width=2.2cm, minimum height=0.62cm, align=center, inner sep=2pt, fill=gray!8},
  pr/.style={draw, dashed, gray!70, rounded corners=2pt, text width=2.6cm, minimum height=0.62cm, align=center, inner sep=2pt, text=gray!70},
  e/.style={-{Latex[length=1.6mm]}, thick, draw=black!60},
  ed/.style={-{Latex[length=1.6mm]}, thick, dashed, draw=black!35},
  t/.style={font=\tiny, fill=white, inner sep=1pt},
  rs/.style={font=\tiny, align=left, text width=5.2cm, draw=rsD, fill=rsF, rounded corners=2pt, inner sep=3pt},
  grp/.style={font=\tiny, anchor=west, text width=1.7cm, align=left}
]
\node[st,cog] (rq0) at (1.2,-0.2)  {Research Question: ideas that meet expert standards};
\node[st,lit] (l0)  at (1.2,-1.8)  {Literature \#0: prompting-based ideation};
\node[st,cog] (in0) at (3.8,-0.2)  {Insight \#0: prompting is misaligned with experts; learn the task};
\node[st,cog] (i0)  at (6.4,-0.2)  {Idea Update \#0: SFT on 1,000 (paper $\to$ idea) pairs};
\node[st,cog] (in1) at (9.0,-0.2)  {Insight \#1: SFT gives the format; the quality dimensions are left to RL};
\node[st,cog] (c1)  at (11.6,-0.2) {Challenge \#1: optimize each dimension and balance them};
\node[st,cog] (i1)  at (14.2,-0.2) {Idea Update \#1: RL with three reward models, one per dimension};
\node[st,wld] (r0)  at (14.2,-3.4) {Exp.\ Results \#0: RL 5.7 (LLaMA-3-8B) vs ResearchAgent 5.3};
\node[st,lit] (l1)  at (10.3,-1.85) {Literature \#1: innovation-feasibility trade-off in prior work};
\node[st,lit] (l2)  at (12.9,-1.75) {Literature \#2: ResearchAgent's iterative feedback};
\node[st,cog] (in2) at (11.6,-3.4) {Insight \#2: the dimensions are interdependent and trade off};
\node[st,cog] (c2)  at (9.0,-3.4)  {Challenge \#2: control each dimension separately at inference};
\node[st,cog] (i2)  at (6.4,-3.4)  {Idea Update \#2: per-dimension controllers with steering values $\varepsilon$};
\node[st,wld] (r1)  at (3.8,-3.4)  {Exp.\ Results \#1: novelty controller alone 5.8 vs 5.7; all, static, 6.1};
\node[st,cog] (in3) at (1.2,-3.4)  {Insight \#3: emphasis depends on the part of the idea};
\node[st,cog] (i3)  at (1.2,-5.6)  {Idea Update \#3: sentence-level decoder sets $\varepsilon$ per sentence};
\node[st,wld] (r2)  at (3.8,-5.6)  {Exp.\ Results \#2: dynamic 6.3; human 5.8, Pearson 0.871};
\node[st,wld] (r3)  at (9.0,-5.6)  {Exp.\ Results \#3 (later): novelty weight 1$\to$4; novelty 6.4$\to$7.3, feasibility 6.1$\to$4.9};
\node[st,cog] (c3)  at (6.4,-5.6)  {Challenge \#3: interpretability, label bias, citation judge (open)};
\draw[dashed, draw=gray!70, rounded corners=2pt] (0.05,0.5) rectangle (2.35,-2.4);
\node[font=\tiny, text=gray!70, anchor=north west, inner sep=1pt] at (0.1,-2.4) {Initial RPG};
\draw[e] (rq0) -- node[t]{1} (in0);
\draw[e] (l0.east) -- (in0.south west);
\draw[e] (in0) -- (i0);
\draw[e] (i0) -- node[t]{2} (in1);
\draw[e] (in1) -- (c1);
\draw[e] (c1) -- node[t]{3} (i1);
\draw[e] (l2) -- (i1);
\draw[e] (i1) -- (r0);
\draw[e] (l1) -- node[t]{4} (in2);
\draw[e] (in2) -- (c2);
\draw[e] (c2) -- node[t]{5} (i2);
\draw[e] (i2) -- (r1);
\draw[e] (r1) -- node[t]{6} (in3);
\draw[e] (in3) -- (i3);
\draw[e] (in2.south) -- (11.6,-4.5) -- (2.0,-4.5) -- (2.0,-3.95);
\draw[e] (i3) -- node[t]{7} (r2);
\draw[e] (r2) -- node[t]{9} (c3);
\draw[e] (i2) -- node[t]{8} (r3);
\node[rs, anchor=north west] (rs4) at (10.3,-4.8) {Reasoning on transition 4, as the paper states it: prior work reports that highly novel ideas often lack feasibility, and the dimensions ``often exhibit interdependency and trade-offs''. The paper therefore adds one controller per dimension to ``adjust the generation to prioritize specific dimensions when necessary'', with steering values that can be set at inference time.};
\draw[ed] (12.9,-4.8) -- (12.9,-2.6) -- (11.05,-2.6);
\node[font=\tiny, anchor=north west, text width=7.3cm] at (0.2,-6.65) {Transitions 1, 3, and 6 have two sources each. Exp.\ Results \#0 feeds no later transition. The trade-off in prior work leads to both the controllers (transition 4) and the decoder (transition 6), and the sweep (Exp.\ Results \#3) confirms it later. One reasoning record is shown.};
\end{tikzpicture}
\caption{Case 3, LDC, as an RPG. Three rounds, seven states of evidence, three of them from the literature. The trade-off enters from prior work at transition 4 and leads to both the controllers and the decoder (transition 6); the sweep that confirms it (Exp.\ Results \#3) is a later analysis. Full account and listing in Appendix~\ref{app:cases:5}.}
\label{fig:case5}
\end{figure}

Case 3 is a method for research-idea generation, LDC \citep{li2024ldc}, based on version 3 of its paper and annotated, or traced and confirmed, by its authors. Prompting-based ideation agents use pre-trained models without task-specific learning, and the paper describes their ideas as ``misaligned with expert standards''. The initial idea is therefore to fine-tune a model on pairs of a paper and the idea extracted from it. Fine-tuning gives the task format, and the paper leaves the quality dimensions to a later stage, so the first update adds reinforcement learning with three reward models, for novelty, feasibility, and effectiveness, a feedback design taken from the iterative feedback of ResearchAgent \citep{baek2024researchagent}. The evidence for the next step comes from the literature, not from the paper's own results. The paper takes from prior work, including \citet{si2024novel}, an ``inevitable innovation-feasibility trade-off'', in which highly novel ideas often lack feasibility, and it reads the dimensions as interdependent. It therefore adds, on top of the per-dimension rewards, one controller per dimension whose weight can be set at inference time. The same trade-off motivated the last update. A fixed setting of the controllers raises novelty at the cost of feasibility across the whole idea, so a decoder sets them sentence by sentence, with novelty emphasized in the method and feasibility in the experiment plan (Figure~\ref{fig:case5}). A later sweep on the finished system confirms the trade-off. Raising the weight of the novelty controller from 1.0 to 4.0 lifts novelty from 6.4 to 7.3 and drops feasibility from 6.1 to 4.9. An agent that receives only the overall score after the first update, 5.7 for reinforcement learning on Llama-3-8B against 5.3 for ResearchAgent, would see a method that already wins, since the score hides the trade-off between novelty and feasibility. The two updates that mattered came from reading prior work as the statement that raising one dimension lowers another. That reasoning sits on the transitions from the literature state to the controllers and to the decoder, and a representation that keeps only the current idea has no place for it or for the paper it came from.

\subsection{Case 4: MLR-Copilot, from a prompted idea agent to a trained one}\label{sec:cases:mlr}

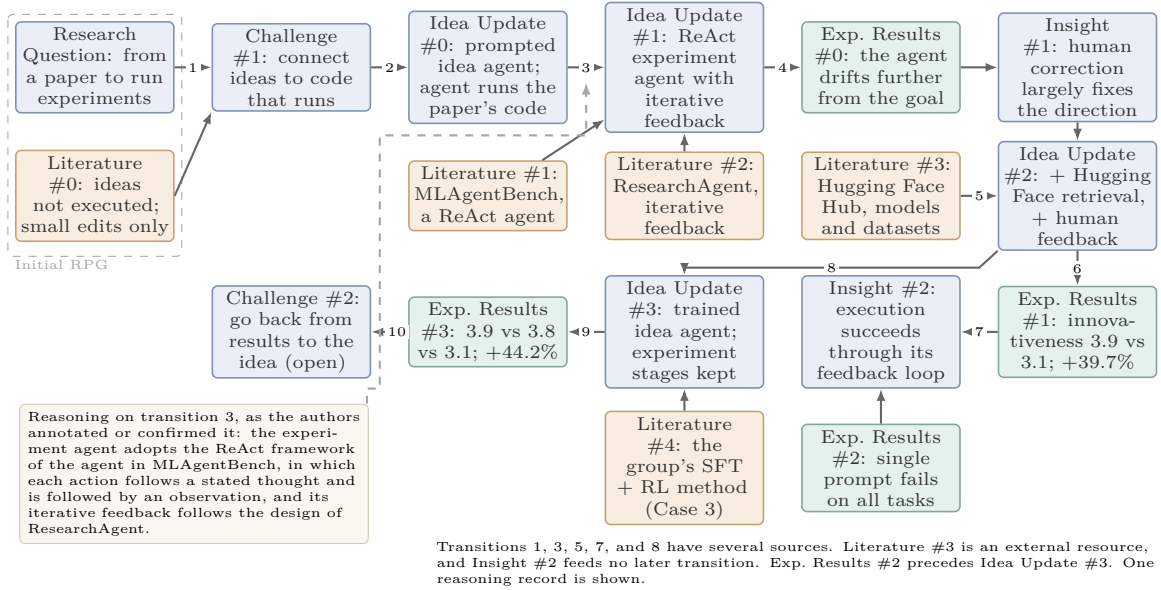
\begin{figure}[t]
\centering
\definecolor{cogF}{HTML}{E9EEF7}\definecolor{cogD}{HTML}{7F91B2}%
\definecolor{litF}{HTML}{F8EDE0}\definecolor{litD}{HTML}{C6A271}%
\definecolor{wldF}{HTML}{E5F1EB}\definecolor{wldD}{HTML}{82A896}%
\definecolor{rsF}{HTML}{FAF7F0}\definecolor{rsD}{HTML}{D1C9B6}%
\begin{tikzpicture}[
  font=\scriptsize,
  st/.style={line width=0.6pt, rounded corners=2pt, text=black!80, text width=1.95cm, minimum height=0.62cm, align=center, inner sep=2pt},
  ty/.style={draw, rounded corners=2pt, text width=1.9cm, minimum height=0.6cm, align=center, inner sep=2pt},
  cog/.style={fill=cogF, draw=cogD},
  lit/.style={fill=litF, draw=litD},
  wld/.style={fill=wldF, draw=wldD},
  txt/.style={draw, rounded corners=2pt, text width=2.2cm, minimum height=0.62cm, align=center, inner sep=2pt, fill=gray!8},
  pr/.style={draw, dashed, gray!70, rounded corners=2pt, text width=2.6cm, minimum height=0.62cm, align=center, inner sep=2pt, text=gray!70},
  e/.style={-{Latex[length=1.6mm]}, thick, draw=black!60},
  ed/.style={-{Latex[length=1.6mm]}, thick, dashed, draw=black!35},
  t/.style={font=\tiny, fill=white, inner sep=1pt},
  rs/.style={font=\tiny, align=left, text width=5.2cm, draw=rsD, fill=rsF, rounded corners=2pt, inner sep=3pt},
  grp/.style={font=\tiny, anchor=west, text width=1.7cm, align=left}
]
\node[st,cog] (rq0) at (1.2,-0.2)  {Research Question: from a paper to run experiments};
\node[st,lit] (l0)  at (1.2,-1.9)  {Literature \#0: ideas not executed; small edits only};
\node[st,cog] (c1)  at (3.8,-0.2)  {Challenge \#1: connect ideas to code that runs};
\node[st,cog] (i0)  at (6.4,-0.2)  {Idea Update \#0: prompted idea agent; agent runs the paper's code};
\node[st,cog] (i1)  at (9.0,-0.2)  {Idea Update \#1: ReAct experiment agent with iterative feedback};
\node[st,lit] (l1)  at (6.4,-1.9)  {Literature \#1: MLAgentBench, a ReAct agent};
\node[st,lit] (l2)  at (9.0,-1.9)  {Literature \#2: ResearchAgent, iterative feedback};
\node[st,wld] (r0)  at (11.6,-0.2) {Exp.\ Results \#0: the agent drifts further from the goal};
\node[st,cog] (in1) at (14.2,-0.2) {Insight \#1: human correction largely fixes the direction};
\node[st,lit] (l3)  at (11.6,-1.9) {Literature \#3: Hugging Face Hub, models and datasets};
\node[st,cog] (i2)  at (14.2,-1.9) {Idea Update \#2: + Hugging Face retrieval, + human feedback};
\node[st,wld] (r1)  at (14.2,-3.7) {Exp.\ Results \#1: innovativeness 3.9 vs 3.1; +39.7\%};
\node[st,cog] (in2) at (11.6,-3.7) {Insight \#2: execution succeeds through its feedback loop};
\node[st,wld] (r2)  at (11.6,-5.5) {Exp.\ Results \#2: single prompt fails on all tasks};
\node[st,cog] (i3)  at (9.0,-3.7)  {Idea Update \#3: trained idea agent; experiment stages kept};
\node[st,lit] (l4)  at (9.0,-5.5)  {Literature \#4: the group's SFT + RL method (Case 3)};
\node[st,wld] (r3)  at (6.4,-3.7)  {Exp.\ Results \#3: 3.9 vs 3.8 vs 3.1; +44.2\%};
\node[st,cog] (c2)  at (3.8,-3.7)  {Challenge \#2: go back from results to the idea (open)};
\draw[dashed, draw=gray!70, rounded corners=2pt] (0.05,0.5) rectangle (2.35,-2.7);
\node[font=\tiny, text=gray!70, anchor=north west, inner sep=1pt] at (0.1,-2.7) {Initial RPG};
\draw[e] (rq0) -- node[t]{1} (c1);
\draw[e] (l0.east) -- (c1.south west);
\draw[e] (c1) -- node[t]{2} (i0);
\draw[e] (i0) -- node[t]{3} (i1);
\draw[e] (l1) -- (i1);
\draw[e] (l2) -- (i1);
\draw[e] (i1) -- node[t]{4} (r0);
\draw[e] (r0) -- (in1);
\draw[e] (l3) -- node[t]{5} (i2);
\draw[e] (in1) -- (i2);
\draw[e] (i2) -- node[t]{6} (r1);
\draw[e] (r1) -- node[t]{7} (in2);
\draw[e] (r2) -- (in2);
\draw[e] (i2.south west) -- (12.9,-2.85) -- node[t]{8} (9.0,-2.85) -- (i3.north);
\draw[e] (l4) -- (i3);
\draw[e] (i3) -- node[t]{9} (r3);
\draw[e] (r3) -- node[t, pos=0.3]{10} (c2);
\node[rs, anchor=north west, text width=4.4cm] (rs3) at (0.2,-4.65) {Reasoning on transition 3, as the authors annotated or confirmed it: the experiment agent adopts the ReAct framework of the agent in MLAgentBench, in which each action follows a stated thought and is followed by an observation, and its iterative feedback follows the design of ResearchAgent.};
\draw[ed] (4.8,-4.65) -- (4.8,-4.45) -- (4.97,-4.45) -- (4.97,-1.1) -- (7.7,-1.1) -- (7.7,-0.4);
\node[font=\tiny, anchor=north west, text width=9.5cm] at (5.6,-6.35) {Transitions 1, 3, 5, 7, and 8 have several sources. Literature \#3 is an external resource, and Insight \#2 feeds no later transition. Exp.\ Results \#2 precedes Idea Update \#3. One reasoning record is shown.};
\end{tikzpicture}
\caption{Case 4, MLR-Copilot, as an RPG. Three rounds and an open challenge, nine states of evidence, five of them literature states. Literature \#3 is an external resource, and Literature \#4 is a later paper by the same group (Case 3). Full account and listing in Appendix~\ref{app:cases:mlr}.}
\label{fig:casemlr}
\end{figure}

Case 4 is a system for automated machine learning research, MLR-Copilot \citep{li2024mlrcopilot}, based on the first and third arXiv versions of its paper (2024 and 2025) and annotated, or traced and confirmed, by its authors. It starts from a gap between two kinds of prior system. Ideation systems produce ideas that are never executed, and systems that automate experiments make small edits such as tuning hyperparameters, without robust feedback. The initial idea connects the two. A prompted idea agent reads a paper and writes a hypothesis and an experiment plan, and an experiment agent adapts the paper's prototype code to the plan and runs it, with no retrieval of models or datasets and no channel for human feedback. The first idea update takes two components from two papers. The experiment agent is rebuilt on the ReAct framework used by the agent of MLAgentBench \citep{huang2024mlagentbench}, and its iterative feedback follows the design of ResearchAgent \citep{baek2024researchagent}. The second update has two sources of different kinds. The Hugging Face Hub enters as an external resource, through one researcher's idea that retrieving candidate models and datasets from it would match models to the task better, with no result behind it. A tool through which a person can be asked for help was added after the agent was seen to drift further from the goal over its iterations, a drift that human correction largely fixed. This gives the system of the first version. On five tasks its ideas score higher than those of a prompt-only baseline (innovativeness 3.9 against 3.1) and its experiments improve on the prototype by 39.7\% on average, while a single-prompt baseline without execution feedback fails on every task. The third version keeps the experiment stages and replaces the idea agent with a trained model, fine-tuned on extracted ideas and refined by reinforcement learning with reward models for novelty, feasibility, and effectiveness. This is the method of a later paper by the same group (Case 3), used without its inference-time control, and the authors give the reason for the change as the availability of that better ideation method. The paper ends with an open challenge, that its stages run largely in sequence and need transitions back from implementation to ideation, which is the loop that property 4 requires (Figure~\ref{fig:casemlr}). Every idea update draws on a literature state, two papers for the first, the Hugging Face Hub for the second, and the group's own later paper for the third, and only the second also rests on a result of the system's own runs. A representation that keeps only the current system description records the final system. It does not record which component came from which paper, nor why each was added.

\subsection{Case 5: on-policy distillation from one training example}\label{sec:cases:3}

\begin{figure}[t]
\centering
\definecolor{cogF}{HTML}{E9EEF7}\definecolor{cogD}{HTML}{7F91B2}%
\definecolor{litF}{HTML}{F8EDE0}\definecolor{litD}{HTML}{C6A271}%
\definecolor{wldF}{HTML}{E5F1EB}\definecolor{wldD}{HTML}{82A896}%
\definecolor{rsF}{HTML}{FAF7F0}\definecolor{rsD}{HTML}{D1C9B6}%
\begin{tikzpicture}[
  font=\scriptsize,
  st/.style={line width=0.6pt, rounded corners=2pt, text=black!80, text width=1.95cm, minimum height=0.62cm, align=center, inner sep=2pt},
  ty/.style={draw, rounded corners=2pt, text width=1.9cm, minimum height=0.6cm, align=center, inner sep=2pt},
  cog/.style={fill=cogF, draw=cogD},
  lit/.style={fill=litF, draw=litD},
  wld/.style={fill=wldF, draw=wldD},
  e/.style={-{Latex[length=1.6mm]}, thick, draw=black!60},
  ed/.style={-{Latex[length=1.6mm]}, thick, dashed, draw=black!35},
  t/.style={font=\tiny, fill=white, inner sep=1pt},
  rs/.style={font=\tiny, align=left, text width=5.2cm, draw=rsD, fill=rsF, rounded corners=2pt, inner sep=3pt},
  grp/.style={font=\tiny, anchor=west, text width=1.7cm, align=left}
]
\node[st,cog] (rq0) at (1.2,-0.2)  {Research Question: how much data does OPD need, and why};
\node[st,lit] (l0)  at (3.8,-0.2)  {Literature \#0: one-shot RLVR as a controlled experiment; Part I of OPD};
\node[st,cog] (i0)  at (6.4,-0.2)  {Idea Update \#0: bring the one-shot lens to OPD: train on one query};
\node[st,wld] (ce0) at (9.0,-0.2)  {Code \& Exp.\ Plan \#0: one query vs 17k; 4 domains $\times$ 3 families};
\node[st,wld] (r0)  at (11.6,-0.2) {Exp.\ Results \#0: 68.5 vs 69.8 at step 300; 87\% of full-data gain};
\node[st,cog] (in0) at (14.2,-0.2) {Insight \#0: rollouts of one query reach 71.5\% of state clusters};
\node[st,cog] (c1)  at (14.2,-2.6) {Challenge \#1: query content, or only the states its rollouts generate?};
\node[st,wld] (ce1) at (11.6,-2.6) {Code \& Exp.\ Plan \#1: content-free prompts; 1--64 rollouts; 1--16 queries};
\node[st,wld] (r1)  at (9.0,-2.6)  {Exp.\ Results \#1: templates nearly match; 16 queries reach 98.9\% coverage};
\node[st,cog] (in1) at (6.4,-2.6)  {Insight \#1: inputs are state generators, not task instructions};
\node[st,cog] (c2)  at (3.8,-2.6)  {Challenge \#2: why does absorption still take hundreds of steps?};
\node[st,wld] (ce2) at (1.2,-2.6)  {Code \& Exp.\ Plan \#2: absorption rate on 1 to 17k queries; frozen states (64 trajectories)};
\node[st,wld] (r2)  at (1.2,-5.0)  {Exp.\ Results \#2: absorption rate declines the same way; frozen run gains for 200 steps};
\node[st,cog] (in2) at (3.8,-5.0)  {Insight \#2: the bottleneck is absorption rate, not supervision quantity};
\node[st,cog] (i1)  at (6.4,-5.0)  {Idea Update \#1: ``data-overfed, algorithm-starved''; select queries by the states they induce};
\draw[dashed, gray!70, rounded corners=2pt] (0.1,0.75) rectangle (7.5,-1.15);
\node[font=\tiny, text=gray!70, anchor=north west, inner sep=1pt] at (0.1,-1.15) {Initial RPG};
\draw[draw=black!35, line width=0.6pt] (rq0) -- (l0);
\draw[e] (l0) -- node[t]{1} (i0);
\draw[e] (i0) -- node[t]{2} (ce0);
\draw[e] (ce0) -- (r0);
\draw[e] (r0) -- node[t]{3} (in0);
\draw[e] (in0) -- node[t]{4} (c1);
\draw[e] (c1) -- node[t]{5} (ce1);
\draw[e] (ce1) -- (r1);
\draw[e] (r1) -- (in1);
\draw[e] (in1) -- node[t]{6} (c2);
\draw[e] (c2) -- node[t]{7} (ce2);
\draw[e] (ce2) -- (r2);
\draw[e] (r2) -- (in2);
\draw[e] (in2) -- node[t]{8} (i1);
\draw[e] (in1) -- (i1);
\draw[e] (in0.east) -- ++(0.5,0) |- (8.0,-6.3) -| (i1.south);
\node[rs, anchor=north east] (rs5) at (15.175,-4.3) {Reasoning on transition 5: Insight \#0 attributes the gain to the states the rollouts reach, not to the query. If so, the query's content should be replaceable: content-free templates and off-domain prompts should drive OPD nearly as well, and more rollouts from one query should help. The mechanism, not the observation, predicts the ablation.};
\draw[ed] (rs5.north) -- ++(0,0.1) -| (12.9,-2.95);
\node[font=\tiny, anchor=north west, text width=5.0cm] at (0.225,-6.1) {Transition 8 has three sources. The dashed box is the initial RPG; Part I is recorded as a literature state of Part II. One reasoning record is shown.};
\end{tikzpicture}
\caption{Case 5, on-policy distillation from one training example, as an RPG. Three rounds, four states of evidence. Full account and listing in Appendix~\ref{app:cases:3}.}
\label{fig:case3}
\end{figure}
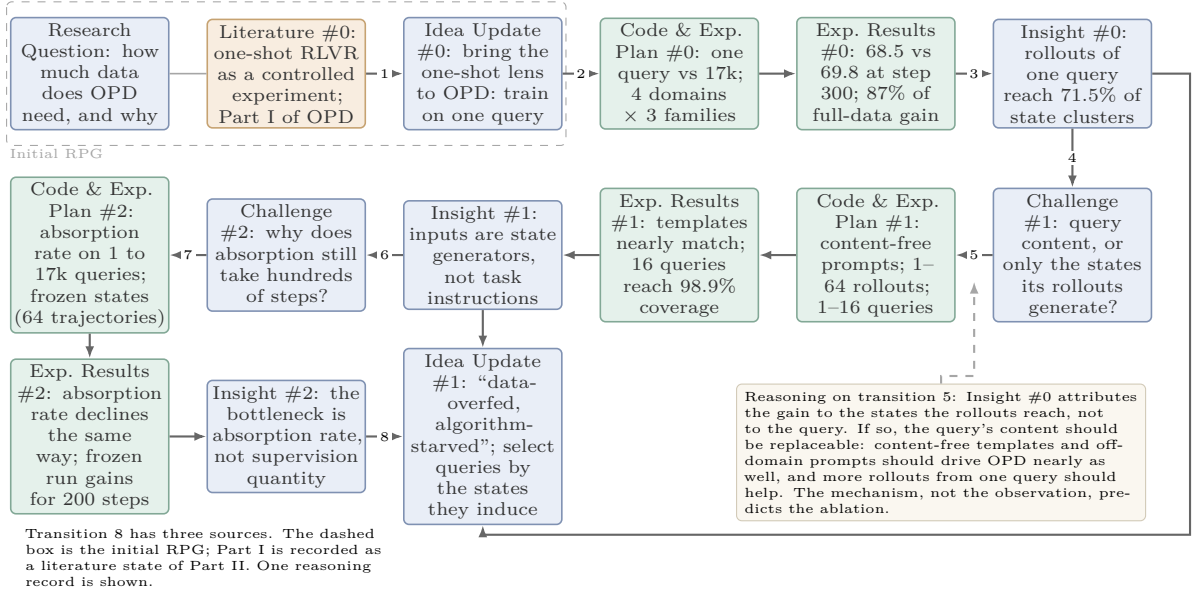

Case 5 is a finding carried across two papers by one group \citep{li2026rethinking1,fu2026rethinking2}, which share eight authors and a title. The second paper names one-shot RLVR \citep{wang2025oneshot}, not the first paper, as the source of its idea, and the first paper enters the graph of the second as a literature state. The initial idea transferred an experimental lens rather than a technique, training on-policy distillation (OPD) on a single query to isolate the role of data, and as an idea it had no mechanism, no prediction, and no boundary. The first result, that one-shot OPD keeps improving for hundreds of steps and recovers 69\% of the teacher--student gap, led to the insight that one query's rollouts already reach most of the states that full-data training reaches, 65.9\% of the state clusters by step 100. That insight raised a challenge that a text idea could not have raised, whether the query's content matters or only the states its rollouts generate, and an ablation answered it, with content-free templates and off-domain prompts nearly matching the full training set. A third round found that the share of the teacher--student distance closed by each update declines the same way at every data size. The final idea is a reframing, that OPD is ``data-overfed but algorithm-starved'' (Figure~\ref{fig:case3}). Its final idea is a re-scoped claim rather than a new idea, each challenge asks about one state, and the ablation exists because the insight predicted it. An agent that receives ``68.5 against 69.8'', one-shot against full-data accuracy at step 300, as a score would either stop or start an unrelated experiment.

\FloatBarrier
\subsection{What the cases establish}\label{sec:cases:summary}

The cases establish five things, which Appendix~\ref{app:cases:summary} tabulates case by case. First, the cycle is not specific to method ideas. The finding (Case 5) passes through it three times with almost only experimental evidence, the application (Case 2) with mostly human and literature evidence, the system (Case 4) with each update confined to one of its stages, and the evaluation idea in the appendix (Case 7) with a change to nothing but the judge. Second, the most valuable transitions are the ones where a \emph{challenge} redirects the search, such as ``how to obtain the language-modeling skill'' in Case 1, ``how to give an LLM a structured cognition'' in Case 2, and whether a small amount of cold-start data can improve reasoning and give a readable model in Case 8. This is why challenges are states of their own and not a sentence in a review. Third, idea updates are often not additions or regenerations. Examples are one slot of a structure substituted (Case 1), a claim re-scoped by two ablations (Case 5), the object of the idea moved from patient to therapist (Case 2), the emphasis set sentence by sentence rather than for the whole idea (Case 3), one stage of a system replaced by a method from another paper (Case 4), and the judge replaced (Case 7). An agent that can only regenerate a fuller idea cannot express any of these. Fourth, forks and abandoned attempts (Case 8, in the appendix) are sibling transitions out of one challenge with different reasons, which a representation holding only the current idea cannot hold at all. Fifth, in every case the decisive change is confined to one state and caused by a specific result or paper. Its reasoning is read from the authors' annotation, their narrative, or their paper, and outside the RPG it is kept, if at all, in prose apart from the state it explains. Agents that pass back a score, a review, or a summary and regenerate the whole idea cannot keep the state, the evidence, or the reasoning; typed idea graphs that no evidence enters have the states and not the evidence. Idea evolution needs both at once, which is what the comparison of \S\ref{sec:systems} finds missing in every existing representation.

\section{Existing Representations Compared with the RPG}
\label{sec:systems}
Every LLM-based research agent represents research ideation with some data structure, whether a proposal schema that is revised in place, a tree of experiments under a fixed hypothesis, or a graph of typed claims. This section groups these representations and compares each group with the RPG, using the eight state types of \S\ref{sec:rpg:states} and the five properties of \S\ref{sec:rpg:props} as the dimensions of comparison. The result is the same for every group. Each representation keeps some of the states and transitions of an RPG and collapses the rest into a text, a score, or a prompt that is discarded, so each can be viewed as an \emph{incomplete RPG}, and none of them represents the authentic research trajectory of an idea. The missing parts are not left blank. Each representation fills them with a device of the agent, a review text, a score, or a regeneration of the whole idea, that has no counterpart in a human trajectory, which is why these representations are artificial as well as incomplete. The comparison therefore has two aims, to locate, for each representation, which parts of the trajectory it lacks, and to show in what sense the RPG is complete and authentic where they are not.
The comparison covers thirty-five agents, thirty-two in the five group tables and three coding agents (AIDE, R\&D-Agent, and Gome) that appear in Table~\ref{tab:axes} but in no group table, and rests on their papers and on public code as of October 2026.

\subsection{How the representations are compared}\label{sec:systems:method}
Six questions are asked of every representation, and the answers come from the agent's paper and, where available, its code.
\begin{enumerate}[itemsep=1pt]
\item Which of the eight state types exist as distinct objects in its data structures, and which are folded into the idea text or into a prompt?
\item What evidence enters after the initial idea exists, whether nothing, a score, a review text, retrieved passages, a result label or summary, or typed states?
\item What does an update modify, the whole idea or named components of it?
\item Are earlier states and the evidence that produced them inputs to the next step, or are they only stored?
\item Where is the reasoning behind a change kept, bound to the change, detached from it as a review, or only in a chain of thought that is discarded?
\item Where can a person enter?
\end{enumerate}
The answers score the five properties. Question 1 scores typed states (P1), which are partial when some kinds of state, such as fields of the idea, retrieved papers, or results, are kept as separate records and absent when the idea and its evidence form one text; question 4 scores the temporal order that later steps read (P2), question 5 reasoning bound to the change (P3), questions 2 and 3 together evidence that enters and changes the idea (P4), and question 6 human intervention (P5). The answers to questions 2 and 3, what enters and what it revises, also place the representation in one of the five groups that follow, and the table of each group records them in its loop column.
 
Table~\ref{tab:axes} shows the answers state by state for twenty of the agents. Each column is one element of the RPG, the eight state types and the two annotations on a transition, and each row is an agent in order of first arXiv version. A cell is full when the element is a distinct object that a later step reads, partial when the information exists but is folded into a text, a prompt, or a score, or is stored and not read, and absent otherwise. The ideation agents type the idea, usually as a proposal schema \citep{si2024novel,baek2024researchagent}, and the literature, where a retriever returns records \citep{wang2024scimon,hu2024nova}. The coding and end-to-end research agents type the code and the results as nodes of an experiment tree or graph \citep{yamada2025aiscientist2,rdagent2025} and read their history. Insights and challenges are distinct states in almost none of them. ASI-Arch's per-node shortcoming analysis \citep{liu2025asiarch} is the only full insight state, and DeepScientist's findings memory \citep{weng2025deepscientist} is the nearest after it. Evolving Idea Graphs' evidence-need nodes \citep{dong2026eig} and the per-hypothesis challenges of R\&D-Agent and Gome \citep{rdagent2025,gome2026} are the only full challenge states. The research question is usually a fixed prompt rather than a state. The temporal index has analogs in ResearchTown's timesteps \citep{yu2025researchtown} and in the parent pointers of the tree-search agents, which the next step reads in the coding agents but rarely in the ideation agents. The reasoning behind a change is bound to the change in two rows only, in R\&D-Agent, where it concerns an experiment on a fixed task, and in Evolving Idea Graphs, which no evidence enters. No agent has all ten elements in full. Each kind of agent has built one part of the trajectory around a different object, the text for the ideation agents, the code for the coding agents, and a closed graph of claims for the typed-idea agents. Ten further agents are not scored because we have not examined their representations in detail. They are Deep Ideation \citep{zhao2025deep}, FlowPIE \citep{wang2026flowpie}, Idea2Story \citep{xu2026idea2story}, DeepInnovator \citep{fan2026deepinnovator}, \citet{ueda2025exploring}, Denario \citep{villaescusanavarro2025denario}, SciAgents \citep{ghafarollahi2025sciagents}, Sparks of Science \citep{oneill2025sparks}, GraphEval \citep{feng2025grapheval}, and Sparks \citep{ghafarollahi2025sparks}.
 
\begin{table}[t]
\centering\small
\caption{What twenty agents keep of an RPG, by element. \emph{v1} is the month of the first arXiv version, and the other columns are the eight state types of \S\ref{sec:rpg:states} and the two edge annotations of \S\ref{sec:rpg:edges}: \emph{Index}, the temporal index, read by the next step; \emph{Reasoning}, the reasoning process bound to the change it explains. \yes{} a distinct object that a later step reads; \half{} the information exists but is folded into a text, a prompt or a score, or is stored and not read; \no{} absent; ? not disclosed (Kosmos is closed source). Entries condense the group tables of \S\ref{sec:systems:A} to \S\ref{sec:systems:E}; AIDE, R\&D-Agent, and Gome appear in no group table, and their entries are read from their papers and code.}
\label{tab:axes}
\setlength{\tabcolsep}{4pt}
\begin{tabular}{llcccccccccc}
\toprule
\textbf{Agent} & \textbf{v1} & \textbf{RQ} & \textbf{I} & \textbf{IN} & \textbf{C} & \textbf{LS} & \textbf{L} & \textbf{CE} & \textbf{R} & \textbf{Index} & \textbf{Reasoning} \\
\midrule
SciMON & 2023-05 & \half & \half & \no & \no & \yes & \yes & \no & \no & \half & \half \\
ResearchAgent & 2024-04 & \half & \yes & \half & \no & \half & \yes & \no & \no & \half & \half \\
AI Scientist v1 & 2024-08 & \half & \yes & \half & \no & \half & \half & \yes & \yes & \half & \no \\
Si et al. & 2024-09 & \half & \yes & \no & \no & \half & \half & \no & \no & \no & \no \\
Chain of Ideas & 2024-10 & \half & \yes & \half & \no & \half & \yes & \half & \no & \half & \half \\
Nova & 2024-10 & \half & \yes & \half & \no & \yes & \yes & \no & \no & \half & \half \\
AIDE (coding) & 2025-02 & \half & \half & \half & \no & \no & \no & \yes & \yes & \yes & \half \\
Curie & 2025-02 & \yes & \half & \half & \no & \no & \no & \yes & \yes & \yes & \half \\
AI co-scientist & 2025-02 & \yes & \yes & \half & \no & \half & \half & \no & \no & \half & \half \\
AI Scientist v2 & 2025-04 & \half & \yes & \half & \no & \half & \half & \yes & \yes & \half & \half \\
IRIS & 2025-04 & \half & \yes & \half & \no & \half & \yes & \no & \no & \yes & \half \\
R\&D-Agent (coding) & 2025-05 & \half & \half & \half & \yes & \half & \half & \yes & \yes & \yes & \yes$^{a}$ \\
InternAgent (NovelSeek) & 2025-05 & \half & \half & \half & \no & \half & \half & \yes & \yes & \half & \half \\
ASI-Arch & 2025-07 & \half & \half & \yes & \half & \half & \yes & \yes & \yes & \yes & \half \\
DeepScientist & 2025-09 & \half & \yes & \half & \no & \half & \half & \yes & \yes & \yes & \half \\
Kosmos & 2025-11 & \yes & ? & ? & ? & \half & \half & \yes & \half & \half & ? \\
AlphaResearch & 2025-11 & \half & \half & \half & \no & \no & \no & \yes & \yes & \half & \no \\
Gome (coding) & 2026-03 & \half & \half & \half & \yes & \no & \no & \yes & \yes & \yes & \half \\
MOOSE-Star & 2026-03 & \half & \yes & \no & \no & \yes & \yes & \no & \no & \half & \half \\
Evolving Idea Graphs & 2026-05 & \yes & \yes & \half & \yes & \no & \half & \half & \no & \half & \yes \\
\midrule
\textbf{RPG (ours)} & & \yes & \yes & \yes & \yes & \yes & \yes & \yes & \yes & \yes & \yes \\
\bottomrule
\end{tabular}
\par\smallskip\raggedright\footnotesize $^{a}$ A reason and a targeted problem on most hypotheses (the class also defines observation and justification fields), but about a change to the code of an experiment on a fixed task, not to an idea state.
\end{table}

\subsection{A single idea text revised against itself}\label{sec:systems:A}

The first group (Table~\ref{tab:groupA}) represents the idea as a single text that is generated once or revised against the model's own critique. \citet{si2024novel} overgenerate thousands of seed ideas, deduplicate them, expand the remaining seeds into full proposals, and rank the proposals; the novelty filter retrieves papers but only discards ideas. LDC \citep{li2024ldc} generates a proposal from a target paper's context with controllable decoding; the Ramon Llull's Thinking Machine \citep{zhao2025llull} recombines theme, domain, and method elements; ResearchTown \citep{yu2025researchtown} simulates a community of proposers and reviewers; CycleResearcher \citep{weng2025cycleresearcher} trains a generator against a learned reviewer. ResearchAgent \citep{baek2024researchagent} is the clearest review loop. Reviewing agents score the problem, the method, and the experiment design, each on five criteria, and each part is refined iteratively, with the scores saturating after three rounds. Compared with the RPG, this representation holds a research question, a set of literature states read once, and a single idea update. The review loop adds a transition from one idea update to the next whose intermediate state, the review text, is untyped and is attached to a whole section of the idea, the problem, the method, or the experiment design, rather than to the part of it that should change, and no result or literature enters after the initial idea exists. What this representation records is a sequence of texts, and the steps between them are set by the prompt. This is the representation of Figure~\ref{fig:intro}, structures (1) and (2).

{\scriptsize
\begin{longtable}{>{\raggedright\arraybackslash}p{0.188\linewidth}>{\raggedright\arraybackslash}p{0.212\linewidth}>{\raggedright\arraybackslash}p{0.212\linewidth}>{\raggedright\arraybackslash}p{0.033\linewidth}>{\raggedright\arraybackslash}p{0.033\linewidth}>{\raggedright\arraybackslash}p{0.033\linewidth}>{\raggedright\arraybackslash}p{0.033\linewidth}>{\raggedright\arraybackslash}p{0.033\linewidth}}
\caption{Group A: a single idea text, generated once or revised against the model's own critique. Scores: \yes{} full, \half{} partial, \no{} absent; P1 to P5 are the properties of \S\ref{sec:rpg:props}; in column P4, lit, exp, and both name the evidence that enters after the initial idea and changes it, and \half{} marks evidence that changes only the code, the plan, or a hypothesis inside a fixed question; ? marks a score that cannot be given because the representation is not disclosed.}\label{tab:groupA}\\
\toprule
\textbf{Agent} & \textbf{Idea representation} & \textbf{Loop: feedback $\rightarrow$ what is revised} & \textbf{P1} & \textbf{P2} & \textbf{P3} & \textbf{P4} & \textbf{P5} \\
\midrule
\endfirsthead
\toprule \textbf{Agent} & \textbf{Idea representation} & \textbf{Loop: feedback $\rightarrow$ what is revised} & \textbf{P1} & \textbf{P2} & \textbf{P3} & \textbf{P4} & \textbf{P5} \\ \midrule \endhead
Si et al. \citep{si2024novel,si2025gap} & Proposal JSON: title, problem, motivation, method, experiment plan, test cases, fallback & None; novelty filter only discards; closed-loop feedback named as future work & \half{} & \no{} & \no{} & \no{} & \no{} \\
\midrule
LDC \citep{li2024ldc} & Text: problem, method, experiment plan, related work; input = a supporting paper & None; literature enters once before generation; novelty reward uses retrieval on the evaluation side only & \half{} & \no{} & \no{} & \no{} & \no{} \\
\midrule
Ramon Llull's Thinking Machine \citep{zhao2025llull} & Theme $\times$ domain $\times$ method combination expanded to a title and abstract & None & \half{} & \no{} & \no{} & \no{} & \half{} \\
\midrule
CycleResearcher \citep{weng2025cycleresearcher} & Full paper text & Training-time only: review scores train the next generator & \no{} & \no{} & \no{} & \no{} & \no{} \\
\midrule
ResearchAgent \citep{baek2024researchagent} & Problem, method, experiment, each with rationale & Yes: 15 reviewing agents, 5 criteria $\rightarrow$ whole-section regeneration, 3 rounds to saturation; no evidence enters the loop & \half{} & \half{} & \half{} & \no{} & \no{} \\
\midrule
ResearchTown \citep{yu2025researchtown} & Typed objects: insight, idea, proposal, review, and meta-review, with keys and timesteps; a rebuttal type is defined but its generation is disabled in the code & None at run time; the review $\rightarrow$ proposal transition is declared and never called & \half{} & \half{} & \half{} & \no{} & \no{} \\
\midrule
GoAI \citep{gao2025goai} & Literature graph with citation-semantic edges; idea = motivation, novelty, method, with reviewer scores & Review score $\rightarrow$ regeneration (mechanism unclear) & \half{} & \no{} & \no{} & \no{} & \no{} \\
\bottomrule
\end{longtable}}

\subsection{An idea text with literature retrieved for it}\label{sec:systems:B}

The second group (Table~\ref{tab:groupB}) retrieves literature for the current idea inside the loop. SciMON \citep{wang2024scimon} retrieves the closest prior work to a generated idea and asks for a ``significantly different'' one; Nova \citep{hu2024nova} plans a search around the current idea and generates child ideas from the results; MOOSE-Star \citep{yang2026moosestar} plans a motivation from the research background, searches a corpus hierarchically for the next inspiration paper, and adds one hypothesis increment per inspiration, with trained models for inspiration retrieval and hypothesis composition; Chain of Ideas \citep{li2025coi} orders retrieved papers into a chain before generating; InternAgent (NovelSeek) \citep{novelseek2025} adds literature from a survey agent while it evolves ideas, and VirSci \citep{su2025virsci} retrieves papers during multi-agent discussion; AI Scientist v2 \citep{yamada2025aiscientist2} retrieves literature that bears on the idea only during ideation, and its later searches only gather citations for the paper. Literature now enters after the initial idea, so the fourth property holds for literature. But no challenge directs the search, the query is derived from the whole idea or from the research goal rather than from an identified gap in the idea, and the update regenerates the idea rather than editing the component that the literature bears on. The insight that the literature yields, and the reasoning from that insight to the change, exist only in the prompt. Nova writes its full idea tree to disk with parent pointers and reads only the last step; the history exists and is not an input.

{\scriptsize
\begin{longtable}{>{\raggedright\arraybackslash}p{0.147\linewidth}>{\raggedright\arraybackslash}p{0.233\linewidth}>{\raggedright\arraybackslash}p{0.233\linewidth}>{\raggedright\arraybackslash}p{0.033\linewidth}>{\raggedright\arraybackslash}p{0.033\linewidth}>{\raggedright\arraybackslash}p{0.033\linewidth}>{\raggedright\arraybackslash}p{0.033\linewidth}>{\raggedright\arraybackslash}p{0.033\linewidth}}
\caption{Group B: an idea text with literature retrieved for it. Results, where experiments are run, do not revise the idea.}\label{tab:groupB}\\
\toprule
\textbf{Agent} & \textbf{Idea representation} & \textbf{Loop} & \textbf{P1} & \textbf{P2} & \textbf{P3} & \textbf{P4} & \textbf{P5} \\
\midrule
\endfirsthead
\toprule \textbf{Agent} & \textbf{Idea representation} & \textbf{Loop} & \textbf{P1} & \textbf{P2} & \textbf{P3} & \textbf{P4} & \textbf{P5} \\ \midrule \endhead
SciMON \citep{wang2024scimon} & One sentence with a record of input, prediction, nearest prior work, regeneration & Literature similarity $\rightarrow$ regeneration until no retrieved work is too close & \half{} & \half{} & \half{} & lit & \no{} \\
\midrule
MOOSE-Star \citep{yang2026moosestar} & Hypothesis built in increments from a research background, one increment (motivation, mechanism, methodology) per inspiration paper & Motivation $\rightarrow$ hierarchical search for the next inspiration $\rightarrow$ increment composed on the previous hypothesis, one step per inspiration; models trained on 108,717 papers decomposed by an LLM; no feedback loop & \half{} & \half{} & \half{} & lit & \no{} \\
\midrule
Chain of Ideas \citep{li2025coi} & Chain of papers in development order; idea = motivation, novelty, method & Novelty check $\rightarrow$ refine from chain; review + retrieval $\rightarrow$ experiment plan, 1 round & \half{} & \half{} & \half{} & lit & \no{} \\
\midrule
Nova \citep{hu2024nova} & Idea JSON with parent pointer & Search plan on current idea $\rightarrow$ child ideas, 3 rounds; tree stored, not read back & \half{} & \half{} & \half{} & lit & \no{} \\
\midrule
VirSci \citep{su2025virsci} & Idea JSON with title, experiment, and self-ratings $\rightarrow$ abstract & Team discussion $\rightarrow$ idea; similarity $\rightarrow$ abstract & \half{} & \half{} & \half{} & lit & \no{} \\
\midrule
AI Scientist v2 \citep{yamada2025aiscientist2} & Idea JSON frozen to a file; experiment tree nodes typed & Literature $\rightarrow$ proposal in ideation only; results $\rightarrow$ experiment tree, never the idea & \half{} & \half{} & \half{} & lit & \half{} \\
\midrule
InternAgent (NovelSeek) \citep{novelseek2025} & Text idea; evolution tree & Self-review + literature + human $\rightarrow$ idea tree; results $\rightarrow$ code & \half{} & \half{} & \half{} & lit & \yes{} \\
\midrule
AI co-scientist \citep{gottweis2025coscientist} & Text hypothesis, rationale, protocol, reviews, Elo & Review, debate, meta-review, web search $\rightarrow$ new hypotheses in a tournament; no lineage described (closed source); wet-lab outside the system & \half{} & \half{} & \half{} & lit & \yes{} \\
\midrule
IRIS \citep{garikaparthi2025iris} & MCTS state: brief, reward, review, retrieved items & Review, literature, user $\rightarrow$ child briefs & \half{} & \yes{} & \half{} & lit & \yes{} \\
\midrule
IdeaSynth \citep{pu2025ideasynth} & Canvas of facet nodes (problem, solution, evaluation, contribution) & Literature feedback + user $\rightarrow$ child facets & \half{} & \half{} & \half{} & lit & \yes{} \\
\midrule
Scideator \citep{radensky2024scideator} & Purpose, mechanism, evaluation facets & Novelty check $\rightarrow$ suggested facet swap $\rightarrow$ user regenerates & \half{} & \no{} & \half{} & lit & \yes{} \\
\bottomrule
\end{longtable}}

\subsection{A fixed idea with a tree of experiments}\label{sec:systems:C}

The third group (Table~\ref{tab:groupC}) runs experiments and feeds the results back, but into the implementation or the plan rather than into the idea. The AI Scientist \citep{lu2024aiscientist} re-plans experiments conditional on results up to five times while the idea is fixed in a file; AI Scientist v2 keeps one experiment tree per stage and starts each new stage from a copy of the previous stage's best node whose parent is set to null, so the lineage is cut at every stage transition, with the hypothesis frozen before the first tree starts; AIDE \citep{jiang2025aide}, Curie \citep{kon2025curie}, Jr.\ AI Scientist \citep{miyai2025jrai}, MLR-Copilot \citep{li2024mlrcopilot}, and Agent Laboratory \citep{schmidgall2025agentlab} revise code, plans, or a hypothesis inside a fixed experimental question from execution feedback. The representation in this group has typed code, experiment plans, and experimental results, and some of these agents read their history. This is the part of the trajectory that these agents represent best. But the cycle between plan and result runs many times under one idea update that never changes. The fourth property holds only in part, because the evidence changes the code, the plan, or a hypothesis inside a fixed question and not the idea, and the reasoning that is recorded, where it is recorded, explains why a piece of code was changed.

{\scriptsize
\begin{longtable}{>{\raggedright\arraybackslash}p{0.173\linewidth}>{\raggedright\arraybackslash}p{0.160\linewidth}>{\raggedright\arraybackslash}p{0.253\linewidth}>{\raggedright\arraybackslash}p{0.034\linewidth}>{\raggedright\arraybackslash}p{0.034\linewidth}>{\raggedright\arraybackslash}p{0.034\linewidth}>{\raggedright\arraybackslash}p{0.055\linewidth}>{\raggedright\arraybackslash}p{0.034\linewidth}}
\caption{Group C: a fixed idea with a tree of experiments; results change the code, the plan, or a hypothesis inside a fixed question, not the idea.}\label{tab:groupC}\\
\toprule
\textbf{Agent} & \textbf{Idea representation} & \textbf{Loop} & \textbf{P1} & \textbf{P2} & \textbf{P3} & \textbf{P4} & \textbf{P5} \\
\midrule
\endfirsthead
\toprule \textbf{Agent} & \textbf{Idea representation} & \textbf{Loop} & \textbf{P1} & \textbf{P2} & \textbf{P3} & \textbf{P4} & \textbf{P5} \\ \midrule \endhead
AI Scientist v1 \citep{lu2024aiscientist} & Idea JSON & Up to two self-reflection rounds per idea in the released pipeline; results $\rightarrow$ next experiment code and paper; score feedback to ideas described but not wired in the released code & \half{} & \half{} & \no{} & \half{} & \no{} \\
\midrule
AI-Researcher \citep{tang2025airesearcher} & Sectioned proposal & Advisor and judge $\rightarrow$ code & \half{} & \half{} & \half{} & \half{} & \no{} \\
\midrule
Agent Laboratory \citep{schmidgall2025agentlab} & Text topic and plan & LLM review + last results $\rightarrow$ new plan; human checkpoints & \half{} & \half{} & \half{} & \half{} & \yes{} \\
\midrule
Curie \citep{kon2025curie} & Experiment knowledge module with a DAG & Results $\rightarrow$ hypothesis within a fixed experimental question & \half{} & \yes{} & \half{} & \half{} & \no{} \\
\midrule
Jr.\ AI Scientist \citep{miyai2025jrai} & Idea JSON as AI Scientist v2 & Results $\rightarrow$ code; hypothesis fixed & \half{} & \half{} & \no{} & \half{} & \no{} \\
\midrule
MLR-Copilot \citep{li2024mlrcopilot} & Task, related work, hypothesis, plan & Results $\rightarrow$ researcher $\rightarrow$ hypothesis & \half{} & \half{} & \no{} & \half{} & \yes{} \\
\bottomrule
\end{longtable}}

\subsection{Ideas revised from experimental results}\label{sec:systems:D}

The fourth group (Table~\ref{tab:groupD}) closes the loop through experiments and into a new idea. Dolphin \citep{yuan2025dolphin} labels each idea's result as improvement, maintenance, or decline and lists the ideas that improved performance as effective in the next round's prompt; DeepScientist \citep{weng2025deepscientist} keeps a findings memory that the next hypothesis step retrieves from; ASI-Arch \citep{liu2025asiarch} stores a shortcoming analysis on every architecture node and retrieves cognitions to address it; AlphaResearch \citep{yu2025alpharesearch}, Kosmos \citep{kosmos2025}, freephdlabor \citep{li2025freephdlabor}, and Zochi \citep{intology2025zochi} pass results or summaries back to a new idea. R\&D-Agent \citep{rdagent2025}, a coding agent scored only in Table~\ref{tab:axes}, runs the same kind of loop over experiments on one fixed task, with several traces in parallel that are fused at the end. Results now enter and produce new idea updates, so the fourth property holds in full. What is missing is the middle of the trajectory. The transition from a result to a new idea carries a label, a score, or a summary rather than an insight and a challenge about a named component, and the new idea is regenerated in full, except in ASI-Arch, which edits the code of a parent architecture and writes a new motivation for the child. The reasoning behind the change, where it is recorded, is therefore not bound to the component that changed. ASI-Arch's shortcoming analysis and R\&D-Agent's hypothesis fields come closest to a reasoning process annotated on a transition.

{\scriptsize
\begin{longtable}{>{\raggedright\arraybackslash}p{0.125\linewidth}>{\raggedright\arraybackslash}p{0.234\linewidth}>{\raggedright\arraybackslash}p{0.215\linewidth}>{\raggedright\arraybackslash}p{0.034\linewidth}>{\raggedright\arraybackslash}p{0.034\linewidth}>{\raggedright\arraybackslash}p{0.034\linewidth}>{\raggedright\arraybackslash}p{0.070\linewidth}>{\raggedright\arraybackslash}p{0.034\linewidth}}
\caption{Group D: ideas revised from experimental results.}\label{tab:groupD}\\
\toprule
\textbf{Agent} & \textbf{Idea representation} & \textbf{Loop} & \textbf{P1} & \textbf{P2} & \textbf{P3} & \textbf{P4} & \textbf{P5} \\
\midrule
\endfirsthead
\toprule \textbf{Agent} & \textbf{Idea representation} & \textbf{Loop} & \textbf{P1} & \textbf{P2} & \textbf{P3} & \textbf{P4} & \textbf{P5} \\ \midrule \endhead
Dolphin \citep{yuan2025dolphin} & Idea JSON & Result label (improvement, maintenance, decline) $\rightarrow$ improved ideas marked effective in the next prompt, others used as a similarity filter; new ideas, not revisions; numbers and analysis do not reach the prompt & \half{} & \half{} & \no{} & exp & \no{} \\
\midrule
DeepScientist \citep{weng2025deepscientist} & Findings memory typed by lifecycle stage & Results + analysis records retrieved by the next hypothesis step & \half{} & \yes{} & \half{} & both & \half{} \\
\midrule
Kosmos \citep{kosmos2025} & Structured world model, format not public & Literature + data analysis summaries $\rightarrow$ world model $\rightarrow$ next tasks & ? & \half{} & ? & both & \no{} \\
\midrule
ASI-Arch \citep{liu2025asiarch} & Node: motivation, code, results, analysis, retrieved cognition; parent pointer & Results + shortcoming analysis + retrieved cognition $\rightarrow$ child architecture & \half{} & \yes{} & \half{} & both & \no{} \\
\midrule
AlphaResearch \citep{yu2025alpharesearch} & Ordered trajectory of idea, program, and result entries & Results $\rightarrow$ next idea; a simulated review score rejects ideas before they run & \half{} & \half{} & \no{} & exp & \no{} \\
\midrule
freephdlabor \citep{li2025freephdlabor} & Working idea JSON & Free-text results $\rightarrow$ manager $\rightarrow$ re-ideation; human can interject & \half{} & \half{} & \no{} & exp & \yes{} \\
\midrule
Zochi \citep{intology2025zochi} & Not public & Claimed only & ? & ? & ? & exp (claimed) & \yes{} \\
\bottomrule
\end{longtable}}

\subsection{Typed idea graphs that no evidence enters}\label{sec:systems:E}

The fifth group (Table~\ref{tab:groupE}) types the internal structure of the idea. Evolving Idea Graphs \citep{dong2026eig} has nine node types, among them problem, hypothesis, method, assumption, risk, and evidence need, records a rationale and a timestamp for every edit, although no later step reads this log at run time; ResearchTown (Table~\ref{tab:groupA}) keys every insight, idea, proposal, and review with a timestep, and Scideator and IdeaSynth (Table~\ref{tab:groupB}) decompose an idea into facets that a person recombines; these three type parts of the idea in the same way but are grouped by their loops.
Evolving Idea Graphs satisfies the first and third properties and the second in part, and these typed representations are the only ones in the comparison that can say which part of an idea changed. But no result or literature enters after the initial idea, since the literature is given in advance and no experiment is run. The graph is edited from the inside, and so the changes it records are not driven by evidence.

{\scriptsize
\begin{longtable}{>{\raggedright\arraybackslash}p{0.150\linewidth}>{\raggedright\arraybackslash}p{0.231\linewidth}>{\raggedright\arraybackslash}p{0.231\linewidth}>{\raggedright\arraybackslash}p{0.033\linewidth}>{\raggedright\arraybackslash}p{0.033\linewidth}>{\raggedright\arraybackslash}p{0.033\linewidth}>{\raggedright\arraybackslash}p{0.033\linewidth}>{\raggedright\arraybackslash}p{0.033\linewidth}}
\caption{Group E: typed idea graphs that no evidence enters.}\label{tab:groupE}\\
\toprule
\textbf{Agent} & \textbf{Idea representation} & \textbf{Loop} & \textbf{P1} & \textbf{P2} & \textbf{P3} & \textbf{P4} & \textbf{P5} \\
\midrule
\endfirsthead
\toprule \textbf{Agent} & \textbf{Idea representation} & \textbf{Loop} & \textbf{P1} & \textbf{P2} & \textbf{P3} & \textbf{P4} & \textbf{P5} \\ \midrule \endhead
Evolving Idea Graphs \citep{dong2026eig} & Typed nodes (problem, hypothesis, method, assumption, risk, evaluation plan, novelty claim, evidence need, repair) and typed edges; every operation logged with rationale and timestamp & Controller edit and commit steps inside the graph; literature given in advance; no retrieval, no experiments & \yes{} & \half{} & \yes{} & \no{} & \no{} \\
\bottomrule
\end{longtable}}

\subsection{Summary of the comparison}\label{sec:systems:gap}

Across the five groups, what flows around the existing loops is a score, a label, a review text, or a summary, and, except in ASI-Arch, the loop uses it to regenerate the whole idea. ResearchAgent's loop carries the reviewers' text and regenerates the reviewed part of the idea in full; Dolphin's carries a label of improvement, maintenance, or decline and an embedding filter that discards new ideas similar to earlier ones, and the numerical results never reach the prompt; AlphaResearch's carries the previous idea, program, and execution result, with a review score used only to reject ideas before they run; Kosmos's carries free-text summaries and freephdlabor's carries file references with short summaries. None of these representations can therefore answer, for a given change to an idea, which part of the idea changed, because of which piece of evidence, into what, and why. A recurring pattern in the code is that the trajectory is stored and not read. Nova writes its idea tree to disk and reads only the last step; AI Scientist v2 cuts the lineage of its experiment tree at every stage boundary and never revises the hypothesis; ResearchTown declares a transition from a rejection back to a new proposal that is never called; ResearchAgent keeps every round's versions and reviews in a history list that its code reads only to pick the best version and its paper does not mention. These agents already store the trajectory, but they do not give it to the next step to reason over.

None of the thirty-five has all five properties, and the two nearest fall short in different ways. R\&D-Agent is the nearest on reasoning. In its code a trace is a list of nodes of type (experiment, feedback), its second version organizes these nodes as a directed acyclic graph and selects a parent for each new node, and the default hypothesis object carries a hypothesis, the pipeline component it targets, the problem it addresses, and a reason, so a reason is recorded on most steps. But the node is an experiment on a fixed data-science task, the hypothesis is a free-text statement tagged with a pipeline component rather than a typed idea state, and the reason explains why a code or feature change was tried, so the graph is over candidate solutions to one task rather than over versions of an idea. Evolving Idea Graphs is the nearest on typing. It has the states and the rationales, but no evidence arrives after the initial idea. Behind these two are DeepScientist and ASI-Arch (real loops and a kept history, but lifecycle or engineering types and reasons linked by retrieval rather than annotated on transitions), IRIS (a version tree, reviews stored per node, and a person, but literature only), and the AI co-scientist (the strongest literature loop and human intervention, but free-text hypotheses with no lineage described).
Many representations with human intervention have no experiment loop, and most with an experiment loop are fully automatic. Agent Laboratory, MLR-Copilot, InternAgent, freephdlabor, and Zochi have both, but they keep the idea as text, or do not disclose it, and record human intervention and experimental evidence separately rather than on one graph.

The RPG is complete in the sense that the parts each group has built, the idea text of the first two, the plans and results of the third, the result-to-idea transition of the fourth, and the typed internal structure of the fifth, sit in one graph together with the two states that almost none of them has, the insight and the challenge, and with the temporal index and the reasoning process on every transition between them. Nothing in an RPG is new to the agents taken together; what is new is that the pieces form a single evolving record, so that the question above has an answer for every change. The RPG is authentic in the sense that the trajectories of \S\ref{sec:cases}, recorded from their authors, fit the representation without a new construct, including their re-readings, derivations, and hand-overs, whereas each representation in this section would keep only a fragment of them.

\section{Evaluating Improvement}\label{sec:eval}

This section argues that if an idea is a trajectory, an improvement to it must be evaluated on the trajectory and not only on the final text.

\subsection{From absolute quality to improvability}\label{sec:eval:absolute}\label{sec:eval:improvability}

None of the signals used to judge research measures the quality of an idea at the time the idea has to be judged. Awards, citations, and venue acceptance score a paper, that is, the idea together with its execution, writing, timing, and authors, and they arrive months or years later. Experts can judge an idea directly, but in the blind study of \citet{si2024novel} they agree on only 56.1\% of rankings, against 66.0\% and 71.9\% that the same authors report for reviewers of complete NeurIPS 2021 and ICLR 2024 submissions, and their judgment is expensive and concerns a description of the idea rather than what becomes of it. A reward model trained on any of these signals inherits its bias, and it becomes circular when the same group trains both the judge and the system being judged. We therefore do not know how to define an absolute score for the text of an idea, and the framework does not require one.

The framework replaces the question of how good an idea is with the question of how far the idea can still be improved from its current state, and at what cost. This depends on where the idea stands in its graph, namely which of its assumptions have been tested, which challenges remain open, what evidence it rests on, and what the next round of the cycle would cost and could change. Two ideas with the same text, one with a result and the insight that explains it and one without, therefore stand in different places, and a judge who reads only the text cannot tell them apart. Improvability is not a score assigned to a text either. What can be measured is which of two processes moved an idea further from the same starting point. Whether improvability can be measured without running the loop, and whether it predicts later outcomes, are open questions.

\subsection{Relative evaluation at three levels}\label{sec:eval:levels}

Improvement is measured by comparison. Two processes start from the same seed with the same model, budget, and tools, and a blind pairwise judge, who cannot see how either result was produced, decides which result is better. On a trajectory this comparison can be made at three levels, none of which assigns a score to a single text. For a single transition, the question is whether the new state is supported by the graph so far, for example whether an insight follows from the result it cites or whether a challenge concerns a component of the current idea. A person can check this in minutes. For a sequence of transitions, the question is whether the sequence reaches the idea update that the authors' own trajectory reached, in no more steps, which requires recorded reference trajectories. For a whole trajectory, the question is whether it shows the cognitive patterns of human trajectories, such as substituting one slot of a structure (Case 1) or sharpening a claim and then ablating what its mechanism predicts (Case 5), which requires a taxonomy of such patterns that does not yet exist. The comparison proposed next works on final versions and, through its hidden target, at the second level.

\subsection{A proposed comparison and its pilot}\label{sec:eval:pilot}

The comparison that would test the framework asks whether the RPG loop improves an idea more than whole-idea rewriting does under equal resources. This paper specifies the comparison and does not run it. Five conditions share one generator and one output-token budget, and those that retrieve share one retrieval tool. The conditions are a one-shot rewrite, a review loop, a literature loop, the RPG loop, and the RPG loop without outside information. In the RPG loop each step is an insight, a challenge, a search, or an idea update, and an idea update edits named fields of the idea and must cite a node created since the previous update. Each seed is an early version of an idea from a paper posted after the generator's training cutoff, and the authors' own next version is kept as a hidden target. A blind pairwise judge compares the final ideas on the five criteria of \citet{si2024novel}, and a second measure records how close each condition came to the step the authors actually took. Appendix~\ref{app:protocol} gives the full protocol.

A pilot on three seeds changed the protocol in two ways and is reported for that reason only (Table~\ref{tab:pilot} in Appendix~\ref{app:protocol}). On overall preference the RPG loop lost to the review loop and to the literature loop, three pairs to zero each, and beat its own ablation three to zero and the one-shot rewrite two to one. The logs point to two causes, the budget and the judge (\S\ref{sec:eval:judge}). At an equal number of calls the review and literature loops generated about 13,000 to 14,000 output tokens per run and the RPG loop about 5,000, because four of its six steps produce no idea text, so the protocol now fixes generated tokens rather than calls. Three seeds are too few to say anything about the framework, and we draw no conclusion about it from the pilot.

\subsection{The judge problem}\label{sec:eval:judge}

A judge that reads only the final text cannot tell a claim that was checked against the literature from one that was not, so it can reward an idea for hiding its prior work and penalize one for citing it. The known biases of language-model judges, toward position, length, and their own generations \citep{zheng2023judging,wang2024fair,panickssery2024llm}, are of the same kind, since each rewards a property of the text rather than of the idea. Three instances show the problem, one published and two from our own runs.

In the first instance, from Case 2, LLM-based evaluators gave the unconstrained GPT-4 patient higher fidelity scores than PATIENT-$\Psi$, the reverse of the experts' ratings, and the authors conclude that automatic evaluation with LLMs cannot assess the fidelity of a simulated patient \citep{wang2024patientpsi}.

In the second instance, from our pilot, the judge read a citation as a lack of originality. Its rationales repeatedly hold the RPG version's own citations against it, for example ``B's main hypothesis, the passive-adaptive gap, is already reported by The Distillation Game (2026), which B cites itself'', and it awarded novelty to the review loop, whose claims were bolder and, since the loop has no tools, never checked. The RPG loop cites its literature because each of its updates must name the node that motivated it.

The third instance, a blind study on the 26 seeds of Appendix~\ref{app:protocol} run after the pilot, shows that a text-only judge often cannot tell an executed idea from an untested one. For each seed, the authors' own next version, written into the fixed schema by the generator from the abstract of the real paper that followed, with its results removed, was paired with a one-shot rewrite of the seed by the same generator, written without tools and never executed. Every target postdates the generator's training cutoff. A text-only judge, a newer model than the generator, compared each pair on the five criteria in both orders, and a preference counts only when the two orders agree (Table~\ref{tab:judgestudy}). The judge preferred the authors' version on novelty (19 pairs to 3) and on excitement (21 to 4). On feasibility and effectiveness it preferred the rewrite, 16 to 8 and 13 to 10, although neither preference is significant with 26 pairs, and on overall quality it was close to even, 13 to 10 for the authors' version. In ten of the 26 pairs, therefore, it rated an untested rewrite above the version the authors carried out. The rewrites were also longer, 641 words against 504 on average. Because the judge is newer than the generator, it may have seen some target papers in training, but a judge that recognized a target would favor the authors' version, so this possible bias works against the finding.

Giving the judge retrieval did not remove the problem. A verified-novelty judge extracts each version's novelty claims, retrieves literature posted before the target for each claim, and decides with that evidence visible. It preferred the authors' version in 13 pairs and the rewrite in 9 (last row of Table~\ref{tab:judgestudy}), and of the pairs that both judges decided, the four on which they disagree all moved toward the rewrite, although with four such pairs the asymmetry is not significant on its own (exact McNemar $p = 0.125$). The rationales explain the shift. The authors' claims sit in a crowded 2025 to 2026 literature in which retrieval finds concurrent work, while several of the rewrite's claims are too general for retrieval to match and pass unchallenged. A verified judge must therefore record a claim that retrieval cannot match as unsupported rather than as unrefuted. The study measures the judge and says nothing about the RPG.

\begin{table}[t]
\centering\small
\caption{Blind judges on 26 pairs, each pairing the authors' own next version with a tool-free one-shot rewrite of the same seed. The first five rows are the text-only judge; the last row (``Novelty, verified'') is the verified-novelty judge, whose retrieval is restricted to work posted before the target. A preference counts only when both presentation orders agree; the remainder are inconsistent. $p$(rewrite) is the share of decisive pairs in which the judge preferred the rewrite, with a 95\% Wilson interval.}
\label{tab:judgestudy}
\begin{tabular}{lcccc}
\toprule
\textbf{Criterion} & \textbf{Authors' version} & \textbf{One-shot rewrite} & \textbf{Inconsistent} & $p$\textbf{(rewrite)} [95\% CI] \\
\midrule
Novelty & 19 & 3 & 4 & 0.14 [0.05, 0.33] \\
Excitement & 21 & 4 & 1 & 0.16 [0.06, 0.35] \\
Feasibility & 8 & 16 & 2 & 0.67 [0.47, 0.82] \\
Effectiveness & 10 & 13 & 3 & 0.57 [0.37, 0.74] \\
Overall & 13 & 10 & 3 & 0.43 [0.26, 0.63] \\
\midrule
Novelty, verified & 13 & 9 & 4 & 0.41 [0.23, 0.61] \\
\bottomrule
\end{tabular}
\end{table}

The three instances set three requirements for the judge of the proposed comparison, all fixed before any run. The judge has the same retrieval access as the systems it judges, it records a claim that retrieval cannot match as unsupported, and a subset judged by people calibrates it. The text-only judgment is reported beside the verified one. The first two requirements give the judge what the trajectory records and the final text omits, namely whether a claim was checked and against what evidence, and the third measures how far the model judge can be trusted.

\section{Actionable Insights}\label{sec:insights}

This section turns the findings of the cases, the comparison of agents, and the evaluation into five recommendations, on what a research agent should store, how its output should be judged, what it should learn, what data it should learn from, and how agents and people should share one record. Each recommendation is presented in the same order. A first paragraph states what to do and which finding of this paper motivates it, a second states what a system or the field gains by following it, and a list gives the concrete steps. Table~\ref{tab:recs} summarizes the five, which build on one another. Evaluation and learning at the level of transitions both need the trajectory to be stored, learning also needs annotated trajectories, and the system design combines the first four.

\begin{table}[t]
\centering\small
\caption{The five recommendations of this section, the problem each addresses, where this paper shows the problem, and what is gained by following it.}
\label{tab:recs}
\begin{tabular}{>{\raggedright\arraybackslash}p{0.17\linewidth}>{\raggedright\arraybackslash}p{0.25\linewidth}>{\raggedright\arraybackslash}p{0.2\linewidth}>{\raggedright\arraybackslash}p{0.27\linewidth}}
\toprule
\textbf{Recommendation} & \textbf{Problem addressed} & \textbf{Shown in} & \textbf{What is gained} \\
\midrule
Store the trajectory & Histories are written and not read; the loop passes a score, a label, a review, or a summary, and regenerates the whole idea & \S\ref{sec:systems:gap}; Case 1 & Every change traced to a component, the evidence that caused it, and a reason; failed ideas kept with the reason they failed \\
\midrule
Evaluate relatively, with a grounded judge & A text-only judge rewards unchecked claims and penalizes cited ones & Case 2; pilot and blind study (\S\ref{sec:eval:judge}) & Grounded ideas are no longer penalized; the effect of the representation is separated from the effect of compute \\
\midrule
Train on next-state prediction & Whole-idea training gives one signal per idea, and that signal comes from a biased judge & \S\ref{sec:eval:absolute}, \S\ref{sec:cases:summary}; MOOSE-Star & Supervision per step that a person can audit; no reward model over whole ideas \\
\midrule
Annotate update paths & Training corpora hold final ideas only & \S\ref{sec:bg:recover}; the eight cases & Training data for next-state prediction and reference trajectories for evaluation \\
\midrule
Record agents and people on one graph & Human input and experimental evidence are recorded apart from each other and from the idea & \S\ref{sec:systems:gap} & One record of the project; the human role defined per transition \\
\bottomrule
\end{tabular}
\end{table}

\paragraph{Store the trajectory, not the current text.} A system that improves ideas should store each idea as a graph of typed states connected by time-indexed edges that carry their reasons, and should give the next step this graph rather than the latest text. The comparison of existing agents found that much of this information is already produced and then discarded. Nova writes its idea tree to disk with parent pointers and reads only the last step, AI Scientist v2 cuts the lineage of its experiment tree at every stage boundary, and ResearchTown declares a transition from a rejection back to a new proposal that is never called. In the loops that do run, what passes from one round to the next is a score, a label, a review text, or a summary, and the next round regenerates the whole idea from it. Case 1 shows what this loses. An agent given the first result as a BLEU score could arrive at ``use GPT-2'', but not at the decision to keep the copy switch, which came from re-reading the pointer-generator after the insight that copying and language modeling are separate skills.

Storing the trajectory gives three things that a text cannot. First, each change to an idea can be traced to the component it changed, the evidence that caused it, and the stated reason, so a person reviewing an agent's output checks one reason against one piece of evidence instead of comparing two full versions. Second, an idea that failed is kept together with the reason it failed, so a later step can avoid proposing it again. Dolphin comes closest with an embedding filter of earlier ideas, which prevents an idea from being proposed again but does not keep the reason it failed. Third, the stored trajectory is the input that the other four recommendations require.
\begin{itemize}[itemsep=1pt]
\item \emph{Make idea updates localized edits.} An update names the components it changes, cites the insight or literature node that justifies each change, and keeps the other components verbatim. This can be added to any structured proposal schema, and the resulting edit log records which component changed because of which node. Evolving Idea Graphs already logs a rationale for every edit, but no evidence enters its graph after the initial idea, so none of its edits can cite a result or a paper.
\item \emph{Type what drives the loop.} A review, a result, or a retrieved paper becomes a node rather than prompt text, so that a later step can refer to it and check an insight against it. R\&D-Agent already records a reason with most hypotheses, which shows that a working agent can record reasons as it runs. What is missing is a reason about a change to the idea rather than about a change to the code.
\item \emph{Read the history.} The next step's input includes the earlier versions and the evidence that produced them, which is property 2. Several systems already store this information, so what has to change is what the next step reads.
\end{itemize}

\paragraph{Evaluate relatively, with a grounded judge.} An improvement process should be evaluated by comparing results produced from the same seed under the same budget, blind and pairwise, at one of the three levels of \S\ref{sec:eval:levels}, and never by giving a single idea a number. The reason is that the judges in common use can be wrong in a direction that works against grounded ideas. In the pilot the text-only judge preferred the unchecked claims of a tool-free loop to the cited claims of the RPG loop, in the blind study it rated an untested rewrite above the authors' executed version in ten of 26 pairs, and in Case 2 LLM evaluators reversed the experts' ranking of two simulated patients. A field that optimizes against such a judge selects for ideas that sound bold and cannot be checked, and penalizes the grounding that the RPG records.

A relative protocol with a grounded judge removes this incentive, so that an agent is no longer penalized for citing the literature behind its claims. Equal budgets measured in generated tokens separate the effect of the representation from the effect of compute. Reporting how close a process came to the authors' own next step, beside the blind verdict, adds a signal that does not depend on any judge's preferences.
\begin{itemize}[itemsep=1pt]
\item \emph{Give the judge retrieval.} The judge has the same retrieval access as the systems it judges, restricted to work that predates the version under judgment. It checks the novelty claims of both sides before comparing them and records a claim that retrieval cannot match as unsupported, since the blind study showed that such claims otherwise pass. The text-only judgment is reported beside it.
\item \emph{Equalize generated tokens, not calls.} A loop whose steps are insights, challenges, and searches produces little idea text per call, so a budget counted in calls penalizes it for the work the framework asks for. In the pilot the RPG loop generated about 5,000 tokens per run against about 13,000 to 14,000 for the review and literature loops at the same number of calls.
\item \emph{Calibrate against people.} A subset of pairs is also judged by people. Each of the three examples above is a disagreement between a model judge and people or evidence, and only human judgments show how large that disagreement is under a given protocol.
\item \emph{Use the authors' own next step as a reference where it exists.} A process is credited for reaching the update the authors reached, or a later one, and for changing the idea along the dimension the authors changed. This is reported beside the blind judgment rather than combined with it, since an idea better than the reference can score low against it, and a disagreement between the two signals is itself informative.
\end{itemize}

\paragraph{Train on next-state prediction.}\label{sec:insights:learning} An ideation agent should be trained to predict the next state of a trajectory. Given a partial RPG, it predicts the next transition and the state that transition produces, whether an insight, a challenge, a search, or an idea update, together with the reasoning on the transition. Training an agent to produce a complete idea gives one training signal per idea, and the only signal available for a complete idea is a judge or a reward model, whose problems \S\ref{sec:eval} describes. The cases show that the information lies in individual steps, since in every case the decisive change is confined to one state and caused by a specific result or paper. MOOSE-Star \citep{yang2026moosestar} already trains this way for two transitions, retrieving the next inspiration and composing the next hypothesis increment, and finds that the decomposition makes training tractable where generating the whole hypothesis at once is not. DeepInnovator \citep{fan2026deepinnovator} calls its objective ``next idea prediction'', but what it predicts at each step is a complete improved idea, scored by a reward model against the idea of the target paper.

Training per step has four benefits. A trajectory with one wrong turn still supervises its correct steps, so the same data yields more supervision. A person can check each target, a single transition, in minutes, so the training data can be audited. No reward model over whole ideas is needed, which avoids the circularity that arises when one group trains both the judge and the system being judged. Finally, an agent trained on trajectories that include abandoned branches, such as the process reward model and the Monte Carlo tree search that DeepSeek-R1 dropped for stated reasons (Case 8), can learn not to propose a step that the graph already records as failed, because the reason for the failure is recorded with it.
\begin{itemize}[itemsep=1pt]
\item \emph{Limit the agent to the cognitive transitions.} A retriever produces the literature state from the search state, and a coding agent or a person produces the plan and result states from the idea state. The ideation agent therefore owns the insight, the challenge, the search, and the idea update, which are exactly the transitions that existing systems compress into one prompt. It can be trained and judged on each of them separately.
\item \emph{Compare continuations, not final texts.} The corresponding reinforcement learning formulation is preference learning over two continuations of the same partial graph. It keeps the same locality, since a person or a grounded judge compares two next steps rather than two complete ideas.
\end{itemize}

\paragraph{Annotate update paths.}\label{sec:insights:data} We propose that the lead authors of published papers annotate the update paths of their own ideas. Each author types the states the project passed through, orders them, and writes the reason on each edge, starting from the initial idea rather than from the paper's narrative, which is the protocol used for Case 1. Such data is needed because next-state prediction and the second level of evaluation both require trajectories, while the corpora that train ideation systems hold only final ideas, in abstracts, proposals, and papers. The nearest exception is MOOSE-Star's corpus \citep{yang2026moosestar}, in which an LLM decomposes 108,717 published papers into a background, inspiration papers, and hypothesis increments. It holds the steps from the literature to the hypothesis but not the results, insights, and challenges that revised it. The reasons can come only from the authors. Of our eight cases four are annotated, or traced and confirmed, by their authors. In the other four the reasons are those the authors chose to publish.

An annotated corpus would provide both the training data for next-state prediction and the reference trajectories for evaluation at the second level. Checked against each project's code and logs, it would also measure how often authors misremember the order of their own steps and the evidence behind them.
\begin{itemize}[itemsep=1pt]
\item \emph{Validate against artifacts.} Earlier code, logs, drafts, and submission histories exist for most projects, and the annotated order and evidence are checked against them, with every disagreement recorded.
\item \emph{Annotate close to the work.} Recall of the reasoning behind a step fades with time. The best record is kept while the project proceeds, which makes the annotation an RPG recorded during the work rather than reconstructed after it.
\item \emph{Keep the schema minimal and shared.} Annotators use the eight state types and add finer types only when they need them, so that corpora from different groups can be combined and compared.
\item \emph{Add cheaper sources.} Three sources need less of an author's time. Trajectories can be traced from papers whose evolution is transparent and then confirmed by their authors, or recovered by retrospection from the final idea as in the seeds of \S\ref{sec:eval:pilot}. Result and analysis pairs extracted from papers train the transition from a result to an insight, which, to our knowledge, no ideation system currently learns. Community-level trajectories, in which each paper's related work is its incoming literature and its limitations are its outgoing challenges, need no annotator. Each source has a different bias, and the practical question is which mixture trains a next-state predictor that transfers to held-out recorded trajectories.
\end{itemize}

\paragraph{Record agents and people on one graph.}\label{sec:insights:design} A research system should be built from specialists, agents and people, that read and write one RPG, so that the question of whether a human is in the loop becomes the question of which transitions the human takes. Agent Laboratory, MLR-Copilot, InternAgent, freephdlabor, and Zochi have both human intervention and an experiment loop, but each keeps the idea as text, or does not disclose it, and records the two kinds of input separately, so a later reader cannot tell whether a change came from a person's comment or from a result. Because the RPG does not distinguish who produced a state (property 5 of \S\ref{sec:rpg:props}), one RPG can hold both.

One graph gives the project a single record, in which every change has an author, a time, and a reason, whoever made it. The role of the person is set per transition, so a group can give the agent the searches and keep the idea updates, or the reverse, without changing the representation. A project that reaches a paper has also produced an author-annotated trajectory for the data agenda above, at no extra cost. The same graph covers the research process beyond ideation, since the ideation agent writes the cognitive transitions, a retriever and an experiment runner write the literature, plan, and result states, and the graph records all of them. When the six questions of \S\ref{sec:systems:method} are asked of the system under design rather than of an existing one, their answers fix its states, its transitions, and the place of the person.
\begin{itemize}[itemsep=1pt]
\item \emph{Fix the states and who writes each.} The idea follows the group's proposal template (problem, key insight, method, experiment plan, expected results, assumptions and risks). Members write the research question and the initial idea, the agent writes insights, challenges, searches, and idea updates, a retriever writes literature, and the group's experiment runner writes plans and results from the current idea. Every node carries its author and time index.
\item \emph{Define one agent transition.} Given the whole graph, including pruned and abandoned branches, the agent emits one transition with its reasoning. It is an insight citing the result or literature nodes it reads, a challenge naming the component of the idea it concerns, a search whose query is recorded, or an idea update that lists the changed components with the nodes justifying each and keeps the rest verbatim. A round that brings in no new outside information cannot produce an update, which implements property 4.
\item \emph{Place the person.} Members can veto or edit any transition before the next one, redirect the work by adding a challenge, and add a result or a paper as a node. Because their operations are the same as the agent's, the graph remains the single record.
\item \emph{Keep alternatives as sibling updates.} If the group wants alternatives, the agent proposes several updates from the same insight, each with its reason, and the choice among them is itself a transition with a reason. A pruned branch therefore keeps the reason it was pruned, and a later turn does not regenerate it.
\item \emph{Evaluate transitions as they are made.} Members rate single transitions as they review them. When the project reaches a paper, its recorded graph becomes an author-annotated trajectory, and the agent's predicted updates can be scored against it at the second level.
\end{itemize}
A system built in this way can answer, for every change to the idea, which component changed, which node caused the change, what the component became, and why, which is the question that none of the existing representations can answer.

\section{Limitations}\label{sec:disc}

The claims of this paper are limited by the scope of its evidence, by how that evidence was obtained, and by questions the framework leaves open. We state each, and close with the results that would count against the framework.

\paragraph{Scope.} The cases and the systems are drawn from machine learning and natural language processing, where results are cheap to obtain and a round of the loop takes days. In the experimental sciences, results are expensive and infrequent, so a trajectory may be dominated by literature and insight states and by challenges that stay open for long periods. Even within machine learning the evidence can be a derivation rather than a run, as in the step from reward modeling to direct preference optimization \citep{rafailov2023dpo}, which the RPG records as a result state, and in theoretical fields this is the usual case. The state types do not depend on the field, but the shape of the loop and the relative weight of its entry points do, and only recorded trajectories from other fields can test this. The cases are also restricted to trajectories within one group of authors wherever the reason on an edge is to be checked, because only the authors can confirm why they made a change. Trajectories across groups, such as the line of on-policy distillation papers in which Cases 5 and 6 sit, can be checked only for what each paper took from the previous one. Finally, since a published paper presents the result rather than the path (\S\ref{sec:bg:recover}), the orderings in our cases may be tidier than the projects were.

\paragraph{Evidence.} The ideal trajectory is annotated by its authors while the research proceeds, with each state and each reason recorded as it happens, and none of our eight cases was collected in this way. Cases 1 to 4 were annotated by their authors after the work was done, or traced from papers that state their steps clearly and then confirmed by their authors. Cases 5 to 8 were traced from what their authors published in papers, a technical report, and a blog post, so the reasons on their edges are the ones the authors chose to state. Where these sources leave the order of states or the reason for a step unstated, we inferred it from the surrounding text, and such inferred steps are not marked in the figures or listings. Tracing a trajectory from papers that describe their evolution clearly and having their authors confirm it can enlarge the set of trajectories at little cost, but it remains retrospective, and annotation during the research remains the goal (\S\ref{sec:insights:data}). The comparison of systems rests on papers and public code as of October 2026. Closed systems are scored on what their papers state, and the paper on Kosmos does not disclose its representation. The controlled comparison of \S\ref{sec:eval:pilot} is specified and piloted but not run, so this paper offers no evidence that a process built on the RPG improves ideas more than whole-idea rewriting. The evidence of \S\ref{sec:cases} supports the representation, in that real trajectories fit it, and not any process built on it. The pilot is evidence about judges and budget units only, and the protocol uses literature as the only outside information, so the result loop that several recent systems implement is not compared. The input of the RPG loop is the linearized graph, which costs more input tokens than any baseline, and we report this cost without correcting for it. The blind study of \S\ref{sec:eval:judge} uses one judge model and 26 pairs, its preferences on feasibility and effectiveness are not significant, and the starting points of its 21 retro seeds are reconstructed by a model.

\paragraph{Open questions.} The eight state types are minimal in the sense of \S\ref{sec:rpg:states}, and the cases already suggest where more may be needed. Application ideas have people as sources of evidence (Case 2). Many idea updates are substitutions, re-scopings, or abandoned branches (Cases 1, 5, and 8), and the RPG records the difference in the reasoning on the transition rather than in the type of the state. Assumptions, mechanisms, and the boundary of a claim are inside the idea text and cannot yet be the target of a challenge. Conclusions about which judge can score an idea are recorded as insights and a change of judge as an idea update, although both alter the idea's criterion of success and may need a state of their own. The reasoning on transitions, which \S\ref{sec:eval:levels} refers to only informally as cognitive patterns, needs a taxonomy grounded in recorded trajectories. Two patterns can already be named, abstracting the structure of a method and mapping a new component onto one slot of it (Case 1), and transferring a clinical instrument as the parameterization a model follows (Case 2). Three further questions are open. Whether improvability (\S\ref{sec:eval:absolute}) can be measured without running the loop, whether it predicts the signals that arrive later, such as expert preference, execution outcomes, and citations, and whether searching over typed idea states improves on the search over idea texts that Nova and IRIS already show to help \citep{hu2024nova,garikaparthi2025iris} are all questions that the framework poses and does not answer.

\paragraph{What would count against the framework.} The claims of this paper can be refuted in three places. The first is the claim of \S\ref{sec:rpg} that the decisive change to an idea is localized to one state and that its reason is bound to the evidence that caused it (properties 3 and 4). If trajectories recorded as the work proceeds showed decisive changes spread across states, or reasons that cannot be bound to evidence, this claim would fail and the state types with it. The second is the claim of \S\ref{sec:eval:judge} that a judge reading only the final text cannot tell grounded versions from ungrounded ones, and that retrieval alone does not fix this. If a text-only judge, or one given retrieval, were shown to agree with experts on such pairs, the requirements we set for the judge would be unnecessary. The third is the design claim of \S\ref{sec:insights}. If, under the protocol of Appendix~\ref{app:protocol}, a loop built on the RPG at an equal output budget were preferred no more often than whole-idea rewriting by a judge with retrieval, and came no closer to the authors' real next step, the design claim would fail, although the descriptive claims of \S\ref{sec:cases} and \S\ref{sec:systems} would stand. We state these so that the proposed protocol can serve as a test of the framework.

\section{Conclusion}\label{sec:concl}

Existing research agents represent a research idea as text and its improvement as rewriting, and each keeps only a fragment of the trajectory through which ideas actually evolve. The authentic trajectories of eight real ideas show what this trajectory contains. Evidence enters as a result or a paper, an insight and a challenge are drawn from it, and the idea changes locally for reasons that only its authors can state. The Research Progress Graph is a minimal representation that keeps these states, their temporal order, and the reasoning on every transition. Compared with it, each of thirty-five existing representations is an incomplete RPG, and the two nearest fall short in different ways. We have argued that the graph, rather than the text, should be the unit of research ideation, that is, what an agent predicts, what it is trained on, and what it is judged by. The judge problem, documented in a published case and in our blind study, shows what is lost when the unit is the text. The same graph, with its literature, plan, and result states written by retrievers, experiment runners, and people, records the research process as a whole; ideation is the part of it this paper has modeled. The most immediate consequence is a data agenda. Authentic trajectories, recorded by their own authors as the work proceeds, are the corpus that research agents lack.

\bibliography{refs}
\bibliographystyle{tmlr}

\appendix
\section{The Cases in Full}\label{app:cases}

This appendix gives each case of \S\ref{sec:cases} in full. For every case except Case 1, whose description is in the body (\S\ref{sec:cases:1}), the full account comes first. It is followed by the listings from which the figure is drawn, one table of typed states and one table of transitions, with the reasoning annotation of each transition abbreviated to a phrase. Node identifiers (RQ, L, LS, I, IN, C, CE, R with an index) follow Table~\ref{tab:states}, and a letter suffix (I3a, I3b) marks the branches of a fork. An abandoned idea update is an I node marked ``abandoned'' and numbered X1, X2, joined by a dashed edge ($\dashrightarrow$) that carries the reason for abandoning it. Within a case, section, table and figure numbers in parentheses refer to the cited source, not to this paper.

\subsection{Case 1: few-shot table-to-text generation}\label{app:cases:1}

\emph{Source: annotated, or traced and confirmed, by its authors; \citet{chen2020fewshot}.}

The description is in \S\ref{sec:cases:1}.

\paragraph{Listings.}
{\scriptsize
\begin{longtable}{>{\raggedright\arraybackslash}p{0.042\linewidth}>{\raggedright\arraybackslash}p{0.050\linewidth}>{\raggedright\arraybackslash}p{0.748\linewidth}}
\toprule
\textbf{id} & \textbf{type} & \textbf{content (short)} \\
\midrule
\endfirsthead
\toprule \textbf{id} & \textbf{type} & \textbf{content (short)} \\ \midrule \endhead
RQ0 & RQ & How to do table-to-text NLG with only a few hundred examples \\
L0 & L & Pointer-generator network (See et al. 2017), built on pointer networks; template NLG baselines \\
I0 & I & Pointer-generator network trained on the few-shot set \\
CE0 & CE & Implement pointer-generator; train at 50/100/200/500 examples \\
R0 & R & Copies table values well; connective language is poor at all sizes \\
IN0 & IN & Two skills are needed: copy (easy from few shots) and language modeling (hard from few shots) \\
C1 & C & How to acquire language modeling with little in-domain data \\
LS1 & LS & Search: few-shot / pre-trained language modeling \\
L1 & L & GPT-2 and contemporaneous pre-trained LMs \\
IN1 & IN & The LM skill can come from pre-training instead of task data \\
C2 & C & How to combine a pre-trained LM with copying \\
IN2 & IN & (from re-reading L0) The pointer-generator is a switch between an LM and a copy distribution; the LM slot can be replaced \\
I1 & I & Pre-trained LM as decoder + copy mechanism + learned switch; explicit copy supervision on matched positions \\
CE1 & CE & Implement I1; same training sizes \\
R1 & R & Large gains, e.g., 36.1 BLEU at 200 examples on Humans against 21.3 for the pointer-generator variant (Base + switch) and 4.9 for the original baseline (Table 1); project concluded \\
\bottomrule
\end{longtable}}
\addtocounter{table}{-1}

{\scriptsize
\begin{longtable}{>{\raggedright\arraybackslash}p{0.041\linewidth}>{\raggedright\arraybackslash}p{0.177\linewidth}>{\raggedright\arraybackslash}p{0.622\linewidth}}
\toprule
\textbf{\#} & \textbf{edge} & \textbf{transition} \\
\midrule
\endfirsthead
\toprule \textbf{\#} & \textbf{edge} & \textbf{transition} \\ \midrule \endhead
1 & RQ0, L0 $\rightarrow$ I0 & Choose the standard architecture for copy-heavy generation \\
2 & I0 $\rightarrow$ CE0 & Implement the pointer-generator \\
3 & CE0 $\rightarrow$ R0 & Train and evaluate at 50 to 500 examples \\
4 & R0 $\rightarrow$ IN0 & Analyze errors; copying succeeded, fluency failed \\
5 & IN0 $\rightarrow$ C1 & Isolate the hard skill as the challenge \\
6 & C1 $\rightarrow$ LS1 & Search for ways to get language modeling cheaply \\
7 & LS1 $\rightarrow$ L1 & Find GPT-2 and contemporaneous pre-trained LMs \\
8 & L1 $\rightarrow$ IN1 & Pre-training supplies the LM skill \\
9 & IN1 $\rightarrow$ C2 & The challenge narrows to combination \\
10 & C2, L0 $\rightarrow$ IN2 & Re-read L0; abstract the pointer-generator to ``switch between LM and copy'' \\
11 & IN2 $\rightarrow$ I1 & Substitute the LM slot, keep the switch and the copy supervision \\
12 & I1 $\rightarrow$ CE1 & Implement I1 \\
13 & CE1 $\rightarrow$ R1 & Run at the same training sizes; stop \\
\bottomrule
\end{longtable}}
\addtocounter{table}{-1}

\subsection{Case 2: from a prompted patient to \texorpdfstring{PATIENT-$\Psi$}{PATIENT-Psi} and CBT-Bench}\label{app:cases:4}

\emph{Sources: \citet{wang2024patientpsi} for the first three rounds, with the order of the intermediate states annotated, or traced and confirmed, by its authors; \citet{zhang2024cbtbench} for the fourth.}

\paragraph{Full account.}
The research question came from a gap reported by practitioners. Their training did not prepare them for real patients, and trainees have few realistic ways to practice formulating a patient's cognitive model, the core skill of cognitive behavioral therapy, before seeing real patients. The initial idea is the simplest one, to prompt GPT-4 to role-play a patient with a given diagnosis. The first outside information was not an experiment but a formative interview study with twelve mental-health experts and trainees, and it played the role R plays in the other cases. The interviewees reported that peer role-play is unrealistic because it does not reflect actual therapy sessions, and that real patients differ in conversational style. The paper reports the therapist-like behavior of the GPT-4 baseline only later, from its user study (\S4.1), where experts described it as ``too forthcoming with responses'' and one called it ``almost like doing therapy with a therapist''. The insight is that realism is a property of the patient's \emph{cognition being structured}, not of the surface text; a trainee learns to infer beliefs from behavior, so the simulated behavior must be generated from beliefs. The paper states that integrating a patient's cognitive model with an LLM can achieve high fidelity (\S1), and the step from the interviews to this insight is annotated, or traced and confirmed, by the authors. The challenge is how to give an LLM a structured cognition.

That challenge sends the search into clinical psychology rather than NLP, and the literature returned is Beck's cognitive model and its clinical instrument, the cognitive conceptualization diagram (relevant history, core beliefs, intermediate beliefs, coping strategies, situations, automatic thoughts, emotions, behaviors). The insight is a transfer of a general kind. An LLM can simulate a cognitive process when given its parameters, so a CCD can be the conditioning input. The first idea update is GPT-4 conditioned on a CCD and instructed to let the diagram inform its answers without stating it. The same formative interviews contributed a second component, six conversational styles (plain, upset, verbose, reserved, tangent, pleasing), because the experts emphasized that real patients show different conversational styles in therapy.

The second round is about where CCDs come from, which is a challenge with no analog in the method cases, since there is no dataset of cognitive models. The outside information is clinical transcripts (Alexander Street, CBT-filtered); the implementation is GPT-4 Turbo summaries of transcripts, then two clinical psychologists writing 106 CCDs from those summaries with their own expertise. The third round is evaluation without ground truth. The group evaluated three dimensions, fidelity to real patients, accuracy to the cognitive model as rated by experts, and perceived training effectiveness, with 20 experts and 13 trainees against a GPT-4 baseline. The results are the fourth injection, and one of them is an insight the paper reports explicitly. LLM-based evaluators give the GPT-4 baseline higher fidelity scores than PATIENT-$\Psi$, the opposite of human experts, although the baseline conveys CBT knowledge ``in a manner resembling a therapist speaking directly and explicitly, rather than a real patient'' (\S5). From that finding the paper concludes that automatic evaluation with LLMs cannot assess the fidelity of a simulated patient, so its claims rest on the expert ratings. The paper leaves objective skill gains for future work, which it notes could be measured in longitudinal randomized controlled trials, since all its measures of training effectiveness are perceived improvements.

The fourth round is the group's next paper, CBT-Bench \citep{zhang2024cbtbench}, which shares five authors with PATIENT-$\Psi$, including its senior author. The reason for the step, as read from the two papers, has been annotated, or traced and confirmed, by the authors. Two results of PATIENT-$\Psi$ point the same way. The expert ratings measured how accurately each simulated patient followed its cognitive model, so the cognitive model is central to the whole evaluation; and the finding that LLM judges give the unstructured baseline higher fidelity scores than PATIENT-$\Psi$, although GPT-4 rates accuracy to the cognitive model much as the experts do (\S5), showed that the models judge the realism of a structured patient poorly. The insight is that the same cognitive model is needed on both sides of therapy, to simulate the patient and to assist the therapist. PATIENT-$\Psi$ used it to make a model behave like a patient, but whether a model can \emph{analyze} a patient's cognition, which is what assisting the therapist would require, had not been tested. CBT-Bench states the gap it fills as ``work on professional assistance in real therapy settings remain limited'', with prior work addressing isolated tasks such as cognitive-distortion classification while ``many other critical stages in CBT ... could potentially be enhanced'' (\S1), and it cites PATIENT-$\Psi$ for the premise that ``with clear modeling of the cognitive model of patients, LLMs could act more like real patients'' (\S2). The idea update is a three-level benchmark on the therapist side, built with the same clinical collaborators, which tests CBT knowledge (CBT-QA, 220 questions), cognitive-model understanding (cognitive-distortion, primary core-belief and fine-grained core-belief classification; 146, 184 and 112 examples) and therapeutic response generation (CBT-DP, 156 deliberate-practice exercises). The result is the sixth injection. Llama-3.1-405B reaches 95.0\% on CBT-QA against 75.9\% for Mistral-7B, but current models ``struggle with detecting fine-grained cognitive disorders'', with Llama-3.1-405B at an F1 of 43.3 on cognitive-distortion classification and 58.2 on fine-grained core-belief classification (Table 5), and, by the authors' account, lag behind human reference responses on deliberate practice (\S5.3; Table 6). The resulting insight is that knowing CBT and analyzing a patient's cognitive structure are different capabilities, and only the first is present. The challenge left open is how to get the second.

The trajectory has four rounds and six states of evidence. Two of the six are studies with people (the formative interviews and the user study) rather than papers or runs, which is what application ideas look like in an RPG, and the fourth round shows the group's idea moving from the patient to the therapist because of its own evaluation results.

\emph{What a whole-idea regeneration loop would have lost.} A review loop would receive ``not realistic'' from the interviewees and regenerate a better patient prompt. The actual changes were three localized steps, each attributed to one piece of evidence. The conditioning input of I0 was replaced by a CCD (transitions 4 and 5) because R0 found practice through role-play unrealistic and L1 supplied a parameterization an LLM can follow; a conversational-style component was added (transition 5) because the same experts named the styles real patients show; and the paper concluded that its claims must rest on expert ratings (transition 8) because R1 showed LLM judges give the baseline higher fidelity scores than PATIENT-$\Psi$. The third step is the one a regeneration loop would have overwritten silently. It is a conclusion about how the idea can be judged, caused by a result about the judge, and it is only recordable as a node (IN2) with a transition back to R1. The fourth round adds a fourth kind of edit. The idea's \emph{object} changed, from the patient to the therapist, and the reason is a pair of results about the evaluation (expert-rated accuracy to the cognitive model; LLM judges that misjudge the realism of a structured patient) that a regeneration loop would have consumed as a score and forgotten.

\paragraph{Listings.}
{\scriptsize
\begin{longtable}{>{\raggedright\arraybackslash}p{0.045\linewidth}>{\raggedright\arraybackslash}p{0.10\linewidth}>{\raggedright\arraybackslash}p{0.50\linewidth}>{\raggedright\arraybackslash}p{0.235\linewidth}}
\toprule
\textbf{id} & \textbf{type} & \textbf{content (short)} & \textbf{source} \\
\midrule
\endfirsthead
\toprule \textbf{id} & \textbf{type} & \textbf{content (short)} & \textbf{source} \\ \midrule \endhead
RQ0 & RQ & How can CBT trainees practice formulating a patient's cognitive model before seeing real patients & Wang et al. 2024c \\
I0 & I & Prompt GPT-4 to role-play a patient with a diagnosis & Wang et al. 2024c \\
R0 & R (formative interviews, 12 experts and trainees) & Training did not prepare them for real practice; peer role-play is unrealistic; real patients show different conversational styles & Wang et al. 2024c \S1, App.\ B \\
IN0 & IN & Realism = behavior generated from a structured cognition; trainees must infer beliefs from behavior & Wang et al. 2024c \\
C1 & C & How to give an LLM a structured cognition & Wang et al. 2024c \\
LS1 & LS & Search in clinical psychology: how CBT models a patient & Wang et al. 2024c \\
L1 & L & Beck's cognitive model; cognitive conceptualization diagram (8 components) & Wang et al. 2024c \\
IN1 & IN & An LLM can simulate a cognitive process when given its parameters; a CCD can be the conditioning input & Wang et al. 2024c \\
I1 & I & GPT-4 conditioned on a CCD + one of six conversational styles; answers informed by, but not stating, the diagram & Wang et al. 2024c \\
C2 & C & Where do CCDs come from? No dataset exists & Wang et al. 2024c \\
L2 & L (data) & Alexander Street CBT therapy transcripts & Wang et al. 2024c \\
CE1 & CE & GPT-4 Turbo summaries of transcripts $\rightarrow$ two clinical psychologists write 106 CCDs (PATIENT-$\Psi$-CM) & Wang et al. 2024c \\
C3 & C & How to evaluate a simulated patient with no ground-truth conversation & Wang et al. 2024c \\
I2 & I & Evaluation on three dimensions: fidelity rated by experts + accuracy to the CCD rated by experts + perceived effectiveness rated by experts and trainees, against the GPT-4 baseline; PATIENT-$\Psi$-TRAINER interface & Wang et al. 2024c \S3 \\
CE2 & CE & User study: 20 experts, 13 trainees, randomized order & Wang et al. 2024c \\
R1 & R & Fidelity and effectiveness significantly above baseline and above textbooks/role-play; CCD components rated 4.1 to 4.3 / 5 & Wang et al. 2024c \\
IN2 & IN & LLM judges rank the baseline above PATIENT-$\Psi$ on fidelity, opposite to experts: automatic evaluation is unreliable for this application & Wang et al. 2024c \\
C4 & C (open) & Objective skill gains need longitudinal RCTs & Wang et al. 2024c \\
IN3 & IN & The cognitive model is central on both sides: PATIENT-$\Psi$ used it to make a model behave like a patient; whether a model can analyze a patient's cognition (the therapist's task) is untested; LLM judges misjudged the realism of a structured patient (reason annotated, or traced and confirmed, by the authors) & Wang et al. 2024c \S4.2, \S5; Zhang et al. 2025 \S1 \\
C5 & C & Can LLMs do the therapist-side cognitive-model work, beyond isolated distortion classification? & Zhang et al. 2025 \S1 (quoted) \\
L3 & L & Prior work: isolated CBT tasks (distortion classification); ``work on professional assistance in real therapy settings remain limited''; PATIENT-$\Psi$ as premise & Zhang et al. 2025 \S1, \S2 \\
I3 & I & CBT-Bench: three levels: CBT-QA (220); cognitive-model understanding: CBT-CD, CBT-PC, CBT-FC (146, 184, 112); CBT-DP deliberate practice (156); built with the clinical collaborators & Zhang et al. 2025 \S3 \\
CE3 & CE & Evaluate open and closed LLMs on all levels; human references for CBT-DP & Zhang et al. 2025 \S4, \S5 \\
R2 & R & CBT-QA 95.0\% (Llama-3.1-405B) vs 75.9\% (Mistral-7B); Llama-3.1-405B F1 43.3 on CBT-CD and 58.2 on CBT-FC (75.3 on CBT-PC); models lag behind human references on deliberate practice by the authors' account & Zhang et al. 2025 Tables 5, 6, \S5.3 \\
IN4 & IN & Reciting CBT knowledge and analyzing a patient's cognitive structure are different capabilities; only the first is present & Zhang et al. 2025 abstract, \S5 \\
C6 & C (open) & How to get cognitive-structure analysis rather than recitation & Zhang et al. 2025 \S6 \\
\bottomrule
\end{longtable}}
\addtocounter{table}{-1}

{\scriptsize
\begin{longtable}{>{\raggedright\arraybackslash}p{0.035\linewidth}>{\raggedright\arraybackslash}p{0.25\linewidth}>{\raggedright\arraybackslash}p{0.595\linewidth}}
\toprule
\textbf{\#} & \textbf{edge} & \textbf{transition} \\
\midrule
\endfirsthead
\toprule \textbf{\#} & \textbf{edge} & \textbf{transition} \\ \midrule \endhead
1 & RQ0 $\rightarrow$ I0 & Default choice \\
2 & I0 $\rightarrow$ R0 & Formative interviews with experts and trainees instead of a run \\
3 & R0 $\rightarrow$ IN0 $\rightarrow$ C1 & Diagnose lack of realism as lack of structured cognition (reason annotated, or traced and confirmed, by the authors; the paper states that a cognitive model gives high fidelity, \S1) \\
4 & C1 $\rightarrow$ LS1 $\rightarrow$ L1 $\rightarrow$ IN1 & Search a different field; identify the CCD as the parameterization an LLM can simulate \\
5 & IN1, R0 $\rightarrow$ I1 & Combine CCD conditioning with the styles experts named \\
6 & I1 $\rightarrow$ C2 $\rightarrow$ L2 $\rightarrow$ CE1 & Build the missing data with clinicians \\
7 & I1 $\rightarrow$ C3 $\rightarrow$ I2 $\rightarrow$ CE2 $\rightarrow$ R1 & Design and run evaluation without ground truth \\
8 & R1 $\rightarrow$ IN2 $\rightarrow$ C4 & Learn that LLM evaluation fails; leave objective outcomes open \\
9 & R1, IN2 $\rightarrow$ IN3 $\rightarrow$ C5 & Read the evaluation results as a statement about the cognitive model on both sides; move the question to the therapist's side (reason annotated, or traced and confirmed, by the authors) \\
10 & C5 $\rightarrow$ L3 $\rightarrow$ I3 $\rightarrow$ CE3 $\rightarrow$ R2 & Survey what exists; build the therapist-side benchmark with the same clinicians; evaluate \\
11 & R2 $\rightarrow$ IN4 $\rightarrow$ C6 & Separate knowledge from analysis; leave analysis open \\
\bottomrule
\end{longtable}}
\addtocounter{table}{-1}

\subsection{Case 3: LDC, from fine-tuning an idea generator to controlling it dimension by dimension}\label{app:cases:5}

\emph{Source: \citet{li2024ldc}, in version 3 of the arXiv paper (2026); quotations, section numbers, and table numbers follow that version. The states are traced from the paper, and the states and reasons are annotated, or traced and confirmed, by its authors.}

\paragraph{Full account.}
The states and reasons below are taken from the published paper and from its authors, and the order of the intermediate states follows the paper's own presentation, which, like every published paper, may be tidier than the project was, as \S\ref{sec:bg:recover} discusses.
The research question is how to make a language model produce research ideas that meet the standards experts actually apply. The literature the idea starts from is the line of ideation systems built on pre-trained models, such as ResearchAgent (\S1, \S2). The paper's reading of it is the first insight. These systems ``primarily rely on pre-trained models without task-specific learning'' (\S1), so their ideas are ``misaligned with expert standards'' (abstract). The initial idea is therefore to \emph{learn} the task by fine-tuning an LLM on pairs of a paper and the research idea extracted from it, with 1,000 ICLR papers in the fine-tuning split (\S4.1) and the most supporting prior work chosen by prompting LLaMA with the paper and the citation counts of the candidates (\S3.2). The paper states what this stage is for. Supervised fine-tuning is meant to ``introduce the model with the general task format'' and to ``improve model training stability in RL'', and its training data ``does not need to achieve high scores in terms of the metrics'', which are left to reinforcement learning (\S3.2). The challenge, stated in the introduction, is that ``optimizing idea generation towards each of the key dimensions while achieving a balanced trade-off remains a critical yet unresolved question'' (\S1).

The first idea update answers the first half of that challenge with reinforcement learning and three reward models, for novelty, feasibility, and effectiveness, whose labels come from review scores or from prompting an LLM on the paper and its reviews (\S3.3). The authors confirm that this feedback design follows ResearchAgent \citep{baek2024researchagent}, which refines ideas iteratively with feedback from reviewing agents on several criteria. In LDC the feedback on each dimension is given by a reward model during training. With the Llama-3-8B backbone used for every later number in this case, the result is 5.7 overall against 5.3 for ResearchAgent (Table 1). The second update is motivated by the literature rather than by this result. The introduction takes from prior work, \citet{si2024novel} and one further study, the ``inevitable innovation-feasibility trade-off'', in which highly novel ideas often lack feasibility (\S1), and the method section gives the same reason for the controllers, since the dimensions ``often exhibit interdependency and trade-offs'' (\S3.4). The insight is that the dimensions are interdependent, so the generator needs a separate handle for each. The second idea update adds dimensional controllers on top of the per-dimension rewards, one per dimension, implemented as layer-wise multiplicative perturbations of the hidden states with a steering value each. The steering values are set to 1 during training and can be set per dimension at inference time (\S3.4). With the same backbone, the novelty controller alone scores 5.8 against 5.7 for reinforcement learning without controllers, the feasibility and effectiveness controllers alone score 5.3 and 5.7, and all three together reach 6.1 (Table 1).

The third round comes from the same trade-off. As the authors confirm, a fixed steering value raises one dimension at the cost of another across the whole idea, so the emphasis has to change within it. The paper states where it should change. Emphasis should be context-aware, ``such as prioritizing novelty in the method part and feasibility in the experiment planning'' (\S1), and its analysis of how the dimensions vary with sentence position supports this (\S5.2, Figure 4). The idea update is a sentence-level decoder, a small recurrent network that predicts the three steering values for each sentence from the text generated so far, trained against per-sentence targets (\S3.5). Dynamic control reaches 6.3 overall, significant against all baselines (Table 1), and human evaluation agrees with the automatic scores (5.8 overall, Table 2; Pearson 0.871 on the overall score, Table 3). The experts' comment names the trade-off again, since novelty steering produced ``more creative, though sometimes less practical, ideas'' (\S4.4). A later analysis of the finished system confirms the trade-off directly. Raising the weight of the novelty controller from 1.0 to 4.0 lifts novelty from 6.4 to 7.3 and drops feasibility from 6.1 to 4.9 (\S5.1, Table 6). The challenges the paper leaves open are the interpretability of the learned controllers, the bias that prompted reward labels may inherit, and whether citation prediction could serve as a judge (Limitations).

The trajectory has three rounds and seven states of evidence, three from the literature and four experimental. Every idea update is a change to one component of the generator, and each is justified by a reading of prior work, a reading of the task, or, for the decoder, the same trade-off together with an analysis of sentence positions. LDC optimizes a one-shot generator against three rewards, and its trade-off finding leaves open whether the unit of improvement should be the generated text or the path by which it is updated, which is the question this paper takes up.

\emph{What a whole-idea regeneration loop would have lost.} A loop that receives ``5.7 vs 5.3'' regenerates a better reward or a bigger model. The update that mattered came from a trade-off reported in prior work (L1), read as a property of the task (IN2). Raising one dimension lowers another, so the fix is not a better reward but a separate handle per dimension. The paper's own sweep (R3, Table 6) confirms the trade-off only later, on the finished controllers. The same trade-off then led the next update to localize the handles to parts of the idea. Both changes are to the \emph{architecture of the generator}, and the reason for the first sits on the edge from the literature to the controllers. In the paper that reason is a sentence in the introduction and one in the method section, separate from the sweep that later confirms it.

\paragraph{Listings.}
{\scriptsize
\begin{longtable}{>{\raggedright\arraybackslash}p{0.045\linewidth}>{\raggedright\arraybackslash}p{0.10\linewidth}>{\raggedright\arraybackslash}p{0.50\linewidth}>{\raggedright\arraybackslash}p{0.235\linewidth}}
\toprule
\textbf{id} & \textbf{type} & \textbf{content (short)} & \textbf{source} \\
\midrule
\endfirsthead
\toprule \textbf{id} & \textbf{type} & \textbf{content (short)} & \textbf{source} \\ \midrule \endhead
RQ0 & RQ & How to generate research ideas that meet expert standards of novelty, feasibility and effectiveness & Li et al. \S1 \\
L0 & L & Ideation systems built on pre-trained models without task-specific learning (e.g., ResearchAgent) & Li et al. \S1, \S2 \\
L2 & L & ResearchAgent: ideas refined iteratively with feedback from reviewing agents on several criteria & annotated, or traced and confirmed, by the authors \\
IN0 & IN & Prompting alone leaves outputs ``misaligned with expert standards''; the task must be learned & Li et al. abstract (quoted), \S1 \\
I0 & I & Supervised fine-tuning on 1,000 (paper $\rightarrow$ extracted idea) pairs; supporting work chosen by prompting with citation counts & Li et al. \S3.2, \S4.1 \\
IN1 & IN & SFT gives the task format and RL stability; the quality dimensions are left to RL & Li et al. \S1, \S3.2 (stated purpose of the SFT stage) \\
C1 & C & How to optimize toward each dimension and balance them (``critical yet unresolved'') & Li et al. \S1 (quoted) \\
I1 & I & RL with three reward models (novelty, feasibility, effectiveness); labels from review scores or prompted from paper and review content & Li et al. \S3.3 \\
R0 & R & SFT + RL 5.7 overall (LLaMA-3-8B) vs ResearchAgent 5.3 & Li et al. Table 1 \\
L1 & L & Prior work: the ``inevitable innovation-feasibility trade-off''; highly novel ideas often lack feasibility & Li et al. \S1 (quoted), citing Si et al. 2024 and Yang et al. 2024 \\
IN2 & IN & The dimensions ``often exhibit interdependency and trade-offs''; each needs its own handle & Li et al. \S1, \S3.4 (quoted) \\
C2 & C & How to control each dimension separately at inference time & Li et al. \S3.4 \\
I2 & I & Dimensional controllers on top of the per-dimension rewards: per-dimension layer-wise multiplicative perturbations with steering values $\varepsilon$ & Li et al. \S3.4 \\
R1 & R & Novelty controller alone 5.8 (feasibility 5.3, effectiveness 5.7) vs 5.7 without controllers; all controllers, static, 6.1 (LLaMA-3-8B) & Li et al. Table 1 \\
IN3 & IN & Emphasis should depend on the part of the idea: novelty in the method, feasibility in the experiment plan & Li et al. \S1 (quoted), \S5.2 \\
I3 & I & Sentence-level decoder: an RNN predicts ($\varepsilon_n$, $\varepsilon_f$, $\varepsilon_e$) per sentence from the generated context & Li et al. \S3.5 \\
R2 & R & Dynamic 6.3 overall (significant against all baselines); human eval 5.8 overall; Pearson 0.871 with automatic on the overall score; experts: more creative, ``sometimes less practical'' & Li et al. Tables 1--3, \S4.4 \\
R3 & R (later analysis) & Novelty controller weight 1.0 $\rightarrow$ 4.0: novelty 6.4 $\rightarrow$ 7.3, feasibility 6.1 $\rightarrow$ 4.9; confirms the trade-off of L1 & Li et al. \S5.1, Table 6 \\
C3 & C (open) & Controller interpretability; bias in prompted reward labels; citation prediction as a judge & Li et al. Limitations \\
\bottomrule
\end{longtable}}
\addtocounter{table}{-1}

{\scriptsize
\begin{longtable}{>{\raggedright\arraybackslash}p{0.035\linewidth}>{\raggedright\arraybackslash}p{0.25\linewidth}>{\raggedright\arraybackslash}p{0.595\linewidth}}
\toprule
\textbf{\#} & \textbf{edge} & \textbf{transition} \\
\midrule
\endfirsthead
\toprule \textbf{\#} & \textbf{edge} & \textbf{transition} \\ \midrule \endhead
1 & RQ0, L0 $\rightarrow$ IN0 $\rightarrow$ I0 & Read prompting-based ideation as unlearned; fine-tune \\
2 & I0 $\rightarrow$ IN1 $\rightarrow$ C1 & State what SFT does and does not optimize; pose the multi-dimension challenge \\
3 & C1, L2 $\rightarrow$ I1 $\rightarrow$ R0 & Add per-dimension reward models, with the feedback design taken from ResearchAgent (annotated, or traced and confirmed, by the authors); measure \\
4 & L1 $\rightarrow$ IN2 $\rightarrow$ C2 & Read the trade-off reported in prior work as a property of the task; ask for separate control (reason stated in \S1 and \S3.4) \\
5 & C2 $\rightarrow$ I2 $\rightarrow$ R1 & Add dimensional controllers; measure \\
6 & IN2, R1 $\rightarrow$ IN3 $\rightarrow$ I3 & A fixed setting trades one dimension against another across the whole idea, so localize emphasis to parts of the idea; add the sentence-level decoder (trade-off annotated, or traced and confirmed, by the authors; parts stated in \S1 and \S5.2) \\
7 & I3 $\rightarrow$ R2 & Measure with automatic and human judges \\
8 & I2 $\rightarrow$ R3 & Later analysis: sweep the novelty controller weight; the trade-off of L1 reappears \\
9 & R2 $\rightarrow$ C3 & Record open questions \\
\bottomrule
\end{longtable}}
\addtocounter{table}{-1}

\subsection{Case 4: MLR-Copilot, from a prompted idea agent to a trained one}\label{app:cases:mlr}

\emph{Source: \citet{li2024mlrcopilot}, versions 1 (August 2024) and 3 (November 2025) of the arXiv paper; section and table numbers follow the version named. The states are traced from the two versions, and the states and reasons are annotated, or traced and confirmed, by the authors.}

\paragraph{Full account.}
The research question is how to carry machine learning research from an existing paper to executed experiments automatically. The first version places the system between two kinds of prior work. Ideation systems produce ideas that are never executed, and systems that automate experiments make small edits such as tuning hyperparameters and lack robust feedback. The initial idea has two agents. An idea agent, a prompted LLM, reads a paper and recent related work and writes first a hypothesis and then an experiment plan, and an experiment agent adapts the paper's prototype code to the plan and runs it. This initial system retrieved no models or datasets and had no channel for human feedback.

The first idea update takes one component from each of two papers. The experiment agent is rebuilt on the ReAct framework, in which each action follows a stated thought and is followed by an observation, as in the research agent of MLAgentBench \citep{huang2024mlagentbench}, and the iterative feedback through which execution results refine the work follows the design of ResearchAgent \citep{baek2024researchagent}. The second update has two sources of different kinds. The Hugging Face Hub, a repository of models and datasets, enters as an external resource, through one researcher's idea that retrieving candidates from it would match models to the task better, and no result preceded it. The tool through which a person can be asked for help was added after the agent was seen to drift further from the goal over its iterations, and human correction largely fixed the direction. The result is the system of the first version. On five tasks, with eight trials each, its ideas score higher than those of a baseline prompted with the core paper alone, for example 3.9 against 3.1 on innovativeness as rated by three experts, and its experiments improve on the prototype by 39.7\% on average with GPT-4, with 40\% of runs improving by at least 10\%. A single-prompt baseline, reported in the third version but run before the idea agent changed, fails on every task because it cannot detect and correct execution errors.

The third version keeps the experiment stages and replaces the idea agent with a trained Llama model. It is first fine-tuned on ideas and plans extracted from 1,000 papers and then refined by reinforcement learning with reward models for novelty, feasibility, and effectiveness, trained on OpenReview feedback for 4,271 ICLR and NeurIPS papers from 2023 and 2024. The paper attributes the method to a later paper by the same group (Case 3) and uses it ``without steering at inference time''. The reason for the change, as the authors give it, is that this better ideation method had become available. The limitations section states the open challenge that the stages run largely in sequence and need transitions back from implementation to ideation.

\emph{What a whole-idea regeneration loop would have lost.} A loop that receives the third version's scores would record a better system. It would not record that two of the three idea updates took components from other papers (a benchmark's agent framework, another system's feedback design, and the group's own idea generator), that one component came from an external resource on a researcher's idea and another from a correction for an observed drift, or that the paper ends by asking for the loop from results back to the idea that the RPG's fourth property describes.

\paragraph{Listings.}
{\scriptsize
\begin{longtable}{>{\raggedright\arraybackslash}p{0.042\linewidth}>{\raggedright\arraybackslash}p{0.050\linewidth}>{\raggedright\arraybackslash}p{0.56\linewidth}>{\raggedright\arraybackslash}p{0.19\linewidth}}
\toprule
\textbf{id} & \textbf{type} & \textbf{content (short)} & \textbf{source} \\
\midrule
\endfirsthead
\toprule \textbf{id} & \textbf{type} & \textbf{content (short)} & \textbf{source} \\ \midrule \endhead
RQ0 & RQ & Carry ML research from an existing paper to executed experiments automatically & v1 \S1 \\
L0 & L & Ideation systems whose ideas are not executed; experiment automation limited to small edits, without robust feedback & v1 \S1; v3 \S1 \\
C1 & C & Connect ideas to code that runs & v1 \S1, \S2 \\
I0 & I & Prompted idea agent (paper + related work $\rightarrow$ hypothesis $\rightarrow$ plan); experiment agent adapts and runs the paper's prototype code; no model or data retrieval, no human feedback & annotated, or traced and confirmed, by the authors \\
L1 & L & MLAgentBench: a research agent built on the ReAct framework (thought, action, observation) & annotated, or traced and confirmed, by the authors \\
L2 & L & ResearchAgent: iterative feedback that refines the work & annotated, or traced and confirmed, by the authors \\
I1 & I & Experiment agent rebuilt on ReAct, with iterative feedback from execution & v1 \S2.2--2.3, \S4; sources annotated, or traced and confirmed, by the authors \\
R0 & R & The agent drifts further from the goal over its iterations & annotated, or traced and confirmed, by the authors \\
IN1 & IN & Human correction largely fixes the direction & annotated, or traced and confirmed, by the authors \\
L3 & L & Hugging Face Hub: a repository of models and datasets (external resource) & v1 \S2.2; annotated, or traced and confirmed, by the authors \\
I2 & I & Add retrieval of candidate models and datasets from Hugging Face and a tool for asking a person for help (the system of v1) & v1 \S2.2--2.3, App.\ A--B \\
R1 & R & Five tasks (SemRel, ELLIPSE, IMDB, Spaceship-Titanic, Identify-Contrails), eight trials. Ideas: innovativeness 3.1 $\rightarrow$ 3.9, similarity 0.32 $\rightarrow$ 0.16; execution: +39.7\% over prototype (GPT-4), 40\% of runs $\geq$10\% & v1 \S3, Tables 1--4 \\
R2 & R & Single-prompt baseline fails on every task (0\% success), since it cannot detect and correct execution errors; run before I3 & v3 \S3.2.2; order annotated, or traced and confirmed, by the authors \\
IN2 & IN & Execution succeeds through its feedback loop & v3 \S3.2.2 \\
L4 & L & The group's idea-generation method (Case 3): fine-tuning, then reinforcement learning with novelty, feasibility, effectiveness reward models & v3 \S2.1, fn.\ 4 \\
I3 & I & Idea agent trained (Llama): fine-tuning on 1,000 papers, reinforcement learning with reward models from OpenReview feedback on 4,271 ICLR/NeurIPS papers, no steering at inference; experiment stages kept & v3 \S2.1, \S3.2.1, App.\ B \\
R3 & R & Ideas: innovativeness 3.9 vs ResearchAgent 3.8 vs base LLM 3.1; similarity 0.13, 0.15, 0.32; execution: +44.2\% (Claude-3.7), 50\% of runs $\geq$10\% & v3 Tables 1--4 \\
C2 & C (open) & Stages run largely in sequence; transitions back from implementation to ideation are needed & v3 \S7 \\
\bottomrule
\end{longtable}}
\addtocounter{table}{-1}

{\scriptsize
\begin{longtable}{>{\raggedright\arraybackslash}p{0.041\linewidth}>{\raggedright\arraybackslash}p{0.192\linewidth}>{\raggedright\arraybackslash}p{0.606\linewidth}}
\toprule
\textbf{\#} & \textbf{edge} & \textbf{transition} \\
\midrule
\endfirsthead
\toprule \textbf{\#} & \textbf{edge} & \textbf{transition} \\ \midrule \endhead
1 & RQ0, L0 $\rightarrow$ C1 & Prior systems either do not execute ideas or only make small edits (annotated, or traced and confirmed, by the authors) \\
2 & C1 $\rightarrow$ I0 & Start execution from the paper's own prototype code (annotated, or traced and confirmed, by the authors) \\
3 & I0, L1, L2 $\rightarrow$ I1 & Take the ReAct framework from MLAgentBench's agent and the iterative feedback from ResearchAgent (annotated, or traced and confirmed, by the authors) \\
4 & I1 $\rightarrow$ R0 $\rightarrow$ IN1 & The agent drifts; a person can correct the direction (annotated, or traced and confirmed, by the authors) \\
5 & L3, IN1 $\rightarrow$ I2 & Add retrieval from the Hub, on one researcher's idea that it matches models to the task better, and human feedback to correct the drift (annotated, or traced and confirmed, by the authors) \\
6 & I2 $\rightarrow$ R1 & Evaluate on five tasks \\
7 & R1, R2 $\rightarrow$ IN2 & Execution feedback, not the prompt, makes the runs succeed (stated in v3 \S3.2.2) \\
8 & I2, L4 $\rightarrow$ I3 & Once the group's ideation method exists, replace the prompted idea agent with it and keep the experiment stages (annotated, or traced and confirmed, by the authors) \\
9 & I3 $\rightarrow$ R3 & Evaluate again \\
10 & R3 $\rightarrow$ C2 & Record what remains open \\
\bottomrule
\end{longtable}}
\addtocounter{table}{-1}

\subsection{Case 5: on-policy distillation from one training example}\label{app:cases:3}

\emph{Source: Fu, He, Zuo, Huang, Zhang, Xiao, Qian, Luo, Gao, Wang, Liu, Ding, Xiao, ``Rethinking On-Policy Distillation of Large Language Models II: One Training Example'', arXiv 2609.04172, September 2026 (Tsinghua, UCAS and collaborators). Part I is Li et al., ``Rethinking On-Policy Distillation of Large Language Models: Phenomenology, Mechanism, and Recipe'', arXiv 2604.13016, April 2026.} A shorter account is in \S\ref{sec:cases:3}.

\paragraph{Full account.}
The second paper names one-shot RLVR, not the first paper, as the source of its idea, and cites the first among studies of the training dynamics of OPD and for one of its metrics. The initial idea was the transfer of an experimental lens rather than of a technique. One-shot RLVR \citep{wang2025oneshot} had been used as an extreme controlled experiment to isolate the role of training data in RL with verifiable rewards, and the group brought the same lens to on-policy distillation. As an idea it is underspecified, since ``train OPD on one example and see what happens'' has no mechanism, no prediction, and no boundary.

The first round is the direct experiment, training on a single query for hundreds of steps, with full-data OPD as the reference (17k queries in mathematics), across four domains and three model families. The result is the first piece of evidence, and it is sharper than anything the idea text contained. One-shot OPD keeps improving for hundreds of steps, recovering 69\% of the teacher--student gap and 87\% of full-data OPD's gain. The insight from that result is about \emph{where} supervision comes from. Clustering the states the student visits shows that rollouts from one query reach 65.9\% of the state clusters by step 100 and 71.5\% by step 300, so a single input exposes the student to a large share of the teacher supervision that full-data training exposes. That insight raises a challenge that a text idea could not have raised, whether it is the query's content that matters or only the states its rollouts generate.

The second round answers it by ablation. Content-free templates and off-domain chat prompts drive OPD nearly as effectively as the full training set, response diversity from one query improves accuracy monotonically, and sixteen diverse queries reach 98.9\% coverage and match full-data OPD. The insight is that an input is useful largely because it starts the student reasoning, so inputs are state generators rather than task instructions. This produces the second challenge (C2), why absorption still takes hundreds of steps if states are cheap and broad. The third round measures the absorption rate, the share of the remaining teacher--student distance that one update closes, in runs on one, four, sixteen, and 17k queries, and finds that it declines the same way at every size. An ablation that holds the states fixed, reusing 64 trajectories sampled once from the initial student in a one-shot run, still gains accuracy for about 200 steps, so a steady supply of fresh states is not what keeps a run going. The final idea is a reframing with a recipe attached. OPD is data-overfed but algorithm-starved, queries should be selected by the states they induce, and future work should go into the step efficiency of the algorithm rather than into data.

The trajectory has three rounds and four states of evidence. The decisive transitions do not combine existing components. They sharpen a claim, propose a mechanism for it, and then ablate what the mechanism predicts.

\emph{What a whole-idea regeneration loop would have lost.} Here the final idea is a re-scoped claim rather than a new idea, and each challenge is a question about one specific state. The first challenge (C1) asks whether the state-coverage mechanism, not the query content, explains the first result, and the second (C2) asks why the mechanism leaves absorption slow. An agent that passes back ``68.5 against 69.8'', one-shot against full-data accuracy at step 300, as a score would either stop, because the score is good, or generate an unrelated next experiment. The ablation of content against states exists because the insight predicted it, and that justification is recorded on one transition and nowhere else. Without attribution to states, the final claim reads as an interpretation of a results table rather than as the conclusion of two ablations that a stated mechanism predicted.

\paragraph{Listings.}

{\scriptsize
\begin{longtable}{>{\raggedright\arraybackslash}p{0.042\linewidth}>{\raggedright\arraybackslash}p{0.050\linewidth}>{\raggedright\arraybackslash}p{0.748\linewidth}}
\toprule
\textbf{id} & \textbf{type} & \textbf{content (short)} \\
\midrule
\endfirsthead
\toprule \textbf{id} & \textbf{type} & \textbf{content (short)} \\ \midrule \endhead
RQ0 & RQ & How much training data does on-policy distillation need, and why \\
L0 & L & One-shot RLVR as a controlled experiment isolating the role of data \citep{wang2025oneshot}; Part I \citep{li2026rethinking1}, cited in Part II for OPD's training dynamics and the top-$k$ overlap metric \\
I0 & I & Bring the one-shot lens to OPD: train on a single query (no mechanism, no prediction, no boundary yet) \\
CE0 & CE & One-shot OPD on one query per domain for hundreds of steps; full-data OPD as reference (17k queries in math); 4 domains (math, code, instruction following, agentic tool use) $\times$ 3 families (Qwen, Llama, OLMo) \\
R0 & R & One-shot keeps improving for hundreds of steps; 68.5 vs 69.8 at step 300 (three-benchmark average); 69\% of the teacher--student gap and 87\% of full-data gain recovered \\
IN0 & IN & State coverage (fraction of 200 state clusters reached): one query's rollouts reach 65.9\% by step 100 and 71.5\% by step 300; supervision breadth comes from rollouts, not from prompts \\
C1 & C & Does the query's \emph{content} matter, or only the states its rollouts generate? \\
CE1 & CE & Ablations: content-free templates; off-domain WildChat prompts; response diversity 1/4/16/64 trajectories; query diversity 1/4/16 queries \\
R1 & R & Templates and off-domain prompts nearly match the full set; accuracy rises with trajectory diversity; 16 diverse queries reach 98.9\% coverage and match full-data OPD \\
IN1 & IN & An input is useful because it starts the student reasoning: inputs are state generators \\
C2 & C & If states are cheap and broad, why does absorption still take hundreds of steps? \\
CE2 & CE & Absorption rate (share of the remaining teacher--student distance closed per update) on 1, 4, 16 and 17k queries; separately, a frozen-state one-shot run on 64 trajectories sampled once from the initial student (accuracy, top-16 overlap, overlap-token advantage) \\
R2 & R & The absorption rate declines the same way at every training-set size; the frozen-state run keeps gaining accuracy for about 200 steps, so fresh states are not what keeps a run going \\
IN2 & IN & The bottleneck is the algorithm's absorption rate, not the quantity of supervision \\
I1 & I & Claim + recipe: ``OPD is data-overfed but algorithm-starved''; select queries by the states they induce; future work on step efficiency of OPD \\
\bottomrule
\end{longtable}}
\addtocounter{table}{-1}

{\scriptsize
\begin{longtable}{>{\raggedright\arraybackslash}p{0.041\linewidth}>{\raggedright\arraybackslash}p{0.192\linewidth}>{\raggedright\arraybackslash}p{0.606\linewidth}}
\toprule
\textbf{\#} & \textbf{edge} & \textbf{transition} \\
\midrule
\endfirsthead
\toprule \textbf{\#} & \textbf{edge} & \textbf{transition} \\ \midrule \endhead
1 & L0 $\rightarrow$ I0 & Transfer an experimental lens from RLVR to OPD \\
2 & I0 $\rightarrow$ CE0 $\rightarrow$ R0 & Sharpen ``how much data'' into a measurable one-shot experiment \\
3 & R0 $\rightarrow$ IN0 & Mechanize: explain the gain by state coverage \\
4 & IN0 $\rightarrow$ C1 & Derive a discriminating question from the mechanism \\
5 & C1 $\rightarrow$ CE1 $\rightarrow$ R1 $\rightarrow$ IN1 & Ablate content vs states; confirm inputs are state generators \\
6 & IN1 $\rightarrow$ C2 & The mechanism leaves a residual: slow absorption \\
7 & C2 $\rightarrow$ CE2 $\rightarrow$ R2 $\rightarrow$ IN2 & Isolate the algorithmic bottleneck \\
8 & IN0, IN1, IN2 $\rightarrow$ I1 & Reframe the field's assumption and attach a recipe \\
\bottomrule
\end{longtable}}
\addtocounter{table}{-1}

\subsection{Case 6: on-policy distillation as told by its authors}\label{app:cases:2}

\begin{figure}[htbp]
\centering
\definecolor{cogF}{HTML}{E9EEF7}\definecolor{cogD}{HTML}{7F91B2}%
\definecolor{litF}{HTML}{F8EDE0}\definecolor{litD}{HTML}{C6A271}%
\definecolor{wldF}{HTML}{E5F1EB}\definecolor{wldD}{HTML}{82A896}%
\definecolor{rsF}{HTML}{FAF7F0}\definecolor{rsD}{HTML}{D1C9B6}%
\begin{tikzpicture}[
  font=\scriptsize,
  st/.style={line width=0.6pt, rounded corners=2pt, text=black!80, text width=1.95cm, minimum height=0.62cm, align=center, inner sep=2pt},
  ty/.style={draw, rounded corners=2pt, text width=1.9cm, minimum height=0.6cm, align=center, inner sep=2pt},
  cog/.style={fill=cogF, draw=cogD},
  lit/.style={fill=litF, draw=litD},
  wld/.style={fill=wldF, draw=wldD},
  e/.style={-{Latex[length=1.6mm]}, thick, draw=black!60},
  ed/.style={-{Latex[length=1.6mm]}, thick, dashed, draw=black!35},
  t/.style={font=\tiny, fill=white, inner sep=1pt},
  rs/.style={font=\tiny, align=left, text width=5.2cm, draw=rsD, fill=rsF, rounded corners=2pt, inner sep=3pt},
  grp/.style={font=\tiny, anchor=west, text width=1.7cm, align=left}
]
\draw[dashed, gray!70, rounded corners=3pt] (0.05,0.7) rectangle (15.35,-1.2);
\node[font=\tiny, text=gray!80, anchor=west, inner sep=0pt] at (0.2,-1.06) {Initial RPG (round 1)};
\node[st,cog] (rq0) at (1.2,-0.2)  {Research Question \#0: cheap reasoning for a small model};
\node[st,cog] (i0)  at (3.8,-0.2)  {Idea Update \#0: off-policy distillation from the teacher};
\node[st,wld] (ce0) at (6.4,-0.2)  {Code \& Exp.\ Plan \#0: SFT on 400k teacher prompts};
\node[st,wld] (r0)  at (9.0,-0.2)  {Exp.\ Results \#0: 60\% AIME'24 at 400k; $\sim$2M needed for 70\%};
\node[st,lit] (l0)  at (11.6,-0.2) {Literature \#0: Qwen3 report: OPD 1,800 GPU-h vs 17,920 for RL};
\node[st,lit] (l1)  at (14.2,-0.2) {Literature \#1: DAgger, GKD, MiniLLM, process rewards};
\node[st,cog] (in0) at (11.6,-2.2) {Insight \#0: off-policy compounds errors; RL is sparse};
\node[st,wld] (ce3) at (9.0,-2.2)  {Code \& Exp.\ Plan \#3: one prompt reused for 20 steps};
\node[st,wld] (r6)  at (6.4,-2.2)  {Exp.\ Results \#6: one prompt nearly matches the teacher};
\node[st,cog] (in3) at (3.8,-2.2)  {Insight \#3: reverse KL learns the distribution, not an answer};
\node[st,cog] (c3)  at (1.2,-2.2)  {Challenge \#3: how far does data efficiency go, and why (studied later in Case 5)};
\node[st,cog] (i1)  at (11.6,-4.2) {Idea Update \#1: student samples, teacher grades each token};
\node[st,wld] (ce1) at (9.0,-4.2)  {Code \& Exp.\ Plan \#1: OPD from the 400k SFT checkpoint of 8B-Base};
\node[st,wld] (r1)  at (6.4,-4.2)  {Exp.\ Results \#1: 70\% in $\sim$150 steps; 9--30x cheaper};
\node[st,cog] (c1)  at (3.8,-4.2)  {Challenge \#1: why is dense supervision so much cheaper?};
\node[st,wld] (ce2) at (1.2,-4.2)  {Code \& Exp.\ Plan \#2: RL-train 8B-Base, distill it back into the base};
\node[st,wld] (r2)  at (1.2,-6.2)  {Exp.\ Results \#2: 7--10x faster than RL; 50--100x less compute};
\node[st,cog] (in1) at (3.8,-6.2)  {Insight \#1: RL is search over strategies; distillation skips it};
\node[st,cog] (i3)  at (11.6,-6.2) {Idea Update \#3: old checkpoint as teacher; OPD after mid-training};
\node[st,wld] (r5)  at (14.2,-6.2) {Exp.\ Results \#5: IF-eval 79\% to 83\%; internal QA 36\% to 41\%};
\node[st,cog] (rq1) at (1.2,-8.2)  {Research Question \#1: assistant that knows company docs, keeps IF};
\node[st,cog] (i2)  at (3.8,-8.2)  {Idea Update \#2: mid-train (SFT) on documents mixed with chat data};
\node[st,wld] (r3)  at (6.4,-8.2)  {Exp.\ Results \#3: QA up, IF-eval down at any mixing weight};
\node[st,wld] (r4)  at (9.0,-8.2)  {Exp.\ Results \#4: SFT on the model's own samples still degrades it};
\node[st,cog] (in2) at (11.6,-8.2) {Insight \#2: finite batches drift though KL is 0; fix the teacher};
\node[st,cog] (c2)  at (14.2,-8.2) {Challenge \#2: add knowledge, keep the post-trained behavior};
\draw[e] (rq0) -- node[t]{1} (i0);
\draw[e] (i0) -- (ce0);
\draw[e] (ce0) -- (r0);
\draw[e] (r0.south east) -- (in0.north west);
\draw[e] (l0) -- node[t,anchor=west,xshift=1.5pt]{2} (in0);
\draw[e] (l1.south west) -- (in0.north east);
\draw[e] (in0) -- node[t,anchor=west,xshift=1.5pt]{3} (i1);
\draw[e] (i1) -- node[t]{4} (ce1);
\draw[e] (ce1) -- (r1);
\draw[e] (r1) -- node[t]{5} (c1);
\draw[e] (c1) -- (ce2);
\draw[e] (ce2) -- (r2);
\draw[e] (r2) -- node[t]{6} (in1);
\draw[e] (rq1) -- node[t]{7} (i2);
\draw[e] (i2) -- (r3);
\draw[e] (r3) -- node[t]{8} (r4);
\draw[e] (r4) -- (in2);
\draw[e] (in2) -- (c2);
\draw[e] (in2) -- (i3);
\draw[e] (i1) -- node[t,anchor=west,xshift=1.5pt]{9} (i3);
\draw[e] (i3) -- (r5);
\draw[e] (i1.north west) -- node[t,inner sep=0.5pt]{10} (ce3.south east);
\draw[e] (ce3) -- (r6);
\draw[e] (r6) -- (in3);
\draw[e] (in3) -- (c3);
\node[rs, anchor=east, text width=4.3cm] (rs9) at (10.15,-6.2) {Reasoning on transition 9: no mixing weight preserves IF-eval (Exp.\ Results \#3), and the post's discussion shows that even SFT on a model's own samples degrades it (Exp.\ Results \#4), so the training target must be a fixed teacher. Reuse the on-policy distillation objective of Idea Update \#1 with the earlier post-trained checkpoint as the teacher.};
\draw[ed] (rs9.east) -- (i3.west);
\node[font=\tiny, anchor=north west, text width=3.0cm] at (13.2,-1.6) {Transitions 2 and 9 have several sources. Research Question \#1 opens a second strand (bottom row). One reasoning record is shown.};
\end{tikzpicture}
\caption{Case 6, on-policy distillation, as an RPG. Three and a half rounds, seven experimental and two literature states of evidence, three idea updates after the initial one. Listing below.}
\label{fig:case2}
\end{figure}
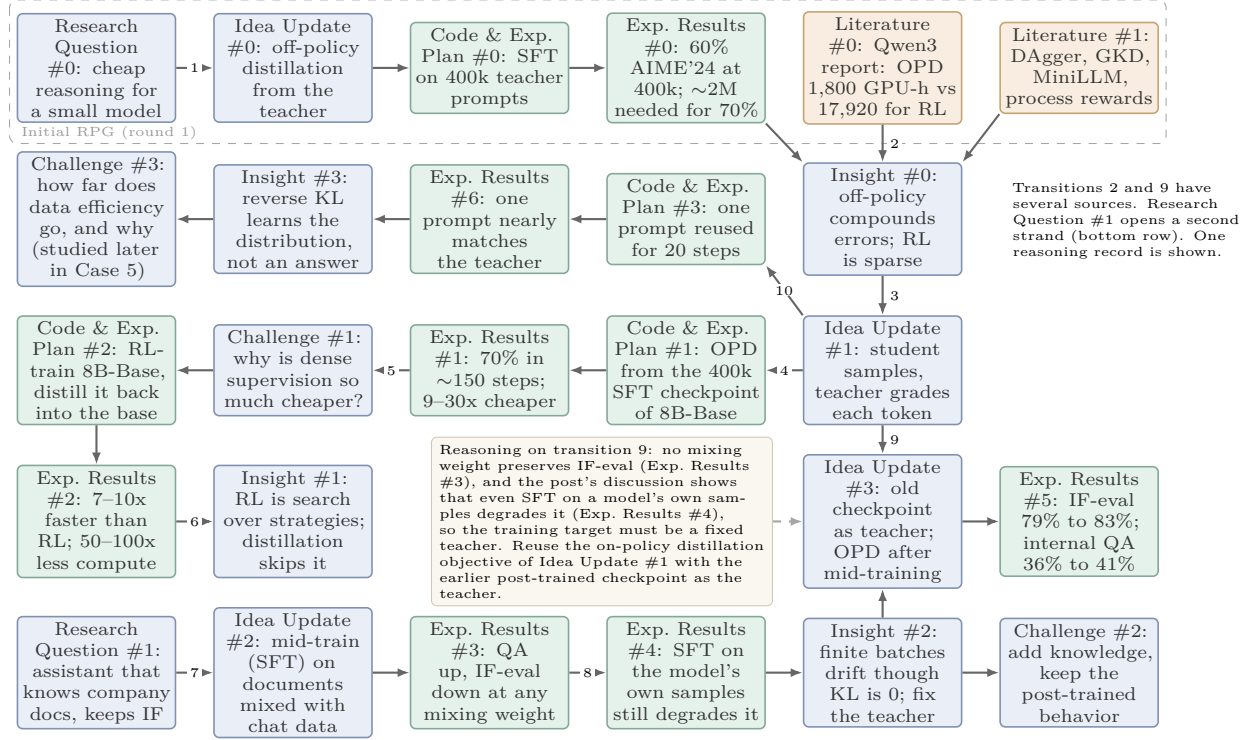

\emph{Source: Kevin Lu with others at Thinking Machines Lab, ``On-Policy Distillation'', blog post, October 27, 2025 (section headings cited in the tables). The prior work the post takes as input is DAgger \citep{ross2011dagger}, GKD \citep{agarwal2024gkd}, MiniLLM \citep{gu2024minillm}, process rewards \citep{lightman2023verify}, and the Qwen3 technical report \citep{qwen3}. This is a single-source case. Every node is traceable to a section of the post.}

\paragraph{Full account.}
The post reports its runs in separate sections and states the reason for most method choices, which is why it can be drawn as an RPG; where it presents an experiment after the result that experiment explains, we order the states by their logical dependence. The research question is how to give a small model the reasoning ability of a large one at low cost. The initial idea is off-policy distillation, which samples solutions from the teacher and fine-tunes the student on them. The authors ran it. Fine-tuned on 400k reasoning prompts with responses generated by QwQ-32B (OpenThoughts-3), the student, Qwen3-8B-Base, reaches 60\% on AIME'24, and extrapolating the log-linear trend says roughly 2M prompts would be needed to reach 70\%. That result is the first injection. Two pieces of literature enter alongside it. One is the Qwen3 report, which had reached 74.4\% with on-policy distillation in 1,800 GPU hours against 17,920 for RL, and the other is the imitation-learning line (DAgger, GKD, MiniLLM). The post states the next insight in its untitled introduction, before it reports any run, so placing it after the first result is our ordering. It is a diagnosis of both baselines at once. Off-policy training makes the student learn in contexts the teacher frequents rather than ones the student will find itself in, so errors compound; RL trains on the student's own contexts but gives very sparse feedback. The idea update combines the two. It samples from the student, has the teacher grade every token, and minimizes the per-token reverse KL, a one-line change on top of an RL loop with a KL regularizer. Run from the same checkpoint, it reaches 70\% AIME'24 in about 150 steps. That completes the first round.

The second round is a why-question about that number. To isolate the effect of dense supervision the authors train Qwen3-8B-Base with RL and then distill the RL-trained policy back into the same base model, and distillation learns the RL-trained policy in 7 to 10 times fewer gradient steps than RL needed. The insight, which the post argues in its discussion without a further experiment, reframes what RL is doing. RL spends its compute on search over semantic strategies, not on the gradient updates, so once a strategy exists, distillation can skip the search.

The third round starts from a different application. Mid-training Qwen3-8B on internal company documents raises internal QA but lowers instruction following, no mixing weight of chat data preserves the original IF-eval score, and LoRA learns less and still forgets. The idea update is to use the earlier post-trained checkpoint, Qwen3-8B, as the teacher and run on-policy distillation on Tulu3 prompts after mid-training, and the reason the post states is that the instruction-following behavior was originally trained with RL, which is expensive and fragile. IF-eval recovers from 79\% to 83\% (85\% before mid-training), and internal QA rises from 36\% to 41\%. On-policy distillation is now a tool for recovering post-trained behavior, which it was not in the initial idea. In its discussion of continual learning the post reports a separate experiment, in which fine-tuning Qwen3-32B on its own samples still degrades it, although the KL between the two is zero in expectation, because every finite batch has a slightly different distribution. The post presents this experiment after the recovery result and argues there that on-policy distillation avoids the drift because the teacher stays fixed, but it does not give the experiment as the reason for the recovery method. It is placed before the idea update because it explains why mixing cannot be the fix. A last half-round raises the data-efficiency question that Case 5 later studies independently. One training prompt, reused for 20 steps, approximately matches the teacher's AIME'24 performance.

The trajectory in Figure~\ref{fig:case2} has three and a half rounds, seven experimental and two literature states of evidence, and three idea updates after the initial one. Two caveats apply. The source is a blog post, not a peer-reviewed paper, and an authorial narrative is itself a retrospective reconstruction, which is the confabulation risk that any author-annotated trajectory carries \citep{nisbett1977telling}, discussed in \S\ref{sec:disc}.

\emph{What a whole-idea regeneration loop would have lost.} The third round is the clearest. A review loop would receive ``instruction following dropped'' and regenerate a data-mixing recipe; the authors did try that, and it is a result node. The change that worked replaced one component of the idea, the training target, with a fixed on-policy teacher. The result that ruled mixing out is R3, in which no mixing weight preserved instruction following, and transition 8 adds the later experiment (R4) showing that even the model's own samples degrade a model. Without nodes for these results and transitions from them to the new idea, the final method reads as ``distill from an old checkpoint'' with no record of the experiments that ruled out the alternatives.

\paragraph{Listings.}
{\scriptsize
\begin{longtable}{>{\raggedright\arraybackslash}p{0.040\linewidth}>{\raggedright\arraybackslash}p{0.040\linewidth}>{\raggedright\arraybackslash}p{0.379\linewidth}>{\raggedright\arraybackslash}p{0.379\linewidth}}
\toprule
\textbf{id} & \textbf{type} & \textbf{content (short)} & \textbf{section of the post} \\
\midrule
\endfirsthead
\toprule \textbf{id} & \textbf{type} & \textbf{content (short)} & \textbf{section of the post} \\ \midrule \endhead
RQ0 & RQ & Give a small model the reasoning ability of a large one at low cost & Distillation for reasoning \\
L0 & L & Qwen3 report: on-policy distillation reaches 74.4\% AIME'24 in 1,800 GPU hours vs 17,920 for RL & Distillation for reasoning: Reinforcement learning \\
I0 & I & Off-policy distillation: sample from the teacher, SFT the student & Distillation for reasoning: Off-policy distillation \\
CE0 & CE & SFT on 400k teacher prompts & Off-policy distillation \\
R0 & R & 60\% AIME'24 at 400k; extrapolated \textasciitilde{}2M prompts for 70\% & Off-policy distillation \\
L1 & L & DAgger; GKD; MiniLLM; process reward models & On-policy distillation --- best of both worlds \\
IN0 & IN & Off-policy: student learns in the teacher's contexts, errors compound. RL: on-policy but sparse, a fixed number of bits per episode & Introduction (untitled) \\
I1 & I & Student samples, teacher grades every token, per-token reverse KL; a one-line change on an RL loop & Implementation: Loss function; Pseudocode \\
CE1 & CE & On-policy distillation into the 400k SFT checkpoint of Qwen3-8B-Base (teacher Qwen3-32B in the text, Qwen3-8B per footnote) & Distillation for reasoning: On-policy distillation \\
R1 & R & 70\% AIME'24 in \textasciitilde{}150 steps; 9x (SFT data given) to 30x cost reduction & On-policy distillation \\
C1 & C & Why is dense supervision so much cheaper? & Discussion: Dense supervision greatly improves compute efficiency \\
CE2 & CE & RL-train Qwen3-8B-Base on DeepMath; distill the RL policy back into the base & Dense supervision \\
R2 & R & Teacher level reached 7 to 10x faster than RL; 50 to 100x less compute & Dense supervision \\
IN1 & IN & RL spends its compute on search over semantic strategies; distillation skips the search & Discussion: RL searches in the space of semantic strategies \\
RQ1 & RQ & Train an internal assistant that knows company documents and keeps instruction following & Distillation for personalization: Training an internal assistant \\
I2 & I & Mid-train (SFT) on internal documents mixed with chat data & Training an internal assistant \\
R3 & R & Internal QA up, IF-eval down; no mixing weight preserves IF-eval; LoRA learns less & Training on new knowledge degrades learned behavior \\
R4 & R & SFT of Qwen3-32B on its own samples (temperature 1.0, Tulu3 prompts) still degrades it; reported after R5 & Discussion: On-policy learning as a tool for continual learning \\
IN2 & IN & KL is 0 in expectation but every finite batch differs; only a fixed teacher keeps the update on-policy & Continual learning \\
C2 & C & How to add knowledge without leaving the post-trained behavior & Training on new knowledge degrades learned behavior \\
I3 & I & Use the earlier post-trained checkpoint as teacher; on-policy distillation on instruction prompts after mid-training & On-policy distillation recovers post-training behavior \\
R5 & R & IF-eval 79\% to 83\% (85\% before mid-training); internal QA 36\% to 41\% & On-policy distillation recovers post-training behavior \\
CE3 & CE & One prompt reused for 20 steps & Discussion: Distillation can effectively reuse training data for data efficiency \\
R6 & R & One prompt reused for 20 steps approximately matches the teacher's AIME'24 performance & Data efficiency \\
IN3 & IN & Reverse KL learns the teacher's distribution, not a memorized answer & Data efficiency \\
C3 & C & How far does data efficiency go, and why (studied later, independently, in Case 5) & Data efficiency \\
\bottomrule
\end{longtable}}
\addtocounter{table}{-1}

{\scriptsize
\begin{longtable}{>{\raggedright\arraybackslash}p{0.041\linewidth}>{\raggedright\arraybackslash}p{0.229\linewidth}>{\raggedright\arraybackslash}p{0.570\linewidth}}
\toprule
\textbf{\#} & \textbf{edge} & \textbf{transition} \\
\midrule
\endfirsthead
\toprule \textbf{\#} & \textbf{edge} & \textbf{transition} \\ \midrule \endhead
1 & RQ0 $\rightarrow$ I0 $\rightarrow$ CE0 $\rightarrow$ R0 & Run the default and extrapolate its cost \\
2 & R0, L0, L1 $\rightarrow$ IN0 & Diagnose both baselines; off-policy compounds error, RL is sparse (the post states the diagnosis in its introduction) \\
3 & IN0 $\rightarrow$ I1 & Combine on-policy sampling with dense teacher grading \\
4 & I1 $\rightarrow$ CE1 $\rightarrow$ R1 & Run from the same checkpoint; measure cost \\
5 & R1 $\rightarrow$ C1 $\rightarrow$ CE2 $\rightarrow$ R2 & Isolate dense supervision by distilling an RL policy back into its base \\
6 & R2 $\rightarrow$ IN1 & Reframe RL as search \\
7 & RQ1 $\rightarrow$ I2 $\rightarrow$ R3 & Try mixing; observe that nothing preserves IF-eval \\
8 & R3 $\rightarrow$ R4 $\rightarrow$ IN2 $\rightarrow$ C2 & Test SFT on own samples; attribute degradation to finite-batch drift; the post reports R4 later, in its discussion, on Qwen3-32B \\
9 & IN2, I1 $\rightarrow$ I3 $\rightarrow$ R5 & Reuse the OPD objective with the old checkpoint as teacher; the stated reason is that the behavior came from expensive, fragile RL \\
10 & I1 $\rightarrow$ CE3 $\rightarrow$ R6 $\rightarrow$ IN3 $\rightarrow$ C3 & Probe data reuse; open the data-efficiency question \\
\bottomrule
\end{longtable}}
\addtocounter{table}{-1}

\subsection{Case 7: from judging research ideas to executing them}\label{app:cases:6}

\begin{figure}[htbp]
\centering
\definecolor{cogF}{HTML}{E9EEF7}\definecolor{cogD}{HTML}{7F91B2}%
\definecolor{litF}{HTML}{F8EDE0}\definecolor{litD}{HTML}{C6A271}%
\definecolor{wldF}{HTML}{E5F1EB}\definecolor{wldD}{HTML}{82A896}%
\definecolor{rsF}{HTML}{FAF7F0}\definecolor{rsD}{HTML}{D1C9B6}%
\begin{tikzpicture}[
  font=\scriptsize,
  st/.style={line width=0.6pt, rounded corners=2pt, text=black!80, text width=1.95cm, minimum height=0.62cm, align=center, inner sep=2pt},
  ty/.style={draw, rounded corners=2pt, text width=1.9cm, minimum height=0.6cm, align=center, inner sep=2pt},
  cog/.style={fill=cogF, draw=cogD},
  lit/.style={fill=litF, draw=litD},
  wld/.style={fill=wldF, draw=wldD},
  e/.style={-{Latex[length=1.6mm]}, thick, draw=black!60},
  ed/.style={-{Latex[length=1.6mm]}, thick, dashed, draw=black!35},
  t/.style={font=\tiny, fill=white, inner sep=1pt},
  rs/.style={font=\tiny, align=left, text width=5.2cm, draw=rsD, fill=rsF, rounded corners=2pt, inner sep=3pt}
]
\node[st,cog] (rq0) at (1.2,-0.2)   {Research Question: can LLMs generate ideas as good as experts'?};
\node[st,lit] (l0)  at (1.2,-2.2)   {Literature \#0: ideation systems; small studies or LLM judges};
\node[st,cog] (i0)  at (3.56,-0.2)  {Idea Update \#0: large blind human study; agent: RAG, 4,000 seeds, ranker};
\node[st,wld] (ce0) at (5.92,-0.2)  {Code \& Exp.\ Plan \#0: 100+ NLP researchers write and review blind};
\node[st,wld] (r0)  at (8.28,-0.2)  {Exp.\ Results \#0: AI ideas more novel, 5.64 vs 4.84; feasibility slightly lower (n.s.)};
\node[st,wld] (r1)  at (10.64,-0.2) {Exp.\ Results \#1: 200 unique of 4,000; ranker 53.3\%, reviewers 56.1\% agreement};
\node[st,cog] (in0) at (13.00,-0.2) {Insight \#0: judging ideas without execution is subjective, even for experts};
\node[st,cog] (c1)  at (15.36,-0.2) {Challenge \#1: do ideas judged better produce better research when executed?};
\node[st,cog] (i1)  at (15.36,-4.4) {Idea Update \#1: execution study; 43 researchers each implement one 2024 idea};
\node[st,wld] (ce1) at (13.00,-4.4) {Code \& Exp.\ Plan \#1: 24 AI, 19 human ideas; 93.7 / 112.6 h; 58 blind reviewers};
\node[st,wld] (r2)  at (10.64,-4.4) {Exp.\ Results \#2: overall drop $-$1.98 vs $-$0.63; effectiveness lead reversed (n.s.)};
\node[st,wld] (r3)  at (8.28,-4.4)  {Exp.\ Results \#3: executors changed only details; eval substitution explains little};
\node[st,cog] (in1) at (5.92,-4.4)  {Insight \#1: execution evaluation uses empirical outcomes; ideation evaluation speculates};
\node[st,cog] (c2)  at (3.56,-4.4)  {Challenge \#2: proxy reward models; closed loops from executed outcomes to ideas};
\draw[dashed, gray!70, rounded corners=3pt] ([shift={(-0.10,0.12)}]i0.north -| rq0.west) rectangle ([shift={(0.12,-0.3)}]l0.south -| i0.east);
\node[font=\tiny, text=gray!70, anchor=south east, inner sep=1pt] at ([shift={(0.08,-0.28)}]l0.south -| i0.east) {Initial RPG (2024 study)};
\draw[e] (rq0) -- node[t]{1} (i0);
\draw[e] (l0.east) -- (i0.south west);
\draw[e] (i0) -- (ce0);
\draw[e] (ce0) -- (r0);
\draw[e] (r0) -- node[t]{2} (r1);
\draw[e] (r1) -- node[t]{3} (in0);
\draw[e] (in0) -- (c1);
\draw[e] (c1) -- node[t]{4} (i1);
\draw[e] (i1) -- (ce1);
\draw[e] (ce1) -- (r2);
\draw[e] (r2) -- node[t]{5} (r3);
\draw[e] (r3) -- (in1);
\draw[e] (in1) -- node[t]{6} (c2);
\node[rs, anchor=north] (rs3) at (11.82,-1.55) {Reasoning on transition 3: the study's own numbers (reviewer--reviewer agreement 56.1\%, ranker--human agreement 53.3\%, 200 unique ideas of 4,000) are read as a limit of judging ideas as text, not as noise, so the next step changes the judge, from idea reviews to executed projects.};
\draw[ed] (rs3.north) -- (11.82,-0.42);
\node[font=\tiny, anchor=north west, text width=6.4cm] at (0.1,-5.65) {Transition 1 has two sources. One round; Challenge \#2 leaves the next round open. The idea update changes the judge, not the idea text, and Insight \#1 is again about the judge. Same three authors across both papers. One reasoning record is shown.};
\end{tikzpicture}
\caption{Case 7, from the ideation study to the ideation--execution gap, as an RPG. One round and an open challenge, five states of evidence. Listing below.}
\label{fig:case6}
\end{figure}
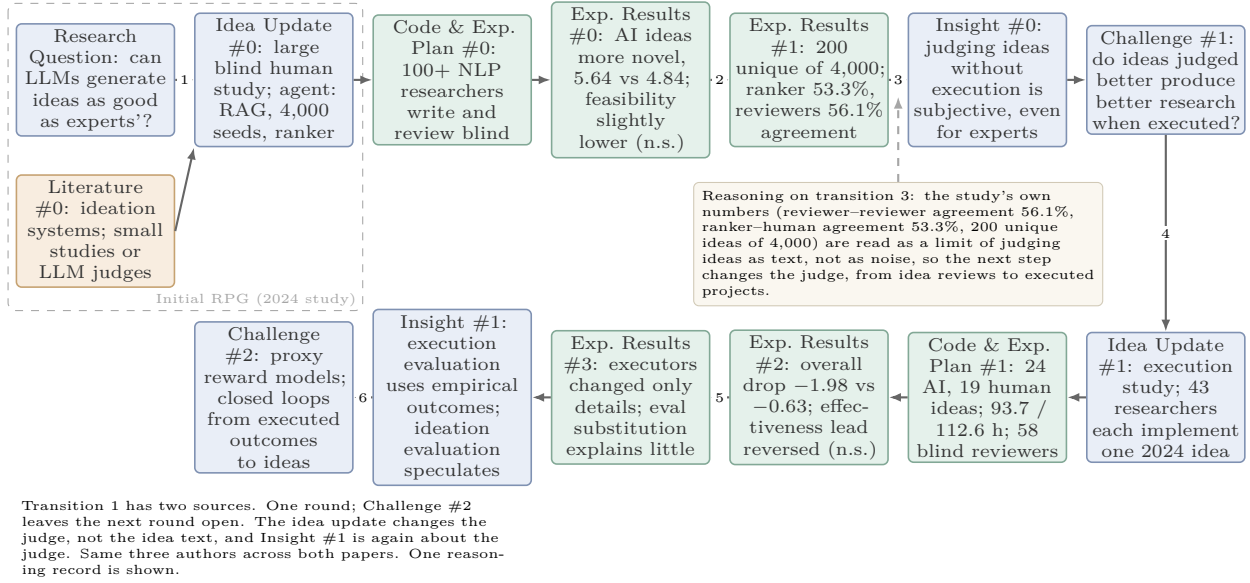

\emph{Sources: \citet{si2024novel} and \citet{si2025gap}, by the same three authors. The second paper restates the first paper's finding and open question in its introduction and executes the same ideas. Every edge cites its section.}

\paragraph{Full account.}
The research question is whether language models can generate research ideas as good as experts'. The literature the first paper starts from is the ideation systems of 2023 and 2024 and their evaluations, which were small or done by LLM judges (\S1). The initial idea is to settle the question with a study large enough to be believed. More than a hundred NLP researchers write and blindly review ideas, against an LLM agent built from retrieval over Semantic Scholar, over-generation of 4,000 seed ideas per topic, deduplication, and a Swiss-tournament pairwise ranker, which is a prompted Claude-3.5-Sonnet model validated on ICLR 2024 submissions (\S3, \S3.3). The headline result is that AI ideas are judged more novel (5.64 vs 4.84, \S5.1), with feasibility slightly lower but not significantly so (abstract, \S5). The paper's own analyses are the second injection. Only about 200 of the 4,000 generated ideas are unique (\S7.1), the ranker agrees with humans 53.3\% of the time (\S7.2), and the human reviewers agree with each other 56.1\% of the time (\S6.3). The insight the authors draw from their own numbers is about the judge, not the generator. Judging an idea without executing it is subjective even for experts, and novelty judged at ideation may not hold once the idea is executed. They propose that evaluation should move to the execution phase (\S6.3, \S10). The challenge is stated as a question, whether the ideas judged better produce better research when executed.

The second paper is the answer, and its introduction records the reason for the step in the authors' own words. The first study ``found that LLM ideas are judged as significantly more novel than human ideas'', but ``evaluating research ideas is difficult even for experts, leaving open the question of whether these ideas would translate into better research outcomes'', because ``during execution, every single step has to be grounded in realistic execution constraints, which impose higher feasibility standards than the ideation stage'' (\S1). The idea update is an execution study. Forty-three researchers each implement one idea, randomly assigned from the human or the AI pool of the first study (24 AI, 19 human), spending on average 93.7 and 112.6 hours, and the finished projects are reviewed blind by 58 reviewers, four to five per project, on novelty, excitement, effectiveness, soundness, and overall quality (\S2 to \S3). The result is the third injection. On the four metrics rated at both stages, AI ideas lose far more than human ideas from ideation to execution, novelty $-$1.05 vs $-$0.01, excitement $-$1.76 vs +0.08, effectiveness $-$1.88 vs $-$0.05, overall $-$1.98 vs $-$0.63 (Table 5, \S4.2; soundness was not rated at ideation). On effectiveness the AI ideas lead at ideation (6.00 vs 4.83) and trail after execution (4.13 vs 4.78), a difference that is not significant when each idea is one data point (Table 4, \S4.2). Two further analyses attribute the gap. Executors changed only experiment details, about three per project, ``rather than altering any algorithms proposed in the original ideas'' (\S5.1, Table 6), so the gap belongs to the ideas. Six AI ideas had a proposed human evaluation replaced by an automatic one, and the authors conclude that this and the other changes ``only explain a small fraction'' of the gap (\S5.1). The insight is a statement about evaluation itself, ``execution evaluation always considers empirical experiment outcomes, while the ideation evaluation often speculates'' (\S5.2). The challenge the authors leave is the one this paper takes up, proxy reward models that predict execution outcomes, and ``closed feedback loops where the outcomes of executed experiments inform iterative idea improvement'' (\S7.3).

The trajectory in Figure~\ref{fig:case6} has one round, which ends at the execution study (I1), and a second round left open (C2). It has five states of evidence, one from the literature and four results, two of them from studies with people. The idea whose path this is was never a system. It was a claim about evaluation, its one update changed the judge, and the insight that followed (IN1) is again about the judge.

\emph{What a whole-idea regeneration loop would have lost.} A loop that receives ``5.64 vs 4.84'' regenerates a more novel generator. The update that mattered was a change to the judge, caused by two agreement numbers and the diversity result (R1) read as a limit of judging ideas as text (IN0), and the insight after it was again about the judge (IN1). Neither change touches the idea text, and both are recordable only as states with edges back to the results that caused them. The case also shows what the two papers could not hold between them. About nine months passed between the question (C1) and its answer, and the reason for the answer had to be restated in the second introduction because there was nowhere else to keep it.

\paragraph{Listings.}
{\scriptsize
\begin{longtable}{>{\raggedright\arraybackslash}p{0.045\linewidth}>{\raggedright\arraybackslash}p{0.10\linewidth}>{\raggedright\arraybackslash}p{0.50\linewidth}>{\raggedright\arraybackslash}p{0.235\linewidth}}
\toprule
\textbf{id} & \textbf{type} & \textbf{content (short)} & \textbf{source} \\
\midrule
\endfirsthead
\toprule \textbf{id} & \textbf{type} & \textbf{content (short)} & \textbf{source} \\ \midrule \endhead
RQ0 & RQ & Can LLMs generate research ideas as good as experts'? & Si et al. 2025 \S1 \\
L0 & L & Ideation systems and their evaluations: small studies or LLM judges & Si et al. 2025 \S1, \S2 \\
I0 & I & A large blind human study; LLM agent = RAG + 4,000 seeds per topic + dedup + Swiss-tournament ranker & Si et al. 2025 \S3 \\
CE0 & CE & 100+ NLP researchers write and review; 79 reviewers; blind & Si et al. 2025 \S4 \\
R0 & R & AI ideas more novel (5.64 vs 4.84); feasibility slightly lower, not significant & Si et al. 2025 abstract, \S5.1 \\
R1 & R (self-analysis) & about 200 unique of 4,000 ideas; ranker 53.3\% agreement with humans; reviewer--reviewer agreement 56.1\% & Si et al. 2025 \S7.1, \S7.2, \S6.3 \\
IN0 & IN & Idea-level judgment without execution is subjective even for experts; judged novelty may not survive execution & Si et al. 2025 \S6.3, \S10 \\
C1 & C & Do ideas judged better produce better research when executed? & Si et al. 2025 \S10; Si et al. 2026 \S1 (quoted) \\
I1 & I & Execution study: 43 researchers each implement one randomly assigned idea from the 2024 pools; blind review of the projects & Si et al. 2026 \S2, \S3 \\
CE1 & CE & 24 AI and 19 human ideas; 93.7 / 112.6 hours per idea; 58 reviewers, 181 reviews, 4--5 per project & Si et al. 2026 \S2, \S3 \\
R2 & R & Ideation $\rightarrow$ execution drop on the four metrics rated at both stages: novelty $-$1.05 vs $-$0.01; excitement $-$1.76 vs +0.08; effectiveness $-$1.88 vs $-$0.05; overall $-$1.98 vs $-$0.63; effectiveness AI ahead at ideation (6.00 vs 4.83), behind after execution (4.13 vs 4.78, not significant per idea) & Si et al. 2026 Tables 4, 5 \\
R3 & R (attribution) & Executors changed only experiment details (2.9--3.1 per project), no algorithms; removing human evaluation (6 AI ideas) explains only a small fraction & Si et al. 2026 Table 6, \S5.1 \\
IN1 & IN & ``Execution evaluation always considers empirical experiment outcomes, while the ideation evaluation often speculates''; execution raises the standard of rigor & Si et al. 2026 \S5.2 (quoted) \\
C2 & C (open) & Proxy reward models for execution outcomes; ``closed feedback loops where the outcomes of executed experiments inform iterative idea improvement'' & Si et al. 2026 \S7.3 (quoted) \\
\bottomrule
\end{longtable}}
\addtocounter{table}{-1}

{\scriptsize
\begin{longtable}{>{\raggedright\arraybackslash}p{0.035\linewidth}>{\raggedright\arraybackslash}p{0.25\linewidth}>{\raggedright\arraybackslash}p{0.595\linewidth}}
\toprule
\textbf{\#} & \textbf{edge} & \textbf{transition} \\
\midrule
\endfirsthead
\toprule \textbf{\#} & \textbf{edge} & \textbf{transition} \\ \midrule \endhead
1 & RQ0, L0 $\rightarrow$ I0 $\rightarrow$ CE0 $\rightarrow$ R0 & Replace small or automatic evaluation by a large blind human study; measure \\
2 & R0 $\rightarrow$ R1 & Analyze the study's own generator and judges \\
3 & R1 $\rightarrow$ IN0 $\rightarrow$ C1 & Read the agreement and diversity numbers as a limit of idea-level judgment; ask the execution question \\
4 & C1 $\rightarrow$ I1 $\rightarrow$ CE1 $\rightarrow$ R2 & Execute a subset of the same ideas; review blind; measure the drop \\
5 & R2 $\rightarrow$ R3 $\rightarrow$ IN1 & Rule out executor changes and evaluation substitution; locate the gap in the ideas and the judge \\
6 & IN1 $\rightarrow$ C2 & Leave the loop open: outcomes should feed back into idea improvement \\
\bottomrule
\end{longtable}}
\addtocounter{table}{-1}

\subsection{Case 8: from GRPO to DeepSeek-R1, with recorded failed explorations}\label{app:cases:7}

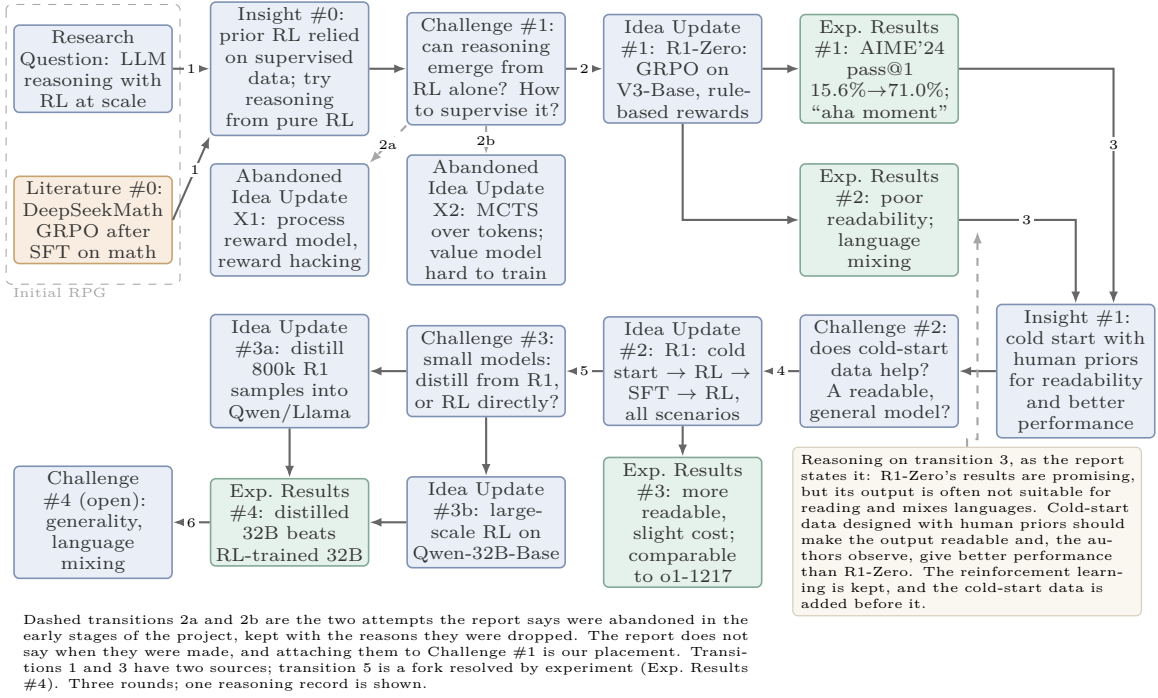
\begin{figure}[htbp]
\centering
\definecolor{cogF}{HTML}{E9EEF7}\definecolor{cogD}{HTML}{7F91B2}%
\definecolor{litF}{HTML}{F8EDE0}\definecolor{litD}{HTML}{C6A271}%
\definecolor{wldF}{HTML}{E5F1EB}\definecolor{wldD}{HTML}{82A896}%
\definecolor{rsF}{HTML}{FAF7F0}\definecolor{rsD}{HTML}{D1C9B6}%
\begin{tikzpicture}[
  font=\scriptsize,
  st/.style={line width=0.6pt, rounded corners=2pt, text=black!80, text width=1.95cm, minimum height=0.62cm, align=center, inner sep=2pt},
  ty/.style={draw, rounded corners=2pt, text width=1.9cm, minimum height=0.6cm, align=center, inner sep=2pt},
  cog/.style={fill=cogF, draw=cogD},
  lit/.style={fill=litF, draw=litD},
  wld/.style={fill=wldF, draw=wldD},
  txt/.style={draw, rounded corners=2pt, text width=2.2cm, minimum height=0.62cm, align=center, inner sep=2pt, fill=gray!8},
  pr/.style={draw, dashed, gray!70, rounded corners=2pt, text width=2.6cm, minimum height=0.62cm, align=center, inner sep=2pt, text=gray!70},
  e/.style={-{Latex[length=1.6mm]}, thick, draw=black!60},
  ed/.style={-{Latex[length=1.6mm]}, thick, dashed, draw=black!35},
  t/.style={font=\tiny, fill=white, inner sep=1pt},
  rs/.style={font=\tiny, align=left, text width=5.2cm, draw=rsD, fill=rsF, rounded corners=2pt, inner sep=3pt},
  grp/.style={font=\tiny, anchor=west, text width=1.7cm, align=left}
]
\node[st,cog] (rq0) at (1.2,-0.2)  {Research Question: LLM reasoning with RL at scale};
\node[st,lit] (l0)  at (1.2,-2.2)  {Literature \#0: DeepSeekMath GRPO after SFT on math};
\node[st,cog] (in0) at (3.8,-0.2)  {Insight \#0: prior RL relied on supervised data; try reasoning from pure RL};
\node[st,cog] (c1)  at (6.4,-0.2)  {Challenge \#1: can reasoning emerge from RL alone? How to supervise it?};
\node[st,cog] (i1)  at (9.0,-0.2)  {Idea Update \#1: R1-Zero: GRPO on V3-Base, rule-based rewards};
\node[st,wld] (r1)  at (11.6,-0.2) {Exp.\ Results \#1: AIME'24 pass@1 15.6\%$\to$71.0\%; ``aha moment''};
\node[st,wld] (r2)  at (11.6,-2.2) {Exp.\ Results \#2: poor readability; language mixing};
\node[st,cog] (x1)  at (3.8,-2.2)  {Abandoned Idea Update X1: process reward model, reward hacking};
\node[st,cog] (x2)  at (6.4,-2.2)  {Abandoned Idea Update X2: MCTS over tokens; value model hard to train};
\node[st,cog] (in1) at (14.2,-4.2) {Insight \#1: cold start with human priors for readability and better performance};
\node[st,cog] (c2)  at (11.6,-4.2) {Challenge \#2: does cold-start data help? A readable, general model?};
\node[st,cog] (i2)  at (9.0,-4.2)  {Idea Update \#2: R1: cold start $\to$ RL $\to$ SFT $\to$ RL, all scenarios};
\node[st,wld] (r3)  at (9.0,-6.2)  {Exp.\ Results \#3: more readable, slight cost; comparable to o1-1217};
\node[st,cog] (c3)  at (6.4,-4.2)  {Challenge \#3: small models: distill from R1, or RL directly?};
\node[st,cog] (i3a) at (3.8,-4.2)  {Idea Update \#3a: distill 800k R1 samples into Qwen/Llama};
\node[st,cog] (i3b) at (6.4,-6.2)  {Idea Update \#3b: large-scale RL on Qwen-32B-Base};
\node[st,wld] (r4)  at (3.8,-6.2)  {Exp.\ Results \#4: distilled 32B beats RL-trained 32B};
\node[st,cog] (c4)  at (1.2,-6.2)  {Challenge \#4 (open): generality, language mixing};
\draw[dashed, gray!70, rounded corners=2pt] (0.05,0.65) rectangle (2.35,-3.05);
\node[font=\tiny, text=gray!70, anchor=north west, inner sep=1pt] at (0.1,-3.05) {Initial RPG};
\draw[e] (rq0) -- node[t]{1} (in0);
\draw[e] (l0.east) -- node[t,pos=0.6]{1} (in0.south west);
\draw[e] (in0) -- (c1);
\draw[e] (c1) -- node[t]{2} (i1);
\draw[e] (i1) -- (r1);
\draw[e] (i1.south) |- (r2.west);
\draw[ed] (c1.south west) -- node[t,pos=0.5]{2a} (x1.north east);
\draw[ed] (c1) -- node[t]{2b} (x2);
\draw[e] (r2.east) -| (in1.north);
\node[t] at (13.55,-2.2) {3};
\draw[e] (r1.east) -| ([xshift=0.5cm]in1.north);
\node[t] at (14.7,-1.2) {3};
\draw[e] (in1) -- (c2);
\draw[e] (c2) -- node[t]{4} (i2);
\draw[e] (i2) -- (r3);
\draw[e] (i2) -- node[t]{5} (c3);
\draw[e] (c3) -- (i3a);
\draw[e] (c3) -- (i3b);
\draw[e] (i3a) -- (r4);
\draw[e] (i3b) -- (r4);
\draw[e] (r4) -- node[t]{6} (c4);
\node[rs, text width=4.4cm, anchor=north west] (rs3) at (10.45,-5.2) {Reasoning on transition 3, as the report states it: R1-Zero's results are promising, but its output is often not suitable for reading and mixes languages. Cold-start data designed with human priors should make the output readable and, the authors observe, give better performance than R1-Zero. The reinforcement learning is kept, and the cold-start data is added before it.};
\draw[ed] (rs3.north) -- ++(0,0.1) -| (12.9,-2.35);
\node[font=\tiny, anchor=north west, text width=9.6cm] at (0.15,-7.3) {Dashed transitions 2a and 2b are the two attempts the report says were abandoned in the early stages of the project, kept with the reasons they were dropped. The report does not say when they were made, and attaching them to Challenge \#1 is our placement. Transitions 1 and 3 have two sources; transition 5 is a fork resolved by experiment (Exp.\ Results \#4). Three rounds; one reasoning record is shown.};
\end{tikzpicture}
\caption{Case 8, DeepSeek-R1, as an RPG. Abandoned updates are drawn as dashed transitions. Three rounds, two recorded abandoned branches, one fork, five states of evidence. The report places the abandoned attempts only in the early stages of the project. Listing below.}
\label{fig:case7}
\end{figure}

\emph{Sources: \citet{shao2024deepseekmath}, which introduces GRPO, and the first arXiv version of \citet{guo2025deepseekr1} (January 2025), from which all quotations, numbers, and section numbers below are taken. The report states the reason for each stage and records two attempts that failed ``in the early stages of developing DeepSeek-R1'', with the reasons for abandoning them. It does not say when they were made relative to R1-Zero.}

\paragraph{Full account.}
The research question is how to improve reasoning in language models with reinforcement learning at scale. The literature the idea starts from is the group's own. DeepSeekMath's GRPO drops the critic and uses a group-relative baseline, and it was applied after supervised fine-tuning on mathematics (DeepSeekMath \S4; R1 \S2.2.1). The first insight is stated at the start of the R1-Zero section. The group's earlier reinforcement learning for reasoning ``heavily depended on supervised data, which are time-intensive to gather'', so the report explores ``the potential of LLMs to develop reasoning capabilities without any supervised data, focusing on their self-evolution through a pure reinforcement learning process'' (\S2.2). The challenge that follows is whether reasoning can emerge from RL alone, with no supervised data at all. The idea update is DeepSeek-R1-Zero, GRPO on the DeepSeek-V3 base model with rule-based rewards for accuracy and format and a think/answer template (\S2.2.1 to \S2.2.3). It uses no neural reward model, because the authors ``find that the neural reward model may suffer from reward hacking in the large-scale reinforcement learning process'' (\S2.2.2). The result has two halves. AIME 2024 pass@1 rises from 15.6\% to 71.0\%, 86.7\% with majority voting, and the authors report the model lengthening its thinking, re-checking its steps, and the ``aha moment'' of rethinking a problem, none of it explicitly programmed (\S2.2.4). The other half is the drawback paragraph, in which R1-Zero ``struggles with challenges like poor readability, and language mixing'' (\S2.2.4), with several languages inside one response (\S2.3.1).

The second round starts from both halves of that result. The report says that the ``promising results'' of R1-Zero raise two questions, whether a small amount of high-quality cold-start data can further improve reasoning or speed up convergence, and how to train a user-friendly model with clear chains of thought and strong general capabilities (\S2.3). It gives two reasons for the cold-start data (\S2.3.1). The first is readability, since R1-Zero's ``content is often not suitable for reading''. The second is potential, since with cold-start data designed with human priors the authors ``observe better performance against DeepSeek-R1-Zero''. The insight is therefore that a small amount of human-designed data should fix the form of the output and also raise performance, while the reinforcement learning is kept. The idea update is DeepSeek-R1, a four-stage pipeline (\S2.3.1 to \S2.3.4). It starts with cold start on thousands of long chains of thought, followed by reasoning-oriented RL with a language-consistency reward. Rejection sampling and supervised fine-tuning follow on about 600k reasoning and 200k non-reasoning samples, and the non-reasoning data is added ``to enhance the model's capabilities in writing, role-playing'' and other general tasks (\S2.3.3). A final RL stage covers all scenarios (\S2.3.4). The result records a trade-off the authors chose to accept. The language-consistency reward causes ``a slight degradation in the model's performance'' but ``aligns with human preferences, making it more readable'' (\S2.3.2), and R1 performs comparably to OpenAI-o1-1217 on reasoning tasks (abstract; for mathematics, \S3.1).

The report also records two attempts that failed ``in the early stages of developing DeepSeek-R1'' (\S4.2). Process reward models were dropped because it is hard to define a fine-grained step in general reasoning and hard to judge whether an intermediate step is correct, and because a model-based process reward ``inevitably leads to reward hacking'' and needs a retraining pipeline. Monte Carlo tree search over tokens was dropped for two reasons. The search space is exponentially larger than in chess, and the maximum extension limit per node that bounds it can leave the model stuck in local optima. The fine-grained value model that guides the search is hard to train, so the model cannot improve itself iteratively through search. The report does not say when the two attempts were made, and we draw both as sibling edges out of the challenge of how to supervise reasoning (C1), each with the reason it was dropped.

The third round is a fork on small models, resolved by experiment. The two branches are to distill R1's 800k samples into Qwen and Llama models from 1.5B to 70B (\S2.4) or to run the same large-scale RL on a 32B model directly. The result is that the distilled 32B model beats the RL-trained one, 72.6 against 47.0 pass@1 on AIME 2024 (Table 6), and the authors conclude that ``distilling more powerful models into smaller ones yields excellent results, whereas smaller models relying on the large-scale RL ... may not even achieve the performance of distillation'' (\S4.1). The conclusion lists the open challenges, which are general capabilities, language mixing beyond Chinese and English, prompt sensitivity, and RL for software engineering, where evaluation is too slow (\S5).

The trajectory in Figure~\ref{fig:case7} has three rounds, two recorded dead ends, and one fork, with five states of evidence, four of them experimental. This is the only case whose source reports abandoned approaches as separate branches, each with the reason it was dropped, and a representation that holds only the current idea cannot hold them at all.

\emph{What a whole-idea regeneration loop would have lost.} A results loop receiving ``71\% on AIME'' keeps training. The update that produced DeepSeek-R1 came from the drawback paragraph (R2) read together with the promising results (R1), and the report states the reason on that edge (IN1). Cold-start data designed with human priors should make the output readable and also raise performance, so the authors added a small amount of it and kept the RL. The two abandoned branches are the sharper point. A loop that regenerates the whole idea has no node for ``we tried a process reward model and here is why it failed'', and it would try it again. The RPG keeps X1 and X2 as sibling edges with their reasons, which is the record a later group, or the same group a year later, needs.

\paragraph{Listings.}
{\scriptsize
\begin{longtable}{>{\raggedright\arraybackslash}p{0.045\linewidth}>{\raggedright\arraybackslash}p{0.10\linewidth}>{\raggedright\arraybackslash}p{0.50\linewidth}>{\raggedright\arraybackslash}p{0.235\linewidth}}
\toprule
\textbf{id} & \textbf{type} & \textbf{content (short)} & \textbf{source} \\
\midrule
\endfirsthead
\toprule \textbf{id} & \textbf{type} & \textbf{content (short)} & \textbf{source} \\ \midrule \endhead
RQ0 & RQ & How to improve LLM reasoning with reinforcement learning at scale & R1 \S1 \\
L0 & L & DeepSeekMath: GRPO (no critic, group-relative baseline) after SFT on mathematics; the group's own prior work & DeepSeekMath \S4; R1 \S2.2.1 \\
IN0 & IN & Earlier RL for reasoning ``heavily depended on supervised data''; explore reasoning ``without any supervised data'', through pure RL & R1 \S2.2 (quoted) \\
C1 & C & Can reasoning emerge from RL on the base model with no supervised data? How to supervise it? & R1 \S2.2, \S2.2.2 \\
I1 & I & DeepSeek-R1-Zero: GRPO on DeepSeek-V3-Base; rule-based accuracy + format rewards; think/answer template; no neural reward model (reward hacking) & R1 \S2.2.1--2.2.3 \\
R1 & R & AIME 2024 pass@1 15.6\% $\rightarrow$ 71.0\%, 86.7\% majority voting; self-evolution: longer thinking, reflection, ``aha moment'' & R1 \S2.2.4 \\
R2 & R (drawbacks) & Poor readability; language mixing within one chain of thought & R1 \S2.2.4 (quoted) \\
X1 & I (abandoned) & Process reward model: hard to define steps, hard to judge intermediate correctness, reward hacking and retraining cost & R1 \S4.2 \\
X2 & I (abandoned) & MCTS over tokens: exponential search space, node extension limit $\rightarrow$ local optima; value model hard to train, no iterative self-improvement & R1 \S4.2 \\
IN1 & IN & Cold-start data designed with human priors should make the output readable and give better performance than R1-Zero; keep the RL & R1 \S2.3, \S2.3.1 (Readability, Potential) \\
C2 & C & Can a little high-quality cold-start data improve reasoning or speed convergence? How to train a user-friendly model with readable CoT and general capability? & R1 \S2.3 \\
I2 & I & DeepSeek-R1: cold start (thousands of long CoT) $\rightarrow$ reasoning RL with language-consistency reward $\rightarrow$ rejection-sampling SFT (600k + 200k) $\rightarrow$ RL over all scenarios & R1 \S2.3.1--2.3.4 \\
R3 & R & Language-consistency reward: slight degradation, more readable (chosen); R1 comparable to o1-1217 on reasoning tasks & R1 \S2.3.2, abstract, \S3.1 \\
C3 & C & Small models: distill from R1, or run large-scale RL directly? & R1 \S2.4, \S4.1 \\
I3a & I & Distill 800k R1 samples into Qwen/Llama 1.5B--70B & R1 \S2.4 \\
I3b & I & Large-scale RL on Qwen-32B-Base directly & R1 \S4.1 \\
R4 & R & Distilled 32B beats RL-trained 32B (AIME 2024 72.6 vs 47.0); ``smaller models relying on the large-scale RL ... may not even achieve the performance of distillation'' & R1 \S4.1, Table 6 (quoted) \\
C4 & C (open) & General capability; language mixing beyond Chinese/English; prompt sensitivity; RL for software engineering (slow evaluation) & R1 \S5 \\
\bottomrule
\end{longtable}}
\addtocounter{table}{-1}

{\scriptsize
\begin{longtable}{>{\raggedright\arraybackslash}p{0.035\linewidth}>{\raggedright\arraybackslash}p{0.25\linewidth}>{\raggedright\arraybackslash}p{0.595\linewidth}}
\toprule
\textbf{\#} & \textbf{edge} & \textbf{transition} \\
\midrule
\endfirsthead
\toprule \textbf{\#} & \textbf{edge} & \textbf{transition} \\ \midrule \endhead
1 & RQ0, L0 $\rightarrow$ IN0 $\rightarrow$ C1 & State the aim of developing reasoning without supervised data; ask whether RL alone suffices and how to supervise it \\
2 & C1 $\rightarrow$ I1 $\rightarrow$ R1, R2 & Run pure RL with rule-based rewards; observe the capability and the drawbacks \\
2a & C1 $\dashrightarrow$ X1 & Try a process reward model; abandon, reasons recorded \\
2b & C1 $\dashrightarrow$ X2 & Try MCTS; abandon, reasons recorded \\
3 & R1, R2 $\rightarrow$ IN1 $\rightarrow$ C2 & Read the promising results and the drawbacks together; expect cold-start data to fix readability and raise performance \\
4 & C2 $\rightarrow$ I2 $\rightarrow$ R3 & Add cold start and three more stages; pay a small accuracy cost for readability \\
5 & I2 $\rightarrow$ C3 $\rightarrow$ I3a, I3b $\rightarrow$ R4 & Fork on small models; resolve by experiment in favor of distillation \\
6 & R4 $\rightarrow$ C4 & Record what remains open \\
\bottomrule
\end{longtable}}
\addtocounter{table}{-1}

\subsection{Cross-case summary}\label{app:cases:summary}

The first table gives the type, source, and size of each trajectory, and the second gives what changed in it and what a loop that regenerates the whole idea would keep. Cases 1 to 5 are presented in \S\ref{sec:cases} and Cases 6 to 8 in this appendix.

{\scriptsize\setlength{\tabcolsep}{3pt}
\begin{longtable}{>{\raggedright\arraybackslash}p{0.04\linewidth}>{\raggedright\arraybackslash}p{0.12\linewidth}>{\raggedright\arraybackslash}p{0.33\linewidth}>{\raggedright\arraybackslash}p{0.11\linewidth}>{\raggedright\arraybackslash}p{0.15\linewidth}>{\raggedright\arraybackslash}p{0.15\linewidth}}
\toprule
\textbf{Case} & \textbf{Type} & \textbf{Who states the reasons} & \textbf{Rounds} & \textbf{Evidence from experiments or studies} & \textbf{Evidence from literature} \\
\midrule
\endfirsthead
\toprule \textbf{Case} & \textbf{Type} & \textbf{Who states the reasons} & \textbf{Rounds} & \textbf{Evidence from experiments or studies} & \textbf{Evidence from literature} \\ \midrule \endhead
1 & method & the authors (annotated, or traced and confirmed) & 1 & 2 & 2 \\
\hline
2 & application & the authors (annotated, or traced and confirmed) & 4 & 3 (formative interviews, user study, benchmark) & 3 (one is a data source) \\
\hline
3 & method (ideation system) & the paper and the authors (annotated, or traced and confirmed), in the paper's order of presentation & 3 & 4 & 3 \\
\hline
4 & system (research agent) & the paper, two versions, and the authors (annotated, or traced and confirmed) & 3 + an open challenge & 4 (one an observation of the agent drifting) & 5 (one an external resource, one the group's own later paper) \\
\hline
5 & empirical insight & the papers (shared title, eight shared authors) & 3 & 3 & 1 \\
\midrule
6 & method & the authors (blog) & 3.5 & 7 & 2 \\
\hline
7 & evaluation of ideation & the same three authors (second intro) & 1 + an open second round & 4 (two with people) & 1 \\
\hline
8 & method & the organization (report, incl. dead ends) & 3 + 2 dead ends + 1 fork & 4 & 1 (own prior work) \\
\bottomrule
\end{longtable}}
\addtocounter{table}{-1}

\medskip

{\scriptsize\setlength{\tabcolsep}{3pt}
\begin{longtable}{>{\raggedright\arraybackslash}p{0.04\linewidth}>{\raggedright\arraybackslash}p{0.24\linewidth}>{\raggedright\arraybackslash}p{0.36\linewidth}>{\raggedright\arraybackslash}p{0.26\linewidth}}
\toprule
\textbf{Case} & \textbf{Decisive cognitive pattern} & \textbf{Node changed / evidence / recorded reason} & \textbf{What a loop that regenerates the whole idea from a score or summary keeps} \\
\midrule
\endfirsthead
\toprule \textbf{Case} & \textbf{Decisive cognitive pattern} & \textbf{Node changed / evidence / recorded reason} & \textbf{What a loop that regenerates the whole idea from a score or summary keeps} \\ \midrule \endhead
1 & abstract structure, substitute one slot & LM slot of the switch / R0 + re-read L0 / pointer-generator is a switch between LM and copy & the final model, not the attribution \\
\hline
2 & transfer a parameterized cognitive model; build the missing data; move from patient to therapist & conditioning input, styles, then the object (patient $\rightarrow$ therapist) / R0 + L1, R0, R1 + IN2 / structured cognition; expert-named styles; the cognitive model is central on both sides & a better prompt, not the conclusion about the judge or the move to the therapist \\
\hline
3 & take a trade-off from prior work as a property of the task; give each dimension its own handle; localize & the generator (per-dimension rewards $\rightarrow$ per-dimension controllers), then where emphasis applies / L1 (trade-off in prior work), for both / the dimensions are interdependent; parts of an idea need different emphasis & a better reward model, not the prior-work trade-off that motivated the controllers and the decoder \\
\hline
4 & take components from other papers one at a time; add a person where the agent drifts; adopt a better method once it exists & the experiment agent, then its resources and a human channel, then the idea agent (prompted $\rightarrow$ trained) / L1 + L2, then L3 + R0, then L4 / ReAct from MLAgentBench and feedback from ResearchAgent; retrieval from the Hub matches models better, and a person corrects the drift; the better ideation method now exists & a better system, not which component came from which paper or why each was added \\
\hline
5 & sharpen, mechanize, ablate what the mechanism predicts & claim scope, twice / R1, R2 / IN0 predicts content-independence; residual slow absorption & a good score, not the two ablations the mechanism predicted \\
\midrule
6 & diagnose both baselines, combine; then rule out alternatives by experiment & sampling source and grading, then the training target / R0 + L0 + L1, then R3 + R4 / compounding error and sparse reward; finite-batch drift & a data-mixing recipe, not the experiment that ruled mixing out \\
\hline
7 & change the judge & the judge (idea reviews $\rightarrow$ executed projects), then an insight about the judge / R1 agreement and diversity numbers, then R2 + R3 / idea-level judgment is subjective; execution evaluation does not speculate & a more novel generator, not the moved judge \\
\hline
8 & act on promising results and stated drawbacks together; record what failed & the data regime (none $\rightarrow$ cold start), plus two abandoned branches / R1 + R2, \S4.2 / readability and expected performance; step definition and reward hacking; search space and value model & a stronger model, not the two branches it must not retry \\
\bottomrule
\end{longtable}}
\addtocounter{table}{-1}

\section{A Proposed Protocol for Comparing Improvement Processes}\label{app:protocol}

This appendix records the design of the controlled comparison referred to in \S\ref{sec:eval:pilot}, which this paper proposes and does not run, together with the details of the pilot from which the judge finding of \S\ref{sec:eval:judge} is taken. Nothing in the paper depends on the comparison's outcome. The pipeline, seeds, prompts and judge code exist and were pilot-tested, and are released with the paper so that the comparison can be run as specified.

\textbf{Question and what is held fixed.} The protocol isolates one variable. It asks whether an idea is improved by localized, reason-recorded edits on a typed graph or by whole-idea regeneration, under the same generator, the same output-token budget and the same retrieval tool. It does not score ideas absolutely, and it does not compare against any system's released output on equal footing.

\textbf{Models.} All five conditions run on the same generator, claude-sonnet-4-5, chosen for its early training cutoff (July 2025, per the vendor's model overview) so that the seed window can be wide; older models of the same family were not reachable, and a smaller model of the same generation failed a protocol test by writing one long draft instead of localized edits. The judge is a newer model with retrieval, claude-opus-5-5. The two cutoffs may differ, because the judge sees only the final idea text and the evidence it retrieves and never reconstructs a target.

\textbf{Seeds.} The seeds are twenty-six usable ideas of type method (15), empirical insight (7), and application (4). Every seed is paired with a hidden target that is the same authors' own next version of the idea, so that seed and target form a within-author update path in the scoping of \S\ref{sec:cases}, and every target is an arXiv v1 submitted after 2025-08-01, the month after the generator's cutoff, so that no condition can recover it from memory. Twenty-one are retro seeds. Following the RPG-retro procedure, a model is given the title and abstract of the target and asked to reconstruct an early version of the idea, keeping the question and general direction and removing the key contribution, which is stored as the hidden target and never shown to any improvement condition. The starting point is reconstructed, but the step from it to the target is the authors' own. Five are same-author pairs in which both papers exist. The earlier paper, mapped to the schema without weakening, is the seed, and the later paper is the target. One pair is the Part I to Part II step of Case 5, with eight shared authors, and the other four share three to six authors with their targets. Four retro seeds whose targets coincided with those of the pairs were replaced by the pairs. Three cross-group pairs used in an earlier version of the protocol are archived, and their follow-up papers became retro seeds. Four screens apply. (i) Date: the target's first version postdates the cutoff. (ii) Contamination: the generator is given the title alone and asked to write the abstract, and the judge model scores its coverage of the actual abstract from 0 to 4; a score of 3 or more excludes the seed, which removed three of 29 candidates whose titles state the mechanism. (iii) Manual review that the seed is genuinely early and does not leak the target. (iv) Retrieval leakage: because web retrieval can return the target paper itself, which happened for two of the three pilot seeds, the retrieval layer excludes the target by title and arXiv identifier and every run is scanned afterwards for the target among its literature nodes. Applied to the eight update paths of \S\ref{sec:cases}, the same screens leave one usable path. Case 1 ends in a 2020 paper, Cases 2, 3, 4, 7, and 8 end in papers whose first versions are inside the generator's training data, and the on-policy distillation post of Case 6, although it postdates the cutoff, is reconstructed from its title alone with coverage 3. Case 5 scores 1. The case paths are therefore evidence for the framework and not data for the protocol.

\textbf{Conditions.} Each of the five conditions (Table~\ref{tab:conditions}) has the same budget of T = 12,000 output tokens, spent over as many calls as fit (at most 24), and each tool condition may run at most Q = 3 searches returning five items each. The pilot of \S\ref{sec:eval:pilot} used an earlier budget unit, K = 6 calls with a cap of 2,000 output tokens per call, and is the reason for the change. The cap was stated in the prompt but not enforced by the interface, and single calls reached 3,700 output tokens, so the protocol counts the tokens actually generated. All conditions produce a final idea in one schema (title, problem, key insight, method, experiment plan, expected results, assumptions and risks) and the judge sees only that schema, never a trajectory, a graph or an edit log.

\begin{table}[h]
\centering\scriptsize
\caption{The five equal-budget conditions.}\label{tab:conditions}
\begin{tabular}{>{\raggedright\arraybackslash}p{0.194\linewidth}>{\raggedright\arraybackslash}p{0.119\linewidth}>{\raggedright\arraybackslash}p{0.374\linewidth}>{\raggedright\arraybackslash}p{0.187\linewidth}}
\toprule
\textbf{Condition} & \textbf{Loop type} & \textbf{Procedure} & \textbf{Corresponds to} \\
\midrule
One-shot rewrite & none & One rewrite with the whole budget & generate-score-rewrite systems \\
Review loop & reviewer feedback & Rounds of critique and whole-idea rewrite until the budget is spent, no tools & ResearchAgent-style \\
Literature loop & retrieval & as above, with one search per round, up to Q, whose results are visible next round; whole-idea rewrite & SciMON / Nova-style \\
RPG loop & typed graph & Transitions, until the budget is spent, of type insight, challenge, search or idea update; the search is executed by the same tool and its results become literature nodes; an idea update is a localized edit of named schema fields, each citing an insight or literature node created since the previous update, with untouched fields kept verbatim; an edit log records which field changed because of which node & this paper \\
RPG without outside information & typed graph, closed & as above with search and literature removed & ablation \\
\bottomrule
\end{tabular}
\end{table}

A result loop (Dolphin-style, real experimental results fed back) is a sixth condition that this version does not run, because the seeds have no experiments; simulating results would be circular. Its absence is a limitation and an agenda item.

\textbf{External one-shot baselines.} Published outputs of one-shot systems, e.g., ResearchTown \citep{yu2025researchtown} and the Ramon Llull's Thinking Machine \citep[not to be confused with the Thinking Machines Lab blog post of Appendix~\ref{app:cases:2}]{zhao2025llull}, enter the same pairwise judge in two roles. Their outputs are the real-system baseline of the one-shot tier, and the same outputs serve as seeds on which the five conditions add a fixed budget. The comparison of an external output with its improved versions is not equal-budget and is reported as ``gain after a fixed added budget'', never as ``RPG beats system X''.

\textbf{Judging: three signals, reported side by side and never combined.} The first is blind pairwise preference between two final ideas from the same seed, on novelty, excitement, feasibility, effectiveness, and overall quality, the five criteria of \citet{si2024novel}. Each pair is judged twice with positions swapped and counts only if both orders agree; win rates carry 95\% Wilson intervals. The judge model differs from, and is newer than, the generation model. Every judgment carries a written rationale that is stored for audit. Node identifiers and version markers are stripped from all outputs before judging so that the judge cannot infer the condition. The second is verified novelty. Two or three novelty claims are extracted from each version, literature is retrieved for each with the same tool the conditions used, with the target excluded, and the pairwise novelty judgment is made with that evidence visible, a step that the pilot below motivated. A claim that retrieval cannot match is recorded as unsupported rather than as unrefuted, because in the judge study of \S\ref{sec:eval:judge} such claims passed the check. The third, reported in its own table, is alignment with the authors' real next step. Given the seed, the hidden target and two final ideas, the judge says which is closer to the step the authors actually took and scores each 0 to 4 for coverage of the target's key idea, mechanism and expected result. For the same-author pairs the judge first writes down the axis along which the authors changed the idea and then says whether each condition's main change lies on that axis, a check less sensitive to wording than coverage. Disagreement between the first and third signals is itself a result, since a process preferred on alignment and dispreferred by the text judge is the judge problem of \S\ref{sec:eval:judge}. Alignment is not quality, since an idea better than the target can score low, and outputs of the external one-shot systems, which have no target, do not enter this table. A human subset of 40 pairs, stratified by condition pair with hidden sources, is judged by PhD students, and its agreement with the model judge is reported; Case 2 is the reason this step is not optional.

\textbf{Hypotheses.} The protocol tests four hypotheses. The first is that the RPG loop is preferred over the one-shot rewrite and the review loop; the second, that it is preferred over the literature loop, which has the same tool, since parity would mean that retrieval and not the graph drives the improvement; the third, that removing outside information removes most of the advantage; and the fourth, that the RPG loop's output lies closer to the authors' actual next step than the output of whole-idea regeneration. These are the predictions the framework makes, stated before any run; \S\ref{sec:disc} says what their failure would mean.

\textbf{Two smaller studies that reuse the same components.} The first is a next-state prediction on annotated update paths. Given the prefix of a path up to a transition, an agent predicts the type and content of the next state in one call, with type accuracy needing no judge and content scored 0 to 4 against the authors' real next node, and the recorded abandoned branches of Case 8 give negative next steps that a system should not re-propose once their reasons are on the graph. On the present case set only Case 5 passes the screens, so the study was run once, on that path, as a demonstration. At nine prediction points the agent given the RPG prefix predicted the type of the next state in seven and a plain-text baseline in three, with mean content coverage of 2.00 and 1.06. One path is not evidence, and the study is the minimal form of learning from update paths (\S\ref{sec:insights:learning}). The second, a judge study on the seeds, was run and is reported in \S\ref{sec:eval:judge}. For each of the 26 seeds, the authors' real next version, written in the schema from the target abstract, is paired with a tool-free one-shot rewrite of the seed, and the text-level and verified-novelty judges each decide blind and position-swapped, with retrieval for the verified judge restricted to work posted before the target. It gives a second instance of the judge problem beside Case 2 and says nothing about the RPG against any system.

\textbf{What the design guards against.} The judge is independent of the RPG (different and newer model, sees no graph, positions swapped, rationales auditable, human calibration). Budgets are equal and actual spend is reported in both call and token units. Seeds and targets are post-cutoff and contamination-screened, and the target is always the same authors' next version. Length and structure bias are limited by the shared schema and an explicit instruction. Retrieval leakage is filtered and scanned. Three weaknesses remain. The retro seeds are model-reconstructed starting points, the loop has no result nodes, and the graph conditions consume more input tokens, since their input is the linearized graph, a difference that is reported and not corrected for.

\begin{table}[htb]
\centering\small
\caption{Pilot (n = 3 seeds): mean spend per run and overall wins against the RPG loop.}
\label{tab:pilot}
\begin{tabular}{lrrrc}
\toprule
\textbf{Condition} & \textbf{Calls} & \textbf{Out. tok.} & \textbf{In. tok.} & \textbf{Wins vs RPG} \\
\midrule
One-shot rewrite & 1.0 & 2,096 & 5,274 & 1 of 3 \\
Review loop & 5.7 & 14,026 & 37,997 & 3 of 3 \\
Literature loop & 6.0 & 13,430 & 43,828 & 3 of 3 \\
RPG loop & 6.0 & 5,006 & 51,031 & --- \\
RPG, no outside info. & 6.0 & 5,793 & 45,363 & 0 of 3 \\
\bottomrule
\end{tabular}
\end{table}

\textbf{Pilot, used as a judge finding.} Three seeds were run through all five conditions and all thirty pairs judged (generator claude-sonnet-5-5, judge claude-opus-5-5; direction only); Table~\ref{tab:pilot} gives spend and outcomes. The pilot used a call-count budget of six calls and a newer generator than the design above, and both the budget unit and the novelty judgment were changed after it and before any run under the protocol, so its numbers are not comparable with such a run and are reported only for the judge finding. The first flaw is the budget unit. At equal \emph{call} count, the review and literature loops spent about 13,000 to 14,000 output tokens per run, the RPG loop about 5,000, because four of its six transitions were insights, challenges or searches that produce no idea text, and its idea updates changed only a few fields. The RPG condition therefore ran on about a third of the generation compute. The remedy, recorded in the protocol's change log, is an equal \emph{output-token} budget of 12,000 generated tokens per condition; the RPG loop then takes as many transitions as fit, roughly twelve to sixteen.

The second flaw is the judge's notion of novelty. In the rationales, the recurring point is that the RPG version's own citations count against it, for example ``B's main hypothesis, the passive-adaptive gap, is already reported by The Distillation Game (2026), which B cites itself'', so novelty goes to the other side; the review loop, which has no tools, makes bolder claims that the judge cannot check and does not penalize. The judge therefore rewards claims that look new without checking whether they are. The remedy is the verified-novelty judgment described under Judging above, and the text-only and verified judgments are both reported.

\end{document}